\documentclass[5p]{elsarticle}
\usepackage{booktabs,multirow,longtable}       
\usepackage{amsmath,amssymb,eucal,mathtools}
\usepackage{url,comment}
\usepackage{furumath}
\usepackage{algorithm,algorithmic}
\usepackage{color}

\newcommand{\SOM}[1]{SOM$^{#1}$}
\newcommand{\KSMM}[1]{KSMM$^{#1}$}
\newcommand{\DT}{{D_\cT}}
\newcommand{\DV}{{D_\cV}}
\newcommand{\DL}{{D_\cL}}

\newcommand{\betai}{\beta\inv}
\newcommand{\eqc}{\stackrel{c}{=}}
\newcommand{\e}{\textit{e}}
\newcommand{\m}{\textit{m}}
\journal{Neurocomputing}
\begin{document}
\begin{frontmatter}%
\title{Multi-task manifold learning for small sample size datasets}
\author[Kyutech]{Hideaki Ishibashi}
\cortext[cor]{Corresponding author}
\ead{ishibashi@brain.kyutech.ac.jp}
\author[Kyutech,Horiba]{Kazushi Higa}
\ead{kazushi.higa@horiba.com}
\author[Kyutech]{Tetsuo Furukawa\corref{cor}}
\ead{furukawa@brain.kyutech.ac.jp}
\address[Kyutech]{Kyushu Institute of Technology,
2--4 Hibikino, Wakamatsu-ku, Kitakyushu 808-0196, Japan}
\address[Horiba]{Horiba Ltd., Kyoto, Japan}
%
\begin{abstract}
In this study, we develop a method for multi-task manifold learning. The method aims to improve the performance of manifold learning for multiple tasks, particularly when each task has a small number of samples. Furthermore, the method also aims to generate new samples for new tasks, in addition to new samples for existing tasks. In the proposed method, we use two different types of information transfer: instance transfer and model transfer. For instance transfer, datasets are merged among similar tasks, whereas for model transfer, the manifold models are averaged among similar tasks. For this purpose, the proposed method consists of a set of generative manifold models corresponding to the tasks, which are integrated into a general model of a fiber bundle. We applied the proposed method to artificial datasets and face image sets, and the results showed that the method was able to estimate the manifolds, even for a tiny number of samples.
\end{abstract}
\end{frontmatter}

\section{Introduction}

To model a high-dimensional dataset, it is often assumed that the data points are distributed in a low-dimensional nonlinear subspace, that is, a manifold. Such manifold-based methods are useful in data visualization and unsupervised modeling, and have been applied in many fields \cite{Huo2007,Pless2009}. A common problem of the manifold-based approach is that it requires a sufficient number of samples to capture the overall shape of the manifold. If there are only a few samples in a high-dimensional space, the samples cover the manifold very sparsely, and then it is difficult to estimate the manifold shape. This is referred to as the {\em small sample size problem} in the literature \cite{Wang2011}.

As a typical example, we consider a face image dataset because face images of a person are modeled by a manifold \cite{Shan2005,Chang2006}. To estimate the face manifold of a person, we need many photographs with various expressions or various poses. However, it is typically difficult to obtain such an exhaustive image set for a single person. If we only have a limited number of facial expressions for the person, it seems almost impossible to estimate a face manifold in which many unknown expressions are contained. Thus, we use multi-task learning; if we can utilize face images of many other people, an exhaustive dataset is no longer necessary.

In the single-task scenario, the aim of manifold learning is to map a dataset to a low-dimensional latent space by modeling the data distribution by a manifold. Thus, in the multi-task scenario, the aim is to map a set of datasets to a common low-dimensional space. For this purpose, multiple manifold learning tasks are executed in parallel, with transferring the information between tasks. Therefore, the success of multi-task learning depends on what and how information is transferred between tasks.

In this study, we use a generative manifold model approach, which is a class of manifold learning methods. Particularly, we chose the method that models a manifold by kernel smoothing, that is, kernel smoothing manifold model (KSMM). Therefore, our target is to develop a multi-task KSMM (MT-KSMM), which works for small sample size datasets. Additionally, MT-KSMM also aims to estimate a general model, which can be applied to new tasks. Thus, our target method covers meta-learning of manifold models \cite{Hospedales2021}.

In the proposed method, we introduce two different types of information transfer between tasks: {\em instance transfer} and {\em model transfer}. For the instance transfer, the given datasets are merged among the similar tasks, whereas for the model transfer, the estimated manifold models are regularized so that they get closer among the similar tasks. To execute these information transfers, we introduce an extra KSMM into MT-KSMM, whereby the manifold models of tasks are integrated into a general model of a fiber bundle. In addition, the extra KSMM also estimates the similarities between tasks, by which the amount of information transfer is controlled. Thus the extra KSMM works as the command center which governs the KSMMs for tasks. Hereafter, the extra KSMM is referred to as the higher-order KSMM (higher-KSMM, in short), whereas the KSMMs for tasks are referred to as the lower-KSMMs. Thus, the key ideas of this work are (1) using two styles of information transfer alternately, and (2) using the hierarchical structure consisting of the lower-KSMMs and the higher-KSMM{}.

The remainder of this paper is structured as follows: We introduce related work in Section~2, and describe the theoretical framework in Section~3. We present the proposed method in Section~4 and experimental results in Section~5. In Section~6, we discuss the study from the viewpoint of information geometry, and in the final section, we present the conclusion.

\section{Related work}

\subsection{Multi-task unsupervised learning}

Multi-task learning is a paradigm of machine learning that aims to improve performance by learning similar tasks simultaneously \cite{Caruana1997,Zhang2018review}. Many studies have been conducted on multi-task learning, particularly in supervised learning settings. By contrast, few studies have been conducted on multi-task unsupervised learning. Some studies on multi-task clustering have been reported \cite{Zhang2012,Zhang2015,Zhang2017,Zhang2018}.

To date, few studies have been conducted on dimensionality reduction, subspace methods, and manifold learning. To the best of our knowledge, a study on multi-task principal component analysis (PCA) is the only research that is expressly aimed at the multi-task learning of subspace methods \cite{Yamane2016}. In multi-task PCA, the given datasets are modeled by linear subspaces, which are regularized so that they are closed to each other on the Grassmannian manifold.

Regarding nonlinear subspace methods, to the best of our knowledge, the most closely related work is the higher-order self-organizing map (\SOM2, which is also called the `SOM of SOMs') \cite{Furukawa2009,Furukawa2005}. Although \SOM2 is designed for multi-level modeling rather than multi-task learning, the function of \SOM2 can also be regarded as multi-task learning. \SOM2 is described in detail below.

\subsection{Multi-level modeling}

Multi-level modeling (or hierarchical modeling) aims to obtain higher models of tasks in addition to modeling each task \cite{Dedrick2009}. Although multi-level modeling does not aim to improve the performance of individual tasks, the research areas of multi-level modeling and multi-task learning overlap. In fact, multi-level modeling is sometimes adopted as an approach for multi-task learning \cite{Zweig2013,Han2015a,Han2015b}.

As in the case of multi-task learning, most studies conducted on multi-level modeling have considered supervised learning, particularly linear models. Among the multi-level unsupervised modeling approaches, \SOM2 aims to model a set of self-organizing maps (SOMs) using a higher-order SOM (higher-SOM) \cite{Furukawa2009,Furukawa2005}. In \SOM2, the lower-order SOMs (lower-SOMs) model each task using a nonlinear manifold, whereas the higher-SOM models those manifolds by another nonlinear manifold in function space. As a result, the entire dataset is represented by a product manifold, that is, a fiber bundle. Thus, the manifolds represented by the lower-SOMs are the fibers, whereas the higher-SOM represents the base space of the fiber bundle. Additionally, the relations between tasks are visualized in the low-dimensional space of the tasks, in which similar tasks are arranged nearer to each other and different tasks are arranged further from each other.

The learning algorithm of \SOM2 is not a simple cascade of SOMs. The higher and lower-SOMs learn in parallel and affect each other. Thus, whereas the higher-SOM learns the set of lower models, the lower-SOMs are regularized by the higher-SOM so that similar tasks are represented by similar manifolds. Thus, \SOM2 has many aspects of multi-task learning. In fact, \SOM2 has been applied to several areas of multi-task learning, such as unsupervised modeling of face images of various people \cite{Jiang2008}, nonlinear dynamical systems with latent state variables \cite{Ohkubo2009,Ohkubo2011}, shapes of various objects \cite{Yakushiji2011,Yakushiji2013}, and people of various groups \cite{Ishibashi2016,Ishibashi2018}. In this sense, \SOM2 is one of the earliest works on multi-task unsupervised learning for nonlinear subspace methods.

Although \SOM2 works like multi-task learning, it is still difficult for it to estimate manifolds when the sample size is small. Additionally, \SOM2 has several limitations that originate from SOM, such as poor manifold representation caused by the discrete grid nodes. In this study, we try to overcome these limitations of \SOM2 by replacing SOM with KSMM{}. Then we extend it for multi-task learning so that it estimates the manifold shape more accurately, even for a small sample size.

\subsection{Meta-learning}

Recently, the concept of meta-learning has been gaining importance \cite{Hospedales2021}. Meta-learning aims to solve unseen future tasks efficiently through learning multiple existing tasks. This is in contrast to multi-task learning, which aims to improve the learning performance of existing tasks. Because MT-KSMM aims to obtain a general model that can represent future tasks in addition to existing tasks, our aim covers both multi-task learning and meta-learning. Similar to the cases of multi-task learning and multi-level learning, meta-learning of unsupervised learning has rarely been performed. Particularly, to the best of our knowledge, meta-learning in an unsupervised manner of unsupervised learning has not been reported, except \SOM2. In this sense, \SOM2 is a method in a unique position. In this study, we aim to develop a novel method for multi-task, multi-level, and meta-learning of manifold models based on \SOM2.

\subsection{Generative manifold modeling}

Nonlinear methods for dimensionality reduction are roughly categorized into two groups. The first group is methods that project data points {\em from} high-dimensional space {\em to} low-dimensional space. Most dimensionality reduction methods, including many manifold learning approaches, are classified in this group. By contrast, the second group explicitly estimates the mapping {\em from} low-dimensional latent space {\em to} the manifold in high-dimensional visible space \cite{Lawrence2005}. Thus a nonlinear embedding is estimated by this group. Because new samples on the manifold can be generated by using the estimated embedding, we refer to the second group as {\em generative manifold modeling} in this study. An advantage of generative manifold model is that it does not suffer from the pre-image problem and the out-of-sample extension problem \cite{Bunte2012}. Thus, by generating new samples between the given data points, generative manifold model completes the manifold shape entirely. Additionally, because the embedding is explicitly estimated, generative manifold model allows the direct measurement of the distance between two manifolds in function space. These are the reasons that we chose the generative manifold model approach.

The representative methods of generative manifold modeling are generative topographic mapping (GTM) \cite{Bishop1998}, the Gaussian process latent variable model (GPLVM) \cite{Lawrence2004,Lawrence2005}, and unsupervised kernel regression (UKR) \cite{Meinicke2005}, which originate from the SOM \cite{Kohonen1982}. To estimate the embedding and the latent variables, GPLVM and GTM employ the Bayesian approach that use a Gaussian process, whereas UKR and SOM employ the non-Bayesian approach that use a kernel smoother. Additionally, GPLVM and UKR represent the manifold in a non-parametric manner, whereas GTM and SOM represent it in parametric manner. In this study, we use a parametric kernel smoothing approach like the original SOM, for ease of information transfer between tasks.

\subsection{Unsupervised manifold alignment}
\label{manifold-alignment}

In nonlinear subspace methods, there is always arbitrariness in the determination of the coordinate system of the latent space. For example, any orthogonal transformation of the latent space yields an equivalent result. Additionally, any nonlinear distortion along the manifold is also allowed.

In single-task learning, such arbitrariness is not a problem, but it causes serious problems in multi-task learning. Because the coordinate system of the latent space is determined differently for each task, the learning results become incompatible between tasks. Therefore, it prevents knowledge transfer between tasks.

To solve this problem, we need to know the correspondences between two different manifolds, or we need to regularize them so that they share the same coordinate system. This is known as the manifold alignment problem, which is challenging to solve under the unsupervised condition \cite{Wang2009}. This is one reason why the multi-task subspace method is difficult. In the case of multi-task PCA, they escaped this problem by mapping them to a Grassmannian manifold \cite{Yamane2016}; however, this approach is only possible for the linear subspace case. Therefore, manifold alignment is an unavoidable problem, which is challenged in this study.

\section{Problem formulation}
\label{sec:formulation}

In this section, we first describe generative manifold modeling in the single-task case, and then we present the problem formulation of the multi-task case.  The notation used in this paper is described in Appendix A, and the symbol list is presented in \tblref{symbols}.

\begin{figure*}
    \centering
    \begin{tabular}{cc} \small
        \includegraphics[scale=0.6]{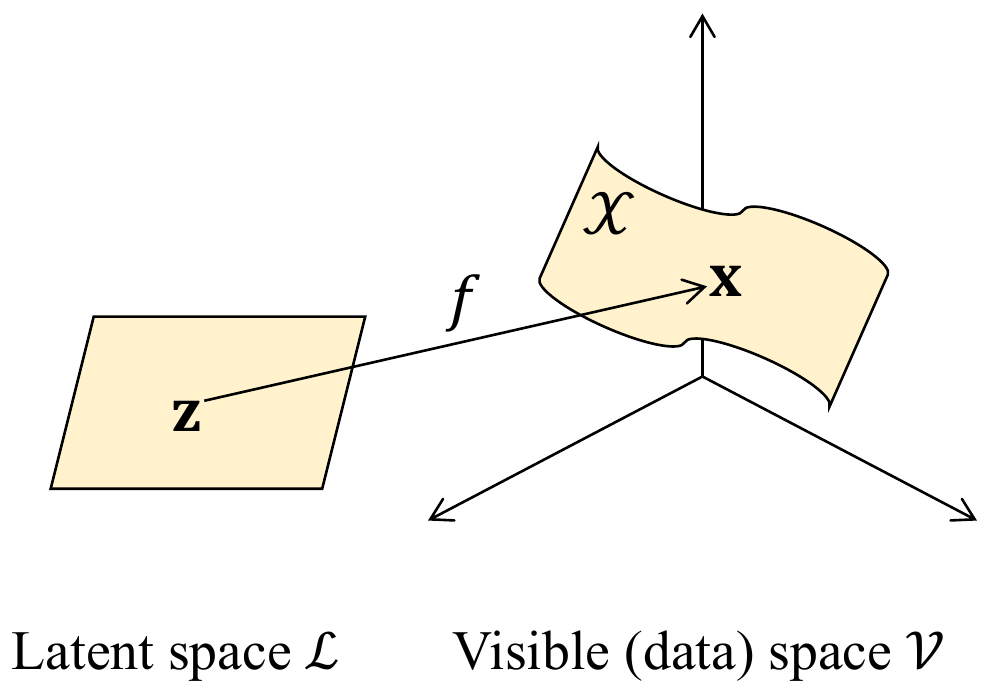} &
        \includegraphics[scale=0.6]{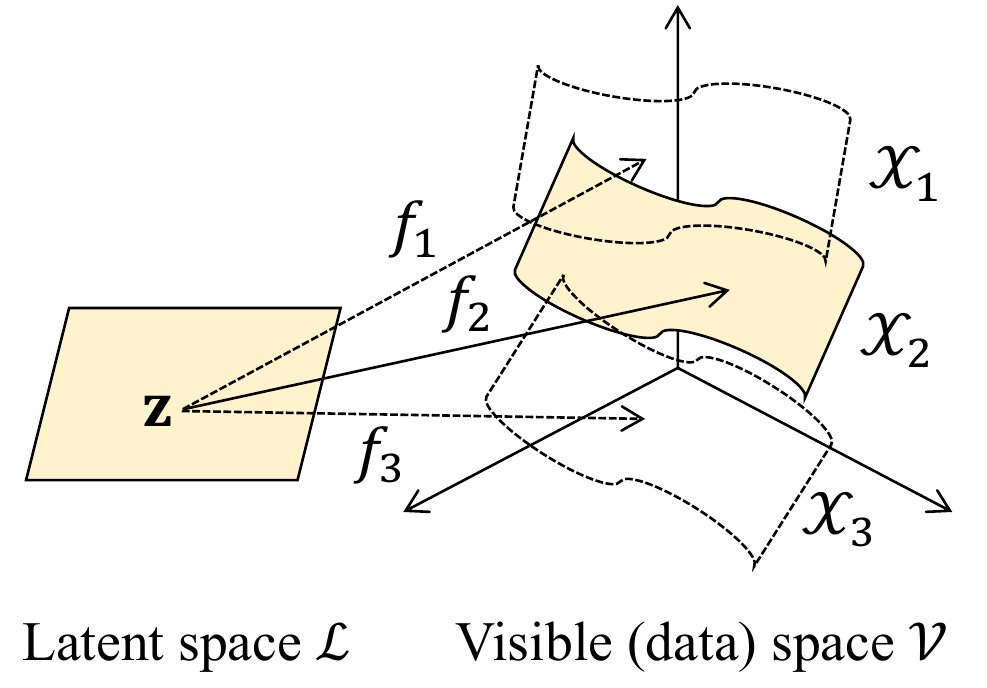} \\
        (a) & (b) \\[1em]
        \includegraphics[scale=0.6]{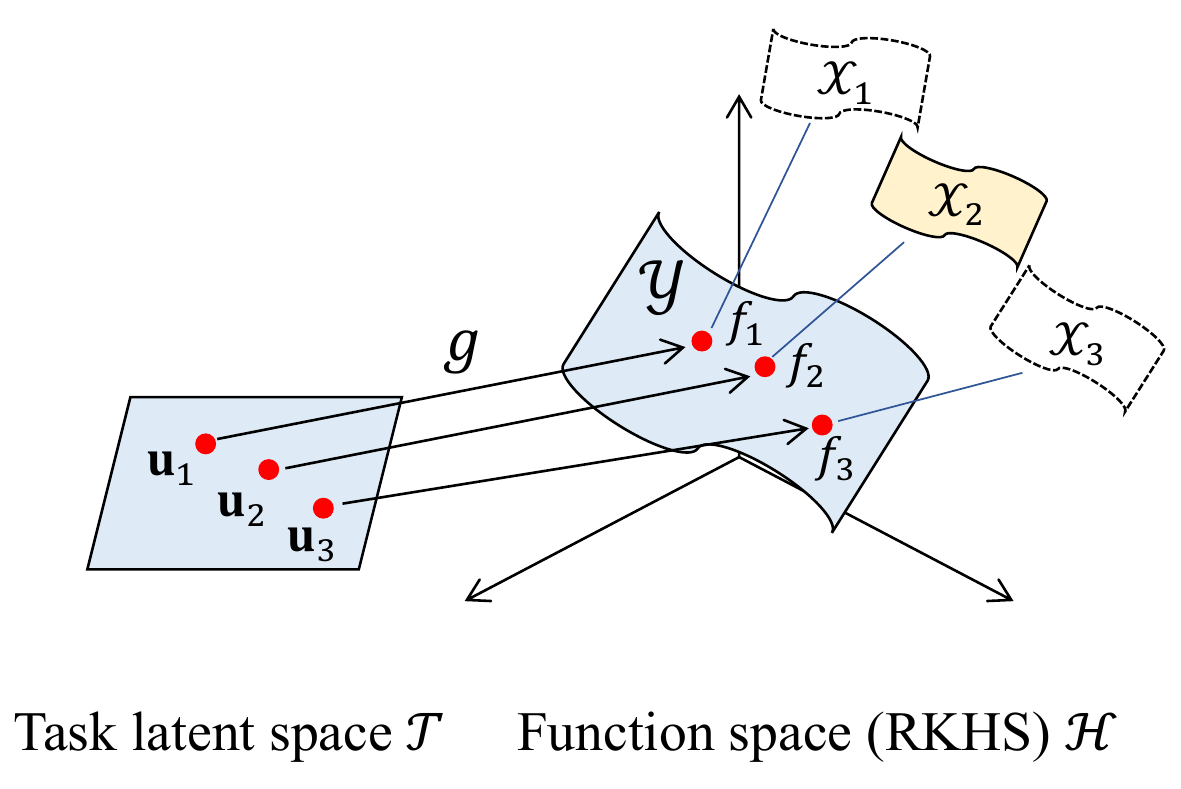} &
        \includegraphics[scale=0.6]{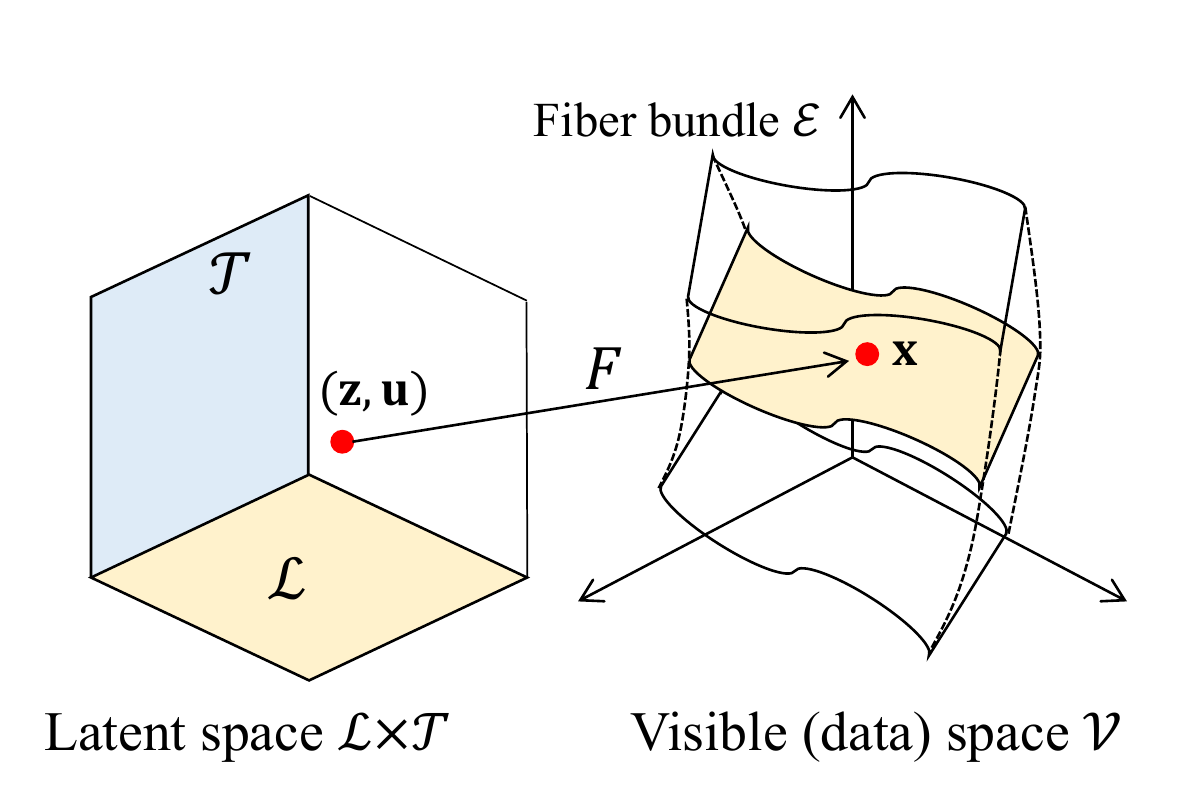} \\
        (c) & (d)
    \end{tabular}
    \caption{Generative model for multi-task manifold learning. (a) Single-task case. A datum $\vx$ on manifold $\cX\subseteq\cV$ is generated from the embedding $f$ and the latent variable $\vz\in\cL$. (b) Multi-task case. The latent space $\cL$ is common to all tasks, whereas the embedding $f$ is different between tasks.  (c) Manifold assumption for tasks. Embedding $f$ is distributed on a manifold $\cY$ in function space $\cH$. (d) Using the manifold assumption, the entire datasets are modeled using a fiber bundle $\cE$.}
    \figlabel{Model}
\end{figure*}

\subsection{Problem formulation for single-task learning}

First, we describe the problem formulation for the single-task case (\figref{Model} (a)). Let $\cV=\bbR^{\DV}$ and $\cL\subseteq\bbR^{\DL}$ be high-dimensional visible space and low-dimensional latent space, respectively, and let $X=\bigl\{\vx_n\bigr\}_{n=1}^N$ be the observed sample set, where $\vx_n\in\cV$. In the case of a face image dataset, $X$ is the set of face images of a single person with various expressions or poses, and $\vz_n\in\eL$ is the sample latent variable corresponding to $\vx_n$, which represents the intrinsic property of the image, such as the expression or pose. The main purpose of manifold learning is to map $X$ to $\cL$ by assuming that sample set $X$ is modeled by a manifold $\cX\subseteq\cV$, which is homeomorphic to $\cL$. Thus, a homeomorphism $\pi\colon \cX\xrightarrow{\sim}\cL$ can be defined, by which $\vx\in\cX$ is projected to $\vz=\pi(\vx)$. Note that actual samples are not distributed in $\cX$ exactly because of observation noise. Thus, the observed sample $\vx_n$ is represented as $\vx_n= \pi\inv(\vz_n)+\vepsilon_n$, where $\vepsilon_n \sim \cN(\v0,\betai\vI_{\DV})$ is observation noise. Generative manifold modeling not only aims to estimate the sample latent variables $Z=\left\{\vz_n\right\}_{n=1}^N$, but also to estimate the embedding $f\coloneqq\pi\inv$ explicitly. Thus, the probabilistic generative model is estimated as $q(\vx,\vz\mid f)=\cN(\vx\mid f(\vz),\betai\vI_\DV)\,p(\vz)$. Because $f$ is expected to be a smooth continuous mapping, we assume that $f$ is a member of a reproducing kernel Hilbert space (RKHS) $\cH=\{f\mid f\colon\cL\to\cV\}$.

\subsection{Problem formulation for multi-task learning}
\label{sec:problem_formulation}

Now we formulate the problem in which we have $I$ tasks. Thus, we have $I$ sample sets $\bigl\{X_1,\dots,X_I\bigr\}$, where $X_i = \bigl\{\vx_{ij}\bigr\}_{j=1}^{J_i}$. For example, in the case of face image datasets, we have $I$ image sets of $I$ people, with $J_i$ expressions or poses for each person. We denote the entire sample set by $X = \bigcup_i X_i = \bigl\{\vx_n\bigr\}_{n=1}^N$, where $N=\sum_i J_i$. We also denote the entire sample set by matrix $\vX=\bigl(\vx\T_n\bigr) \in \bbR^{N\times D_\cV}$. In the same manner, $Z_i=\left\{\vz_{ij}\right\}_{j=1}^{J_i}$ is the latent variable set of task $i$, which corresponds to $X_i$, whereas the entire latent variable set is represented as $Z=\left\{\vz_n\right\}_{n=1}^N$. Additionally, let $i_n$ be the task index of sample $n$, and let $\cN_i$ be the index set of samples that belong to task $i$. To simplify the explanation, we consider the case in which each dataset has the same number of samples ({\em sample per task\/}: S/T), that is, $J_i\equiv J$.

Similar to the single-task case, we assume that $\{X_i\}$ is modeled by manifolds $\{\cX_i\}$, which are all homeomorphic to the common latent space $\cL\subseteq\bbR^{\DL}$. Thus, we can define a smooth embedding $f_i\colon\cL\to\cX_i\subseteq\cV$ for each task, where $f_i\coloneqq\pi_i\inv$ (\figref{Model} (b)). By considering the observation noise, we represent the probability distribution of task $i$ by $q_i(\vx,\vz\mid f_i)=\cN(\vx\mid f_i(\vz),\betai\vI_{D_\cV})\,p(\vz)$. In this study, the individual embedding $f_i$ is referred to as the {\em task model}.

To apply multi-task learning, the given tasks need to have some similarities or common properties. We make the following two assumptions on the similarities of tasks. First, the sample latent variable $\vz$ represents the intrinsic property of the sample, which is independent of the task. Thus, if $\vx_{ij}$ and $\vx_{i'j'}$ are mapped to the same $\vz$, they are regarded as having the same intrinsic property, even if $\vx_{ij}\ne\vx_{i'j'}$. For example, if both $\vx_{ij}$ and $\vx_{i'j'}$ are face images of different people with the same expression (e.g., smiling), they should be mapped to the similar $\vz$, even if the images look different. Using this assumption, the distance between two manifolds $\cX_i$ and $\cX_{i'}$ can be defined using the norm $\norm{f_i-f_{i'}}_\cH$ as
\begin{align}
 L^2(\cX_i,\cX_{i'})
   &\coloneqq \norm{f_i-f_{i'}}^2_\cH  \nncr
   &= \int_\cL \norm{f_i(\vz)-f_{i'}(\vz)}^2\,dP(\vz),
   \eqlabel{distance}
\end{align}
where $dP(\vz)$ is the measure of $\vz$, which is common to all tasks. In this study, we define the measure as $dP(\vz)=p(\vz)\,d\vz$, where $p(\vz)$ is the prior of $\vz$.

Second, we assume that the set of task models $F=\{f_i\}_{i=1}^I$ is also distributed in a nonlinear manifold $\cY$ embedded into RKHS $\cH$ (\figref{Model} (c)). Thus, $\cY$ is defined by a nonlinear smooth embedding $g\colon\cT\to\cY\subseteq\cH$, where $\cT\subseteq\bbR^{\DT}$ is the low-dimensional latent space for tasks. Under this assumption, all tasks are assigned to the task latent variables $U=\{\vu_i\}_{i=1}^I$, $\vu_i\in\cT$ so that $f_i=g(\vu_i)$. It also means that if $i$ and $i'$ are similar, then $\vu_i\sim\vu_{i'}$ and $\cX_i\sim\cX_{i'}$. As a result, the entire model distribution is represented as
\begin{align}
  q(\vx,\vz,\vu\mid G)
    = \N{\vx}{G(\vz,\vu),\betai\vI_{D_\cV}}\,p(\vz)\,p(\vu),
  \eqlabel{Q(x,z,u)}
\end{align}
where $G\colon\cL\times\cT \to\cE\subseteq\cV$ is defined as $G(\cdot,\vu)=g(\vu)$, and $\eE$ is the embedded product manifold of $\cL\times\cT$ into $\cV$ (\figref{Model} (d)). In this study, $G$ is referred to as the {\em general model}. Using the general model, each task model is expected to be $f_i(\vz) = G(\vz,\vu_i)$.

We can describe this problem formulation using the terms of the fiber bundle. Suppose that $\cE\subseteq\cV$ is the total space of the fiber bundle, where $\cE\coloneqq G\left(\cL,\cT\right)$. Then we can define a projection $\eta\colon\cE\rightarrow\cT$ so that $\eta(\cX_i)\equiv\vu_i$, where $\cT$ is now referred to as the base space. Thus, the samples that belong to the same task are all projected to the same $\vu$. Under this scenario, $(\cE,\cT, \eta)$ becomes a fiber bundle, and each task manifold $\cX_i$ is regarded as a fiber. Additionally, we can define another projection $\pi\colon\cE\rightarrow\cL$, which satisfies $\pi(\vx_{ij})\equiv\pi_i(\vx_{ij})$. Such a formulation using the fiber bundle has been proposed in some studies \cite{Furukawa2009,Daouda2020}.

Under this scenario, our aim is to determine the general model $G$ in addition to the task model set $F=\bigl\{f_i\bigr\}_{i=1}^I$, and to estimate the latent variables of samples $Z=\left\{\vz_n\right\}_{n=1}^N$ and tasks $U=\left\{\vu_i\right\}_{i=1}^I$. Considering compatibility with SOM, we assume that the latent spaces are square spaces $\cL=[-1,+1]^{D_\cL}$ and $\cT=[-1,+1]^{D_\cT}$, and the priors $p(\vz)$ and $p(\vu)$ are the uniform distribution on $\cL$ and $\cT$\footnote{Rigorously speaking, we need to consider the border effect in the case of compact latent space, such as $[-1,+1]^{D}$. In this study, we ignore such a border effect for ease of implementation. Such an approximation is common in SOM and its variations.}. The difficulty of learning depends on S/T (i.e., $J$) and the number of tasks $I$. Particularly, we are interested in the case in which S/T is too small to capture the manifold shape, whereas we have a sufficient number of tasks for multi-task learning.

\section{Proposed method}

In this section, we first describe the single task of manifold modeling, which is called KSMM{}. Then we describe the information transfer among the tasks. Finally, we present the multi-task KSMM, that is, MT-KSMM{}. In this section, we outline the derivation of the MT-KSMM algorithm, and in Appendix B, we describe the details.

\subsection{Kernel smoothing manifold modeling (KSMM)}
\label{sec:KSMM}

We chose KSMM as a generative manifold modeling method, which estimates the embedding using a kernel smoother. Thus, our aim is to develop the multi-task KSMM, that is, MT-KSMM{}. KSMM is a theoretical generalization of SOM, which has been proposed in many studies \cite{Luttrell1989,Cheng1997,Graepel1998,Heskes2001,Verbeek2005}. The main difference between SOM and KSMM is that the former discretizes the latent space $\cL$ to regular grid nodes, whereas the latter treats it as continuous space. According to previous studies, the cost function of KSMM is given by
\begin{multline}
 E[Z,f\mid X] \\
   = \frac{\beta}{2N} \sum_n
   \int_\cL h_\cL(\vz\mid\vz_n) \norm{f(\vz)-\vx_n}^2\,d\vz,
 \eqlabel{KSMM1}
\end{multline}
where $h_\cL(\vz\mid\vz')$ is the non-negative smoothing kernel defined on $\cL$, which is typically $h_\cL(\vz\mid\vz') = \cN\bigl(\vz\given\vz', \lambda_\cL^2\vI\bigr)$~\cite{Heskes2001}. If we regard $h_\cL(\vz\mid\vz')$ as the probability density that represents the uncertainty of $\vz_n$, the cost function \eqref{KSMM1} equals the cross-entropy between the data distribution $p(\vx,\vz)$ and the model distribution $q(\vx,\vz)$ (ignoring the constant), which are given by
\begin{align}
 p(\vx,\vz\given X,Z)
   &= \frac1N \sum_n \cN(\vx\mid\vx_n,\betai\vI)\,h_\cL(\vz\mid\vz_n)
   \eqlabel{p(x,z)}\\
 q(\vx,\vz\given f)
   &= \N{\vx}{f(\vz),\betai\vI}\,p(\vz).
   \eqlabel{q(x,z)}
\end{align}
Thus, $E[Z,f\mid X] \eqc H[p(\vx,\vz\mid X,Z),q(\vx,\vz\mid f)]$, where `$\eqc$' means that both hands are equal except the constant, and $H[\cdot,\cdot]$ denotes the cross-entropy.

The KSMM algorithm is the expectation maximization (EM) algorithm in the broad sense \cite{Heskes2001,Verbeek2005}, in which the E and M steps are iterated alternately until the cost function \eqref{KSMM1} converges (\figref{architecture} (a)). In the E step, the cost function \eqref{KSMM1} is optimized with respect to $Z$, whereas $f$ is fixed as follows:
\begin{align}
   \Hat\vz_n &\coloneqq \argmin_{\vz_n} \int h_\cL(\vz\mid\vz_n) \norm{\Hat f(\vz)-\vx_n}^2 d\vz
      \eqlabel{KSMM-Estep1}\\
      &\simeq \argmin_{\vz_n} \norm{\Hat f(\vz_n)-\vx_n}^2,
    \eqlabel{KSMM-Estep2}
\end{align}
where $\Hat\vz_n$ and $\Hat f$ represent the estimators of $\vz_n$ and $f$, respectively\footnote{In this study, the (tentative) estimators are denoted by hat $\Hat{\ }$ when we need to indicate expressly. We also denote the information transfer result by tilde $\Tilde{\ }$.}. The latent variable estimation obtained by \eqref{KSMM-Estep2} is used in the original SOM \cite{Kohonen1982} and some similar methods \cite{Meinicke2005}. Note that we can rewrite \eqref{KSMM-Estep2} as
\begin{align}
  \Hat\vz_n
  &\simeq \argmax_{\vz_n} \log q(\vx_n,\vz_n \mid \Hat f), \notag
\end{align}
which means the maximum log-likelihood estimator. We optimize \eqref{KSMM-Estep2} using SOM-like grid search to avoid local minima for early loops of iterations, and then they are updated by the gradient method in the later loops.

By contrast, in the M step, \eqref{KSMM1} is optimized with respect to $f$, whereas $Z$ is fixed, as follows:
\begin{align}
  \Hat f & \coloneqq \argmin_f \sum_n \int h_\cL(\vz\mid\Hat\vz_n)\norm{f(\vz)-\vx_n}^2 d\vz \nncr
  &= \argmin_f H\left[p(\vx,\vz\mid X,\Hat Z),q(\vx,\vz,\mid f)\right].
    \eqlabel{KSMM-Mstep1}
\end{align}
The solution of \eqref{KSMM-Mstep1} is given by the kernel smoothing of samples as
\begin{align}
    \Hat f(\vz)  & \coloneqq \frac{\sum_n h_\cL(\vz\mid\Hat\vz_n)\,\vx_n}{\sum_{n'} h_\cL(\vz\mid\Hat\vz_{n'})}.
    \eqlabel{KSMM-Mstep2}
\end{align}
These E and M steps are executed repeatedly until the cost function \eqref{KSMM1} converges. In the proposed method, embedding $f$ is represented parametrically using orthonormal bases. The details are provided in Section~\ref{sec:parametric}.

There are several reasons why we chose kernel smoothing-based manifold modeling rather than Gaussian process-based modeling. First, because KSMM is a generalization of SOM, it is easy to extend \SOM2 to MT-KSMM{} naturally. Second, because the information transfers are made by the non-negative mixture of datasets or models, a kernel smoother can be used consistently through the method. Third, because the kernel smoother acts as an elastic net that connects the data points \cite{Durbin1987,Utsugi1997}, the kernel smoothing of task models is expected to solve the unsupervised manifold alignment problem by minimizing the distance between manifolds. Finally, as far as we examined, kernel smoothing-based manifold modeling is more stable and less sensitive to sample variations than the Gaussian process. This property seems to be desirable for multi-task learning, particularly when the sample size is small.

\begin{figure*}
    \centering\small
        \includegraphics[width=\textwidth]{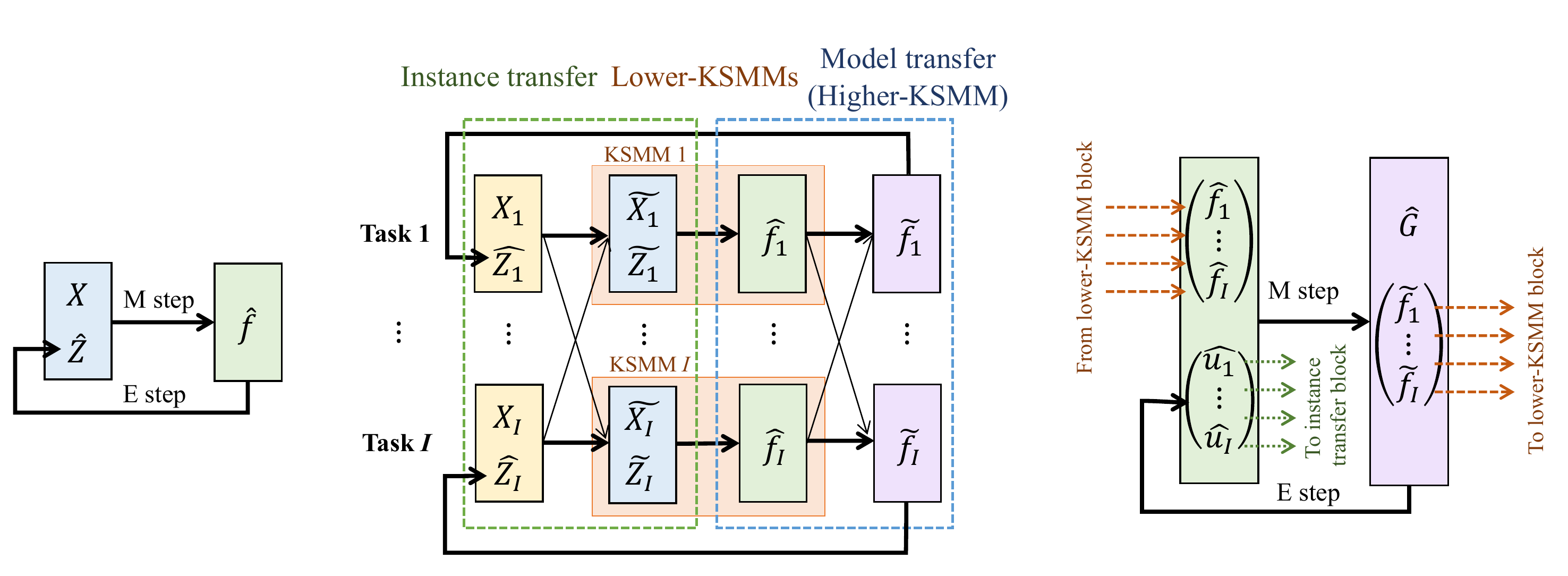} \\
        (a) KSMM \hspace{40mm} (b) MT-KSMM \hspace{40mm} (c) Higher-KSMM \rule{10mm}{0mm}
    \caption{Block diagram of the learning process. (a) Single-task KSMM{}. In the M step, the estimator of embedding $\Hat f$ is updated by the dataset $X=\{\vx_n\}$ and latent variables $\Hat Z=\{\Hat\vz_n\}$, whereas in the E step, $\Hat Z$ is updated by $X$ and $\Hat f$. (b) MT-KSMM{}. To execute $I$ tasks in parallel, MT-KSMM consists of $I$ lower-KSMMs and two information transfer blocks, that is, the instance transfer block and model transfer block (higher-KSMM). (c) The higher-KSMM receives the task models $\Hat F=\{\Hat f_i\}$ from the lower-KSMMs as the input data and outputs the general model $\Hat G$. The general model is fed back to the lower-KSMMs as the model transfer result, whereas the task latent variables $\Hat U=\{\Hat\vu_i\}$ are fed back to the instance transfer block.}
    \figlabel{architecture}
\end{figure*}

\subsection{Information transfer between tasks}

For multi-task learning, the information about each task needs to be transferred to other tasks. Generally, there are three major approaches: {\em feature-based}, {\em parameter-based}, and {\em instance-based} \cite{Zhang2018review}. To summarize, the feature-based approach transfers or shares the intrinsic representation between tasks, the parameter-based approach transfers the model parameters between tasks, and the instance-based approach shares the datasets among the tasks. In the proposed method, we use two information transfer styles: {\em model transfer}, which corresponds to the parameter-based approach, and {\em instance transfer}, which corresponds to the instance-based approach. Although we do not use the feature-based approach explicitly in the proposed method, we can regard the latent space $\cL$, which is common to all tasks, as the intrinsic representation space shared among the tasks. Therefore, the proposed method is related to all information transfer styles of multi-task learning.

In instance transfer, samples of a task are transferred to other tasks, and form a weighted mixture of samples. Thus, if task $i'$ is a neighbor of task $i$ in the task latent space $\cT$, the sample set $X_{i'}$ is merged with the target set $X_i$ as an auxiliary sample set with a larger weight. By contrast, if task $i''$ is far from task $i$ in $\cT$, then $X_{i''}$ is merged with $X_i$ with a small (or zero) weight. We denote the weight of sample $n$ in source task $i_n$ with respect to the target task $i$ by $\rho_{in}$ ($0\leq \rho_{in}\leq 1$). Typically $\rho_{in}$ is defined as
\begin{align}
 \rho_{in} \equiv \rho(\Hat\vu_i\mid\Hat\vu_{i_n}) &=
   \exp\left[-\frac1{2\lambda_\rho^2}\norm{\Hat\vu_i-\Hat\vu_{i_n}}^2\right],
 \eqlabel{step1a}
\end{align}
where $\lambda_\rho$ determines the neighborhood size for data mixing, and $\Hat\vu_i$ is the estimator of the task latent variable $\vu_i$. Then we have the merged dataset $\Tilde X_i=\{(\vx_n,\rho_{in})\}$, where $(\vx_n,\rho_{in})\in\Tilde X_i$ indicates that $\vx_n$ is a member of $\Tilde X_i$ with weight $\rho_{in}$. Similarly, the latent variable set $Z_i$ is also merged with the same weight as $\Tilde Z_i=\left\{(\Hat\vz_n,\rho_{in})\right\}$. The data distribution corresponding to the merged dataset $(\Tilde X_i,\Tilde Z_i)$ becomes
\begin{align}
 \Tilde p_i(\vx,\vz\given \Tilde X_i,\Tilde Z_i)
    &= \frac1{\Tilde J_i} \sum_n \rho_{in} \cN(\vx\mid\vx_n,\betai\vI)
         \,h_\cL(\vz\mid\Hat\vz_n) \notag\\
    &= \frac{\sum_{i'} \rho(\Hat\vu_i\mid\Hat\vu_{i'})\, p_i(\vx,\vz\mid X_i,\Hat Z_i)}{\sum_{i''} \rho(\Hat\vu_i\mid\Hat\vu_{i''})},
 \eqlabel{m-ks}
\end{align}
where $\Tilde J_i=\sum_n \rho_{in}$. Eq.~\eqref{m-ks} means that {\em the data distributions} $P=\{p_i\}$ are smoothed by kernel $\rho(\vu,\vu')$ {\em before} the task models are estimated. By creating a weighted mixture of samples, we can avoid the small sample set problem.

By contrast, in model transfer, each task model $f_i$ is modified so that it becomes similar to the models of the neighboring tasks, as follows:
\begin{align}
  \Tilde f_i(\vz)
    &=\frac{\sum_{i'} h_\cT(\Hat\vu_i\mid\Hat\vu_{i'})\,\Hat f_{i'}(\vz)}{\sum_{i''} h_\cT(\Hat\vu_i\mid\Hat\vu_{i''})},
    \eqlabel{model_transfer}
\end{align}
where $h_\cT(\vu\mid\vu')=\cN(\vu\mid\vu', \lambda^2_\cT\vI)$. Eq.~\eqref{model_transfer} means the kernel smoothing of the task models $F=\{f_i\}$. Thus, the model distribution becomes
\begin{align}
  \Tilde q_i(\vx,\vz\mid \Tilde f_i) &= \cN(\vx\mid \Tilde f_i(\vz), \betai\vI) \, p(\vz).
  \eqlabel{model_transfer2}
\end{align}
Note that \eqref{model_transfer} and \eqref{model_transfer2} are integrated as
\begin{align}
  \log \Tilde q_i(\vx,\vz)
    &\eqc\frac{\sum_{i'} h_\cT(\Hat\vu_i\mid\Hat\vu_{i'}) \log q_{i'}(\vx,\vz)}{\sum_{i''} h_\cT(\Hat\vu_i\mid\Hat\vu_{i''})}.
    \eqlabel{model_transfer3}
\end{align}
Therefore, {\em the log of model distributions} $Q=\{q_i\}$ is smoothed by kernel $h_\cT(\vu\mid\vu')$ {\em after} the task models are estimated.

\subsection{Architecture of multi-task KSMM (MT-KSMM)}

When we have $I$ datasets (i.e., $I$ tasks), we need to execute $I$ KSMMs in parallel. To execute the multi-task learning of KSMMs, the proposed method MT-KSMM introduces two key ideas. The first idea is to use both instance transfer and model transfer, and the second idea is to introduce the higher-KSMM, which integrates the $I$ manifold models estimated by the KSMMs for tasks (i.e., the lower-KSMMs).

As described in \ref{sec:KSMM}, KSMM estimates $\Hat f$ from $X$ and $\Hat Z$ in the M step, whereas $\Hat Z$ is estimated from $X$ and $\Hat f$ in the E step. The first key idea is to use the merged dataset $(\Tilde X_i, \Tilde Z_i)$ that is obtained by instance transfer instead of using the original dataset $(X_i,\Hat Z_i)$. Thus, $\Hat f_i$ is estimated from $\Tilde X_i$ and $\Tilde Z_i$. Similarly, to estimate the latent variables $\Hat Z_i$, the proposed method uses the embedding $\Tilde f_i$ that is obtained using model transfer instead of using $\Hat f_i$ directly (\figref{architecture} (b)).

To execute instance transfer and model transfer, we need to determine the similarities between tasks. For this purpose, the proposed method introduces the higher-KSMM{} (\figref{architecture} (c)). Thus, the higher-KSMM estimates the latent variables of tasks $\Hat U=\left\{\Hat\vu_i\right\}_{i=1}^I$, by which the similarities between tasks are determined. Another important role of the higher-KSMM is to make the model transfer by estimating the general model $\Hat G$ from the task models $\Hat F=\{\Hat f_i\}_{i=1}^I$.

After introducing these two key ideas, the entire architecture of MT-KSMM consists of three blocks: the instance transfer block, lower-KSMM block, and higher-KSMM block (i.e., the model transfer block) (\figref{architecture} (b)). The aim of the lower-KSMM is to estimate the task models $f_i$ for each task $i$ from the merged dataset $(\Tilde X_i, \Tilde Z_i)$. Without considering information transfer, the cost function of the lower KSMM for task $i$ is given by
\begin{align}
  & E_i^\text{lower}[\Tilde Z_i,f_i\mid \Tilde X_i] \notag\\
  & \qquad \coloneqq \frac\beta{2\Tilde J_i} \sum_n \int h_\cL(\vz,\vz_n)
             \rho_{in} \norm{f_i(\vz)-\vx_n}^2\, d\vz \notag\\
  & \qquad \eqc H\left[\Tilde p_i(\vx,\vz\mid \Tilde X_i, \Tilde Z_i), q_i(\vx,\vz\mid f_i)\right].
  \eqlabel{costfunction1}
\end{align}
However, the minimization of the cost function is affected by the information transfer at every iteration of the EM algorithm. Thus, before executing the M step, $(X_i,\Hat Z_i)$ are merged by the instance transfer as $(\Tilde X_i,\Tilde Z_i)$, whereas before executing the E step, $\Hat f_i$ is replaced by $\Tilde f_i$ as a result of model transfer.

By contrast, the aim of the higher-KSMM is to estimate the general model $G$ by regarding the task models $\Hat F=\{\Hat f_i\}_{i=1}^I$ as the dataset. Thus, the cost function (without considering the influence of other blocks) is given by
\begin{multline}
  E^\text{higher}[U,G\mid \Hat F]
    = \frac{\beta}{2I} \sum_i \int h_\cT(\vu\mid\vu_i) \\
         \norm{G(\cdot,\vu)-\Hat f_i(\cdot)}^2_\cH\, d\vu.
  \eqlabel{costfunction2}
\end{multline}
This cost function equals the cross-entropy between $p(\vx,\vz,\vu\mid \Hat F,U)$ and $q(\vx,\vz,\vu\mid G)$ as
\begin{align*}
        E^\text{higher}[U,G,\mid \Hat F]
      &\eqc H\left[p(\vx,\vz,\vu\mid F,U), q(\vx,\vz,\vu\mid G)\right],
\end{align*}
where
\begin{align*}
    p(\vx,\vz,\vu\mid \Hat F,U)
      &= \frac1I \sum_i \cN(\vx\mid f_i(\vz),\betai\vI)\,
        p(\vz)\,h_\cT(\vu\mid\vu_i) \\
    q(\vx,\vz,\vu\mid G)
      &= \cN(\vx\mid G(\vz,\vu),\betai\vI)\,
        p(\vz)\,p(\vu).
\end{align*}
Note that $p(\vx,\vz,\vu)$ is determined by $\Hat F=\{\Hat f_i\}$ and $U=\{\vu_i\}$, regarding $\Hat F=\{\Hat f_i\}$ as the input data. Therefore, $p(\vx,\vz,\vu)$ is not determined by the observed samples $X$ naively; thus,
\begin{align*}
    p(\vx, \vz, \vu)
      &\ne \frac1N \sum_n \cN(\vx\mid \vx_n,\betai\vI)
       \, h_\cL(\vz\mid\vz_n)\, h_\cT(\vu\mid\vu_{i_n}).
\end{align*}

The lower and higher-KSMMs do not simply optimize the cost functions \eqref{costfunction1} and \eqref{costfunction2} in parallel. The estimated task models $\Hat F=\{\Hat f_i\}_{i=1}^I$ are fed forward to the higher-KSMM from the lower-KSMM as the dataset, whereas the estimated general model $\Hat G$ is fed back from the higher to the lower as a result of model transfer, so that $\Tilde f_i(\cdot)=\Hat G(\cdot,\Hat\vu_i)$. Additionally, the estimated task latent variables $\Hat U=\{\Hat\vu_i\}_{i=1}^I$ are also fed back from the higher-KSMM to the instance transfer block, whereby the merged datasets $\{(\Tilde X_i,\Tilde Z_i)\}_{i=1}^I$ are updated. Such a hierarchical optimization process satisfies the concept of meta-learning \cite{Hospedales2021}.

\begin{algorithm}[t]
\caption{MT-KSMM algorithm}
\label{algorithm}
\begin{algorithmic}
  \STATE For all $n$, initialize $Z=\{\vz_n\}$ randomly. (Step 0)
  \STATE For all $i$, initialize $U=\{\vu_i\}$ randomly. (Step 0)
  \REPEAT
    \STATE For all $(i,n)$, calculate $\rho_{in}$ using \eqref{step1a}. (Step 1)
    \STATE For all $i$, calculate $\vV_i$ using \eqref{step2a}--\eqref{step2b}. (Step 2)
    \STATE Calculate $\uW$ using \eqref{step3a}--\eqref{step3b}. (Step 3)
    \STATE For all $i$, update $\vu_i$ using \eqref{step4}. (Step 4)
    \STATE For all $n$, update $\vz_n$ using \eqref{step5}. (Step 5)
  \UNTIL{the calculation converges.}
\end{algorithmic}
\end{algorithm}

\subsection{MT-KSMM algorithm}

The MT-KSMM algorithm is described as follows (also see Algorithm \ref{algorithm}).
\begin{description}
  \item[\textbf{Step 0: }] \textbf{Initialization}

    For the first loop, the latent variables $\Hat Z=\left\{\Hat\vz_n\right\}$ and $\Hat U=\left\{\Hat\vu_i\right\}$ are initialized randomly.
  \item[\textbf{Step 1: }] \textbf{Instance transfer}

    To execute instance transfer, $\rho_{in}$ is calculated using \eqref{step1a}. Then we have the merged datasets $\Tilde X_i=\{(\vx_n,\rho_{in})\}$ and $\Tilde Z_i=\{(\vz_n,\rho_{in})\}$ for all $i$. This step can be interpreted as the kernel smoothing of the data distribution set $P=\{p_i(\vx,\vz)\}$ using \eqref{m-ks}, and we have $\Tilde P=\{\Tilde p_i(\vx,\vz)\}$ as a result.
  \item[\textbf{Step 2:}] \textbf{M step of the lower-KSMM}

    Task models $\Hat F=\{\Hat f_i\}$ are updated so that the cost function of the lower-KSMM is minimized as
    \begin{align}
      \Hat f_i &\coloneqq \argmin_{f_i} E_i^\text{lower}[\Tilde Z_i,f_i\mid \Tilde X_i], \notag
    \end{align}
    for each task $i$. The solution is given by the kernel smoothing of the merged sample set as follows:
    \begin{align}
      \Hat f_i(\vz) &\coloneqq \frac{\sum_n
        h_\cL(\vz,\Hat\vz_n) \rho_{in}\, \vx_n}{\sum_{n'}
        h_\cL(\vz,\Hat\vz_{n'})\rho_{in}}.
    \eqlabel{step2}
    \end{align}
    Then we have the model distribution set $Q=\{\Hat q_i(\vx,\vz)\}$ by \eqref{q(x,z)}.
  \item[\textbf{Step 3:}] \textbf{M step of the higher-KSMM}

    The general model $\Hat G$ is updated so that the cost function of the higher-KSMM is minimized as
    \begin{align}
       \Hat G &\coloneqq \argmin_G E^\text{higher}[\Hat U,G\mid \Hat F]. \notag
    \end{align}
    The solution is given by
    \begin{align}
        \Hat G(\vz,\vu)&\coloneqq \frac{\sum_i h_\cT(\vu,\Hat\vu_i)\,\Hat f_i(\vz)}{\sum_{i'} h_\cT(\vu,\Hat\vu_{i'})}.
        \eqlabel{step3}
    \end{align}
    Then we have the entire model distribution $q(\vx,\vz,\vu\mid\Hat G)$ using \eqref{Q(x,z,u)}. This process is regarded as model transfer.
  \item[\textbf{Step 4:}] \textbf{E step of the higher-KSMM}

    The task latent variables $\Hat U=\{\vu_i\}$ are updated so that the log-likelihood is maximized as
    \begin{align}
        \Hat\vu_i &\coloneqq \argmax_{\vu_i} \sum_{n\in\cN_i} \log  q(\vx_n,\Hat\vz_n,\vu_i\mid \Hat G) \nncr
          &= \argmin_{\vu_i} \sum_{n\in\cN_i} \norm{\Hat G(\Hat\vz_n,\vu_i)-\vx_n}^2.
        \eqlabel{step4}
    \end{align}
    Then we have the task models as
    \begin{align*}
        \Tilde f_i(\vz) &:= \Hat G(\vz,\Hat\vu_i) \\
        \Tilde q_i(\vx,\vz) &:= q(\vx,\vz\mid \Hat\vu_i, \Hat G),
    \end{align*}
    which are the results of model transfer using \eqref{model_transfer} and \eqref{model_transfer2}. Then $\Tilde F=\{\Tilde f_i\}$ is fed back to the lower-KSMM{}.
  \item[\textbf{Step 5:}] \textbf{E step of the lower-KSMM}

    Finally, the sample latent variables $\Hat Z=\{\vz_n\}$ are updated so that the log-likelihood is maximized as
    \begin{align}
      \Hat\vz_n & \coloneqq \argmax_{\vz_n} \log \Tilde q_{i_n}(\vx_n,\vz_n) \nncr
      &= \argmin_{\vz_n} \norm{\Tilde f_{i_n}(\vz_n)-\vx_n}^2.
      \eqlabel{step5}
    \end{align}
\end{description}
These five steps are iterated until the calculation converges. During the iterations, the length constant of the smoothing kernels $\lambda_\cL$ and $\lambda_\cT$ are gradually reduced just like the ordinary SOM{}.

\subsection{Parametric representation of embeddings}
\label{sec:parametric}

To apply the gradient method, we represent embedding $f$ parametrically using orthonormal basis functions defined on the RKHS (e.g., normalized Legendre polynomials), such as
\begin{align}
 f(\vz) = f(\vz\mid\vV) = \sumL \varphi_l(\vz) \vv_l = \vV\T\vvphi(\vz), \notag
\end{align}
where $\vvphi=(\varphi_1,\dots,\varphi_L)\T$  is the basis set and $\vV\in\bbR^{L\times D_\eX}$ is the coefficient matrix. In the case of single-task KSMM, the estimator of the coefficient matrix $\Hat\vV$ is determined as
\begin{align}
    \Hat\vV=\vA\inv\vB\vX,
    \eqlabel{KSMM-parametric1}
\end{align}
where
\begin{align}
 \vA &= \int \vvphi(\vz)\vvphi\T(\vz)\,\ol{h}_\cL\,d\vz \\
 \vB &= \int \vvphi(\vz)\,\vh_\cL(\vz)\T\,d\vz,
\end{align}
and
\begin{align}
\vh_\cL(\vz) &= \bigl(h_\cL(\vz\mid\Hat\vz_1),\dots,h_\cL(\vz\mid\Hat\vz_N)\bigr)\T\\ \ol{h}_\cL(\vz)&=\sum_n h_\cL(\vz\mid\vz_n).
 \eqlabel{KSMM-parametric3}
\end{align}
Thus, \eqref{KSMM-Mstep2} is replaced by \eqref{KSMM-parametric1}--\eqref{KSMM-parametric3}.

Similarly, in MT-KSMM, the coefficient matrices of the task models are estimated as
\begin{align}
     \Hat \vV_i &= \vA_i\inv \vB_i \vX,
       \eqlabel{step2a}
\end{align}
where
\begin{align}
     \vA_i &= \int \vvphi(\vz)\,\vvphi\T(\vz)\,\ol h_i(\vz)\,d\vz \\
     \vB_i &= \int \vvphi(\vz) \,\vh_i(\vz)\T \,d\vz\\
     \vh_i(\vz) &= \Bigl(\rho_{in}\,h_\cL(\vz\mid\vz_n)\Bigr)_{n=1}^N \\
     \ol h_i(\vz) &= \sum_n \rho_{in}\,h_\cL(\vz\mid\vz_n).
       \eqlabel{step2b}
\end{align}
Thus, \eqref{step2} is replaced by \eqref{step2a}--\eqref{step2b}. Note that $\ul\vV=(\vV_i)_{i=1}^I\in\bbR^{I\times L \times \DV}$ becomes a tensor of order 3.
For the higher-KSMM, the general model $G$ is represented as
\begin{align}
  G(\vz, \vu\mid \ul{\vW})
  &= \sumK \sumL \vw_{kl} \psi_k(\vu) \varphi_l(\vz) \nncr
  &= \ul{\vW} \ttimes1 \vpsi(\vu) \ttimes2 \vvphi(\vz),
  \eqlabel{general_model}
\end{align}
where $\uW \in \bbR^{K\times L \times \DV}$ is the coefficient tensor and $\times_m$ represents the tensor--matrix product, where $\vpsi=(\psi_1,\dots,\psi_K)\T$ is the basis set for the higher-KSMM{}. The coefficient tensor $\ul\vW$ is estimated as
\begin{align}
     \Hat{\ul\vW} &= \ul{\Hat\vV} \ttimes1 \left(\vC\inv \vD\right) \eqlabel{step3a}\\
     \vC &= \int \vpsi(\vu)\,\vpsi\T(\vu)\,\ol{h}_\cT(\vu)\,d\vu
     \eqlabel{step3a-C}\\
     \vD &= \int \vpsi(\vu)\, \vh_\cT\T(\vu)\,d\vu,
     \eqlabel{step3a-D}
\end{align}
where
\begin{align}
     \vh_\cT(\vu) &= \Bigl(h_\cT(\vu\mid\vu_i)\Bigr)_{i=1}^I \\
     \ol{h}_\cT(\vu) &= \sum_i h_\cT(\vu\mid\vu_i).
     \eqlabel{step3b}
\end{align}
Thus, \eqref{step3} is replaced by \eqref{step3a}--\eqref{step3b}. Note that the computational cost of the proposed method with respect to the data size is $\cO(NID_\cV)$ for the lower-KSMMs and $\cO(ID_\cV)$ for the higher-KSMM{}. Therefore, the computation cost is proportional to both the data size $N$ and number of tasks $I$.

\section{Experimental results}

\subsection{Experiment protocol}

To demonstrate the performance of the proposed method, we applied the method to three datasets. The first was an artificial dataset generated from two-dimensional (2D) manifolds, and the second and third were face image datasets of multiple subjects with various poses and expressions, respectively.

To examine how the information transfer improved performance, we compared methods that used different transfer styles, and KSMM was used for all methods as the common platform. Thus, (1) MT-KSMM (the proposed method) used both instance transfer and model transfer, (2) \KSMM2 (the preceding method) used model transfer only, and (3) single-task KSMM did not transfer information between tasks\footnote{It is not able to examine the case of using instance transfer only because the task latent variable $\Hat U=\{\Hat\vu_i\}$ cannot be estimated without using the higher-KSMM{}, which also makes the model transfer.}. \KSMM2 is a modification of the preceding method \SOM2 that replaces the higher and lower-SOMs with KSMMs. Note that this modification improves the performance of \SOM2 by eliminating the quantization error. Similarly, single-task KSMM can be regarded as a SOM with continuous latent space.

In the experiment, we trained the methods on a set of tasks by changing S/T{}. After training was complete, we evaluated the performance both qualitatively and quantitatively using the test data. We used two types of test data: test data of existing tasks and test data of new tasks. The former evaluates the generalization performance for new data of existing tasks that have been learned, and the latter evaluates the generalization performance for new unknown tasks.

We assessed the learning performance quantitatively using two approaches: the root mean square error (RMSE) of the reconstructed samples and the mutual information (MI) between the true and estimated sample latent variables. RMSE (smaller is better) evaluates accuracy in the visible space $\cV$ without considering the compatibility of the sample latent variables, whereas MI (larger is better) evaluates accuracy in the latent space $\cL$, and considers how consistent the sample latent variables are between tasks. We evaluated both RMSE and MI for the test data of existing tasks and new tasks. For existing tasks, we used the task latent variable $\Hat\vu_i$ that was estimated in the training phase, whereas $\Hat\vu_\text{new}$ was estimated by minimizing \eqref{step4} for each new task.

\begin{figure}[t]
\centering
\begin{tabular}{cc}
\small
\includegraphics[scale=0.22]{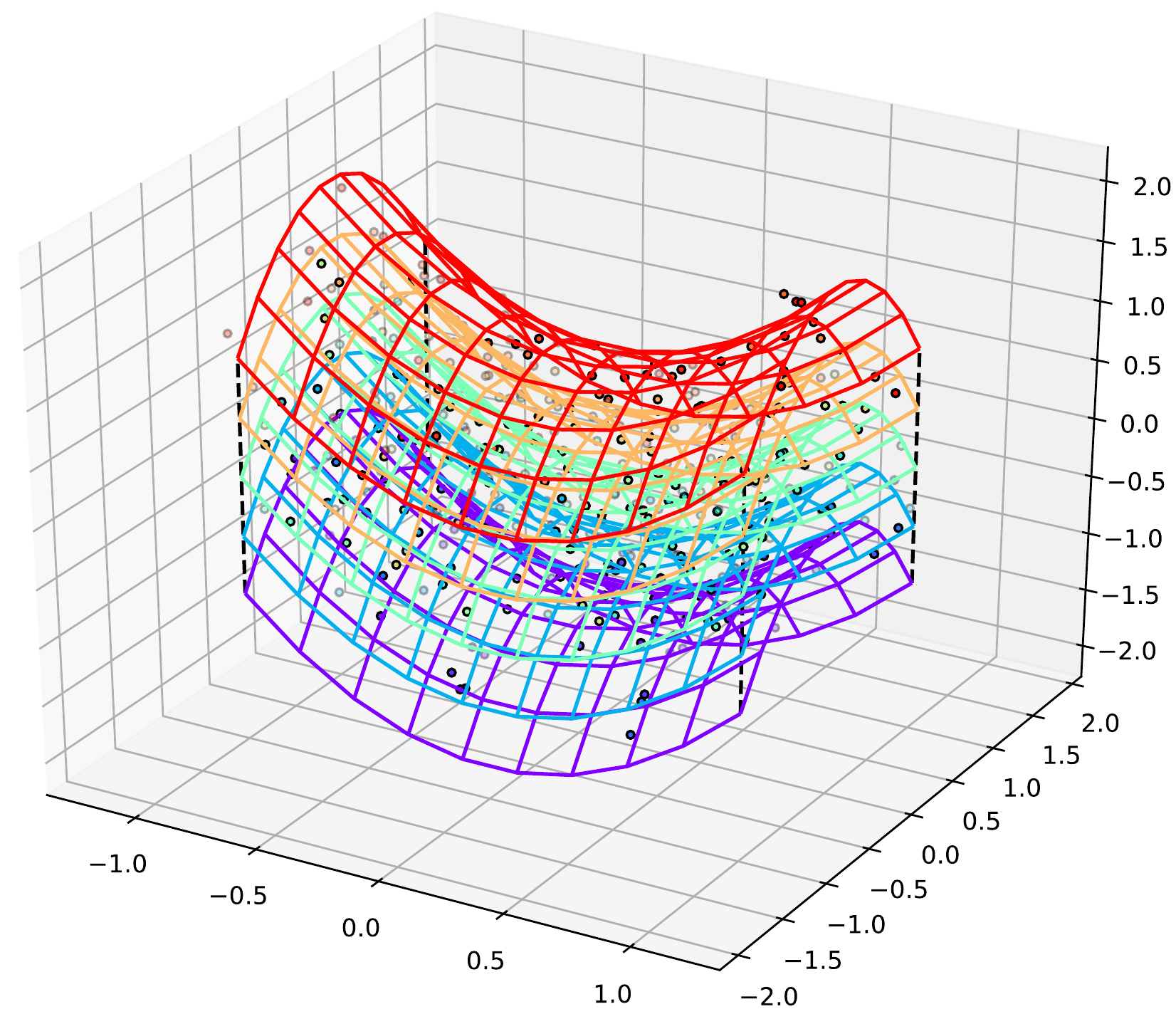} &
\includegraphics[scale=0.22]{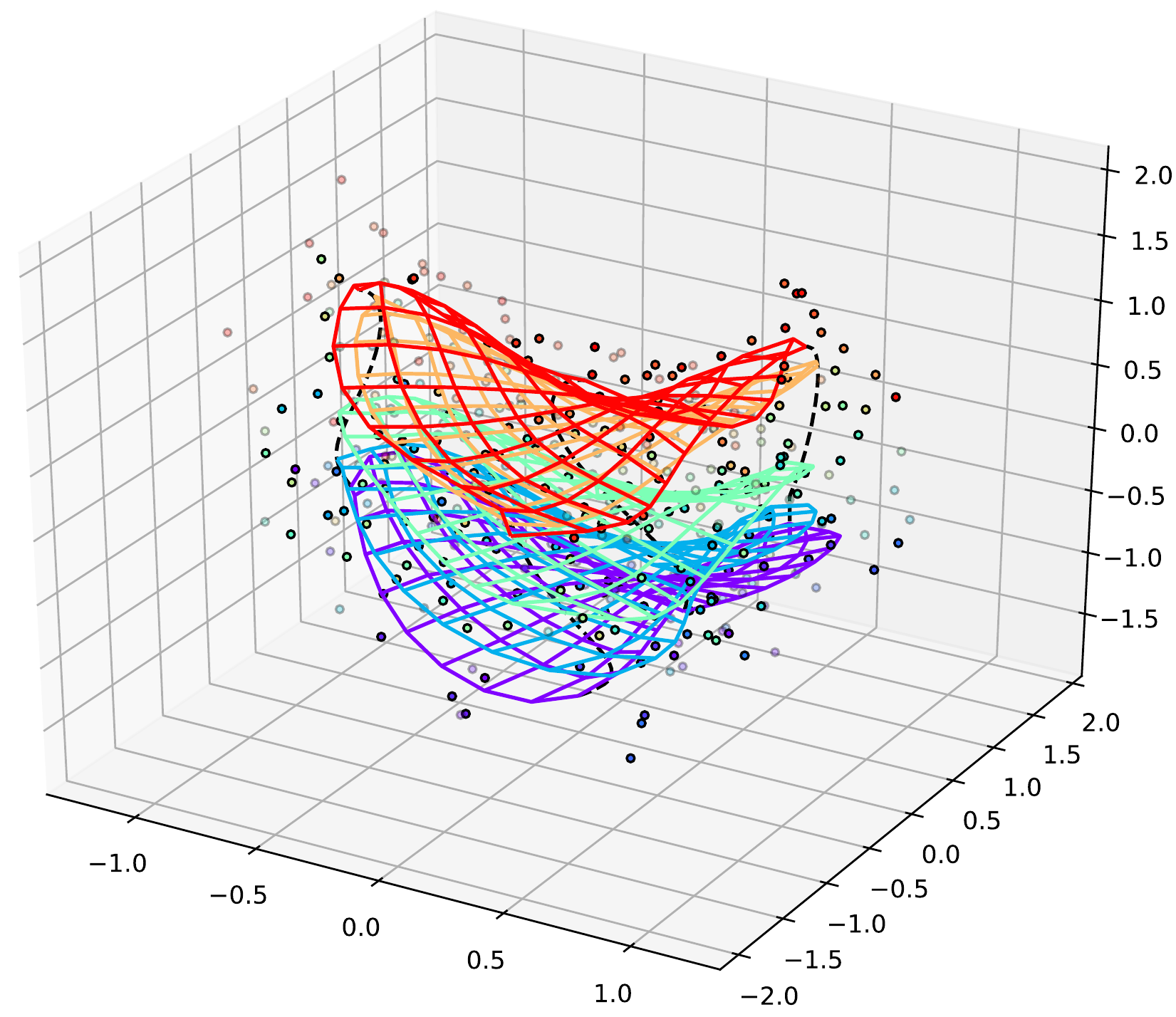} \\
(a) & (b) \\
\includegraphics[scale=0.22]{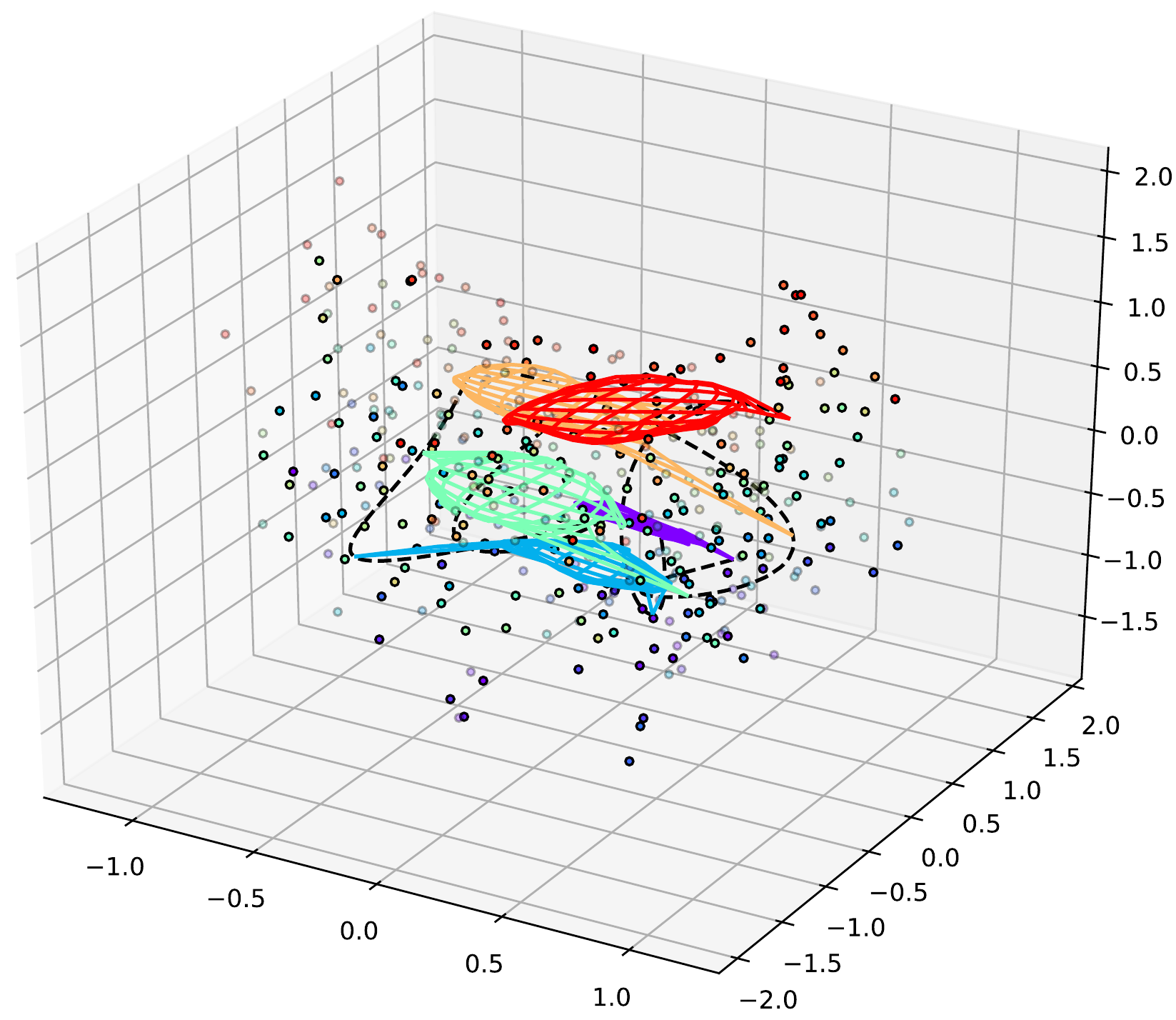} &
\includegraphics[scale=0.22]{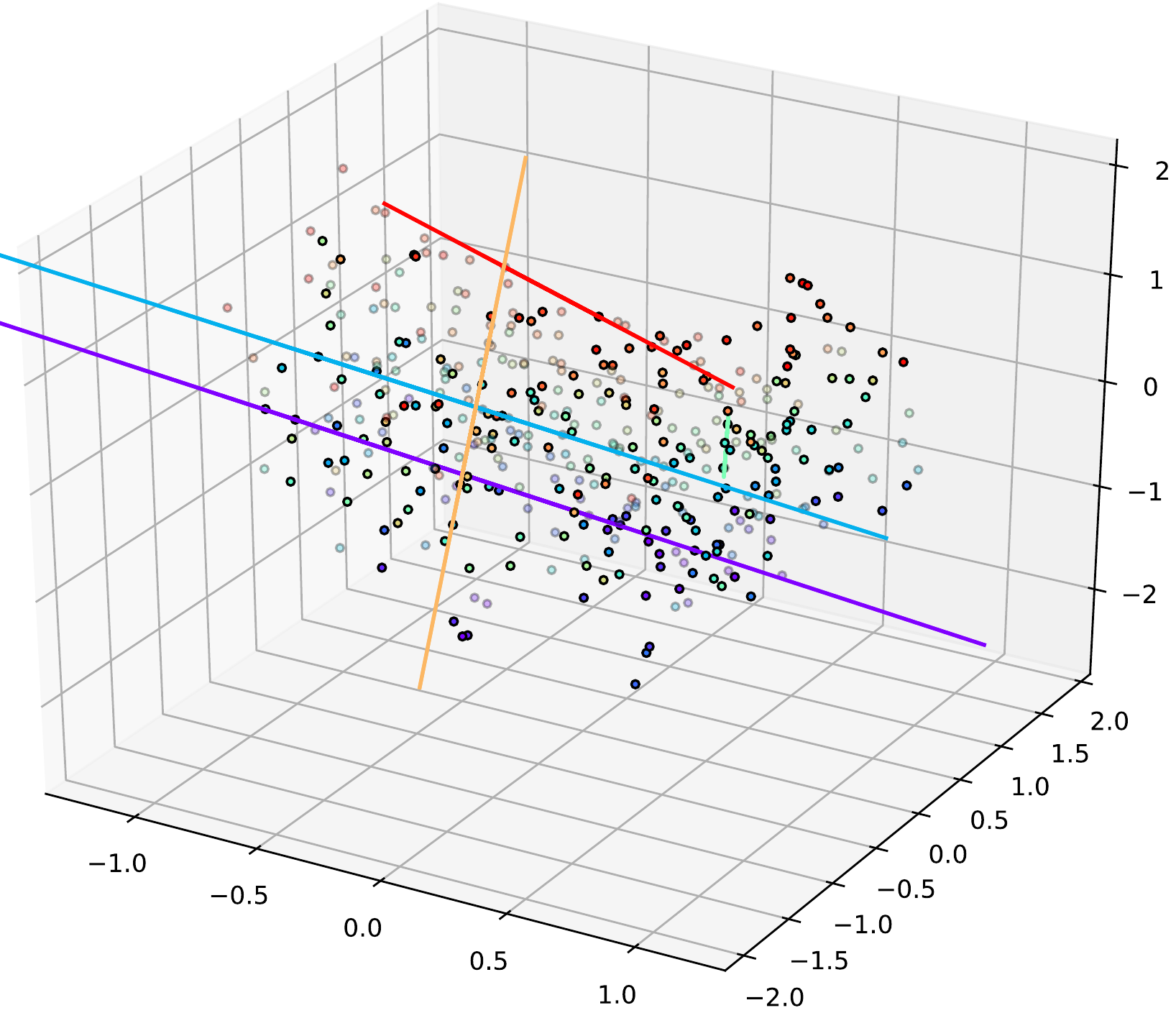} \\
(c) & (d)
\end{tabular}
\caption{Results for the saddle shape artificial dataset. For training, 200 manifolds with different biases and two samples per task (2 S/T) were used. (a) Ground truth. (b) MT-KSMM. (c) \KSMM2 (d) KSMM{}. Five out of 200 manifolds are indicated.}
\figlabel{ex1-1}
\end{figure}
\begin{figure}[t]
\centering
\begin{tabular}{cc}
\small
\includegraphics[scale=0.19]{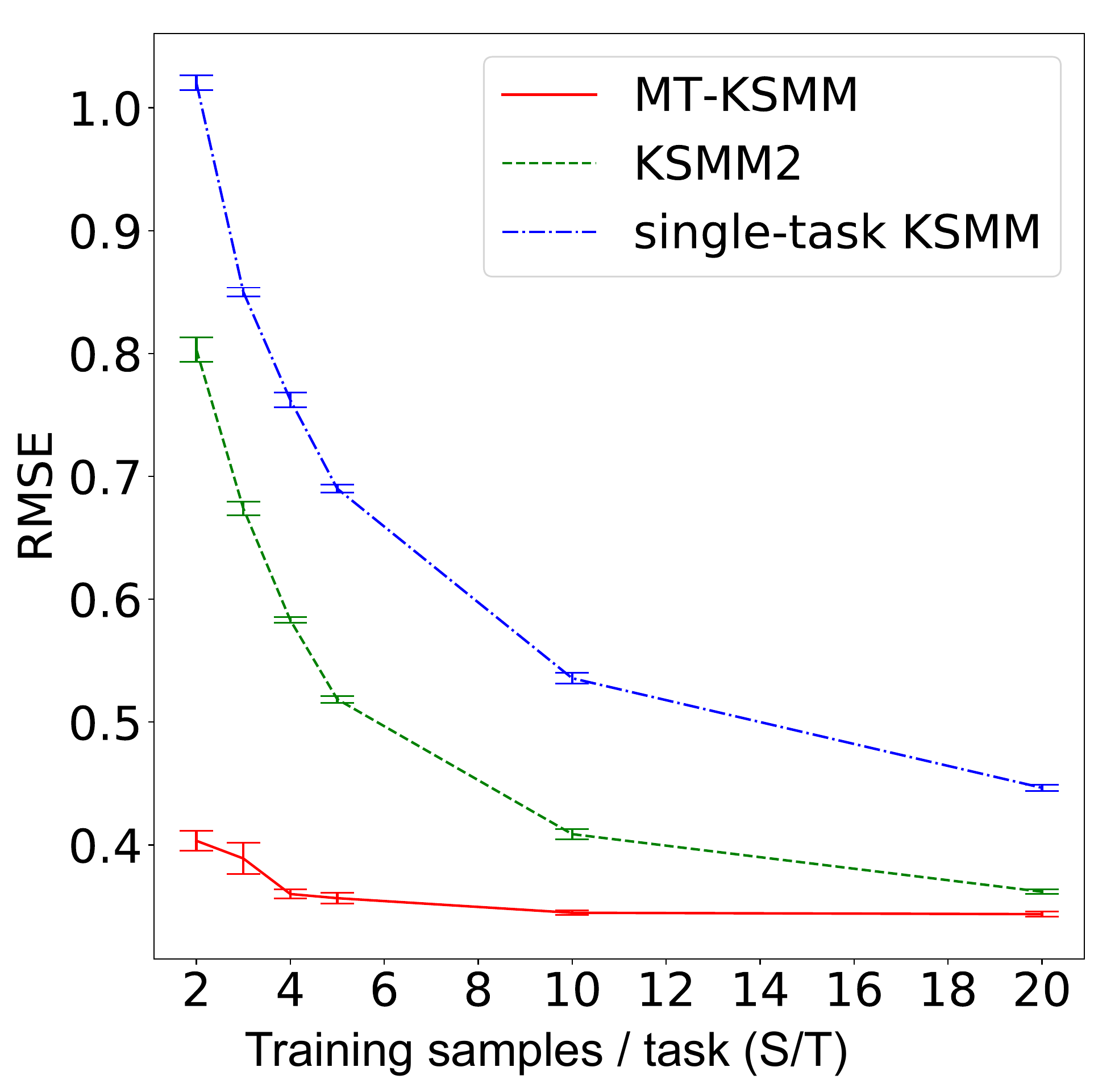} &
\includegraphics[scale=0.19]{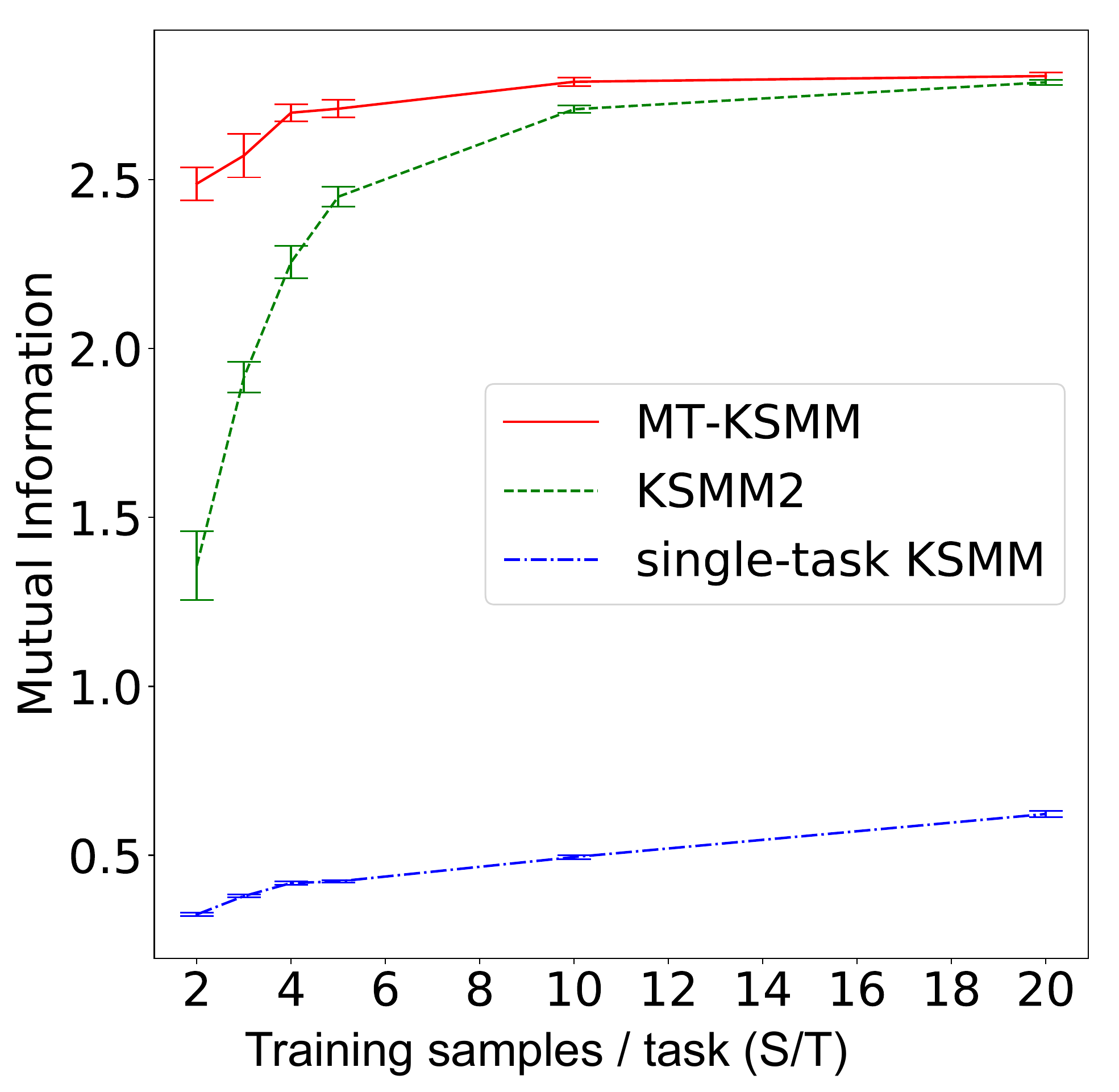} \\
(a) & (b) \\
\includegraphics[scale=0.19]{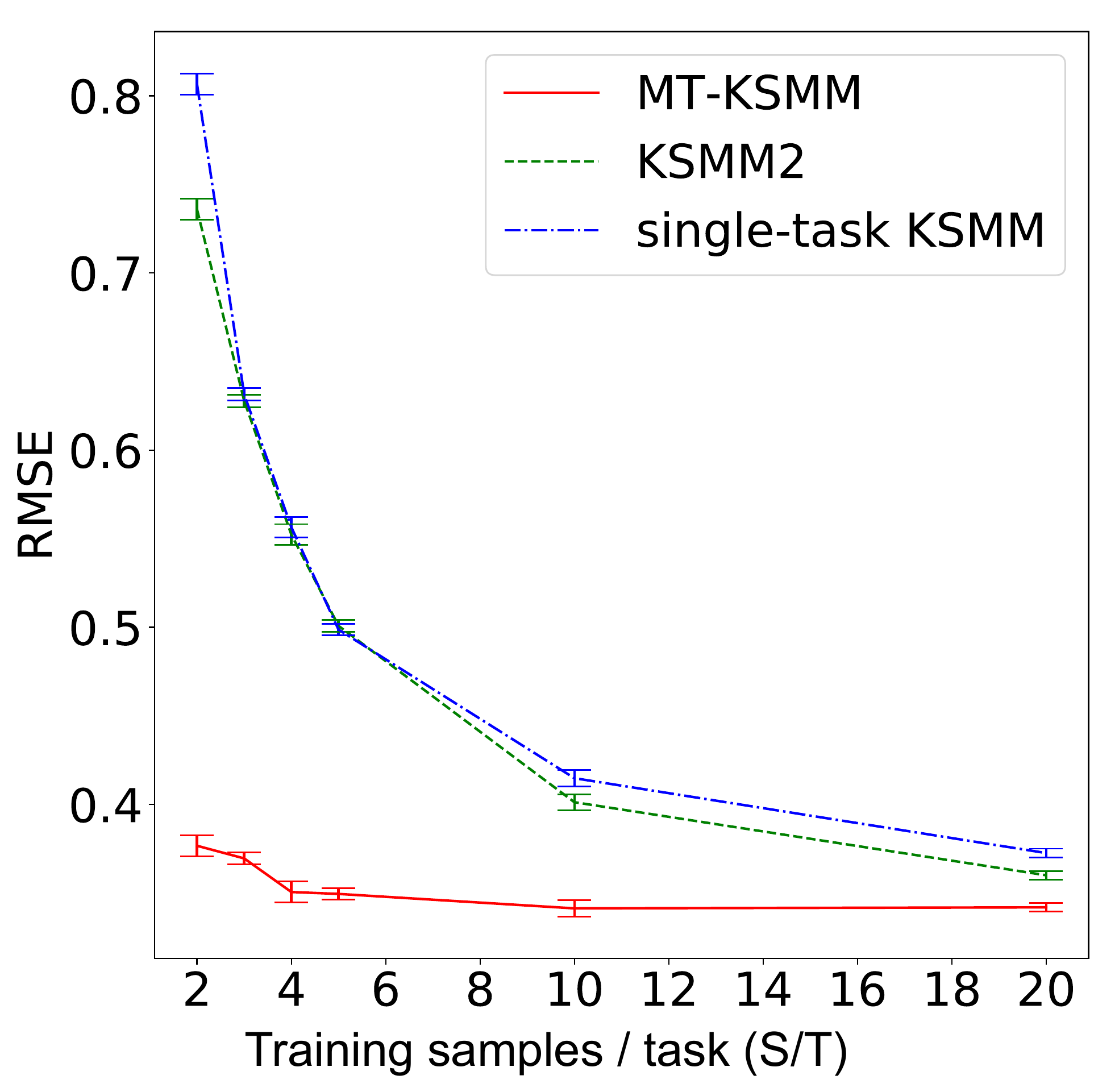} &
\includegraphics[scale=0.19]{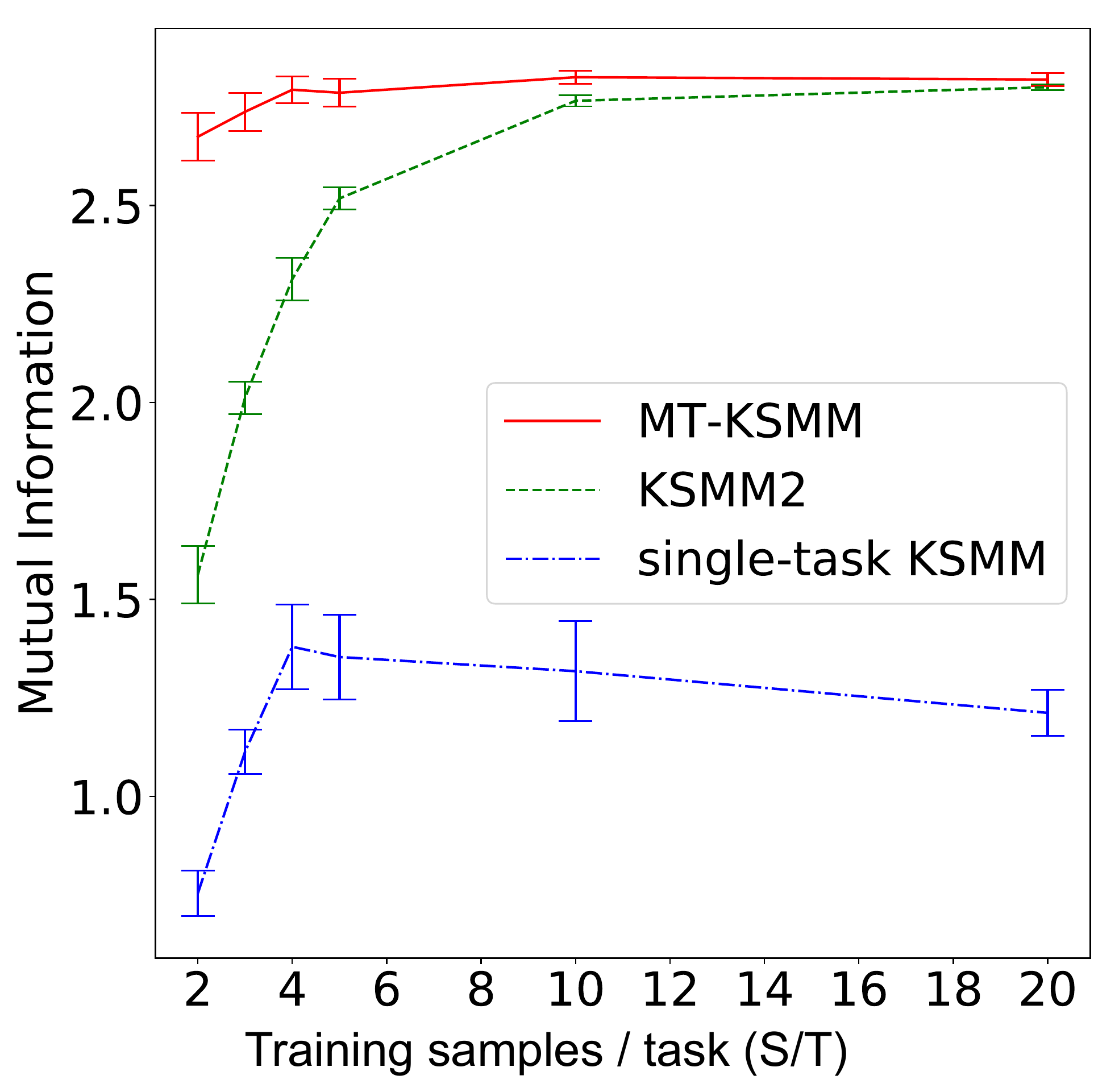} \\
(c) & (d)
\end{tabular}
\caption{Quantitative evaluation when training samples per task (S/T) was changed. The root mean square error (RMSE, smaller is better) and mutual information (MI, larger is better) were evaluated for the test data of existing tasks and for new generated tasks. Error bars indicate the standard deviations. (a) RMSE for existing tasks. (b) MI for existing tasks. (c) RMSE for new tasks. (d) MI for new tasks.}
\figlabel{ex1-2}
\end{figure}

\subsection{Artificial datasets}

We examined the performance of the proposed method using an artificial dataset. We used 2D saddle shape hyperboloid manifolds embedded in 10-dimensional (10D) visible space. The data generation model is given by
\begin{align}
  \vx = F(\vz,u)+\vepsilon &=
  \begin{pmatrix}
    z_1 \\
    z_2 \\
    z_1^2-z_2^2+u\\
    \v0_7
  \end{pmatrix}+\vepsilon,
  \eqlabel{hyperboloid}
\end{align}
where $\vz=(z_1,z_2)\in[-1,+1]^2$ and $u\in[-1,+1]$ are the latent variables generated by the uniform distribution and $\v0_7$ is a seven-dimensional zero vector. $\vepsilon$ is 10D Gaussian noise $\vepsilon\sim \cN(\v0_{10},\sigma^2\vI_{10})$, where $\sigma=0.1$. We generated 300 manifolds (i.e., tasks) by changing $u$, each of which had 100 samples. For training, 200 out of 300 tasks were used for training, each of which had $J$ S/T, and the other samples were used as test data. Thus, samples of 100 tasks were used as test data for new tasks.

A representative result is shown in \figref{ex1-1}. In this case, each task had only two samples for training (2 S/T). Thus, it was impossible to estimate the 2D manifold shape using single-task learning (\figref{ex1-1} (d)). Despite this, the proposed algorithm estimated manifold shapes successfully (\figref{ex1-1} (b)), although the border area of the manifolds was shrunk because of missing data. Compared with the proposed method, \KSMM2 performed poorly in capturing the manifold shapes (\figref{ex1-1} (c)).

\begin{figure*}[p]
\centering
  \includegraphics[width=0.9\textwidth]{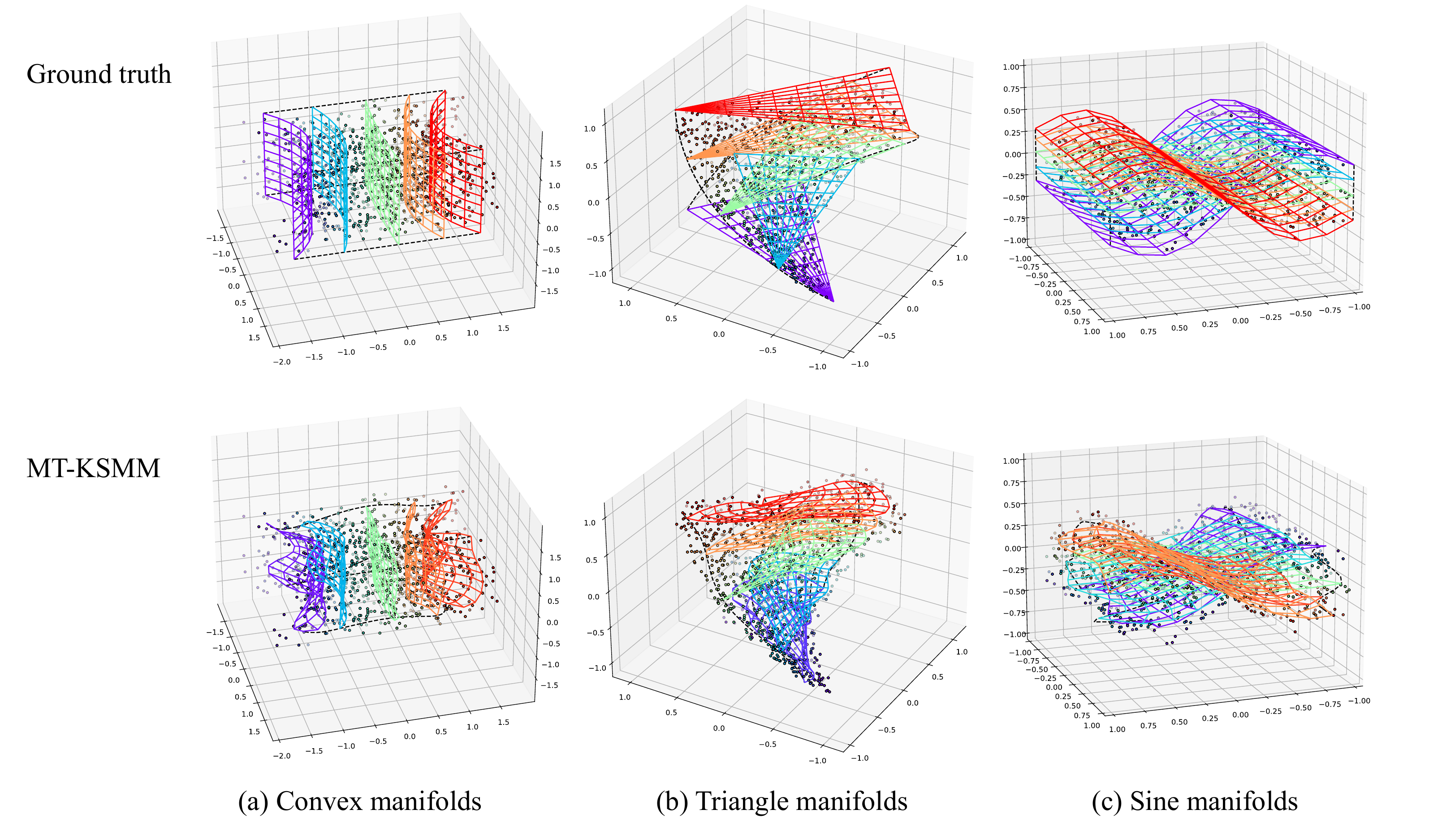}
  \caption{Results for three types of manifold datasets: 400 tasks were used for training, each of which consisted of 3 S/T{}.}
  \figlabel{ex1-3}
\end{figure*}

\begin{table*}[p]\small
  \centering
  \caption{Generalization abilities evaluated using RMSE and MI for four types of manifold datasets: 400 training tasks were used, each of which consisted of 3 S/T. RMSE and MI were evaluated using the test samples of the existing tasks and new tasks.}
  \tbllabel{results}
  \vspace{2mm}
  \begin{tabular}{ccccccc}
    \toprule
      & & & Saddle shape & Convex & Triangle & Sine \\
      \midrule
      \multirow{6}{*}{RMSE}
      & & MT-KSMM & $\mathbf{0.368\pm0.004}$
                & $\mathbf{0.346\pm0.004}$
                & $\mathbf{0.315\pm0.001}$
                & $\mathbf{0.327\pm0.003}$ \\
      & Existing tasks
      & \KSMM2  & $0.679\pm0.005$
                & $0.544\pm0.004$
                & $0.419\pm0.004$
                & $0.591\pm0.005$ \\

      & & KSMM  & $0.854\pm0.010$
                & $0.750\pm0.010$
                & $0.546\pm0.005$
                & $0.742\pm0.009$ \\
    \cmidrule{2-7}
    & & MT-KSMM & $\mathbf{0.353\pm0.003}$
                & $\mathbf{0.339\pm0.004}$
                & $\mathbf{0.309\pm0.002}$
                & $\mathbf{0.319\pm0.002}$ \\
    & New tasks
    & \KSMM2    & $0.633\pm0.006$
                & $0.534\pm0.003$
                & $0.409\pm0.005$
                & $0.551\pm0.007$ \\
    & & KSMM    & $0.619\pm0.009$
                & $0.492\pm0.009$
                & $0.374\pm0.004$
                & $0.432\pm0.019$ \\
    \midrule
      \multirow{6}{*}{MI}
      & & MT-KSMM & $\mathbf{2.662\pm0.012}$
                & $\mathbf{2.670\pm0.014}$
                & $\mathbf{1.467\pm0.066}$
                & $\mathbf{2.706\pm0.013}$ \\
      & Existing tasks
      & \KSMM2  & $1.895\pm0.072$
                & $2.152\pm0.042$
                & $1.458\pm0.050$
                & $1.631\pm0.056$ \\
      & & KSMM  & $0.207\pm0.004$
                & $0.207\pm0.004$
                & $0.205\pm0.005$
                & $0.210\pm0.004$ \\
    \cmidrule{2-7}
    & & MT-KSMM & $\mathbf{2.812\pm0.020}$
                & $\mathbf{2.724\pm0.023}$
                & $\mathbf{1.598\pm0.049}$
                & $\mathbf{2.768\pm0.019}$ \\
    & New tasks
    & \KSMM2    & $2.025\pm0.079$
                & $2.133\pm0.053$
                & $1.529\pm0.022$
                & $1.977\pm0.163$ \\
    & & KSMM    & $1.036\pm0.019$
                & $1.253\pm0.084$
                & $0.937\pm0.054$
                & $1.717\pm0.181$ \\
    \bottomrule
  \end{tabular}
\end{table*}

\figref{ex1-2}~(a) and (b) show the RMSE and MI for the test data of existing tasks when S/T was changed. The results show that MT-KSMM performed better than the other methods, particularly when S/T was small. As S/T increased, all methods reduced the RMSE{}. By contrast, in the case of MI, the performance of single-task KSMM did not improve, even when a sufficient number of training samples was provided. Because no information was transferred between tasks in single-task KSMM, the manifolds were estimated independently for each task. Consequently, the sample latent variables $\vz$ were estimated inconsistently between tasks.

\figref{ex1-2}~(c) and (d) show the RMSE and MI for the test data for new tasks. We generated 100 new tasks using \eqref{hyperboloid}, each of which had 100 samples. Then we estimated the task latent variable $u$ and sample latent variable $\vz$ using the trained model. For new tasks, MT-KSMM also demonstrated excellent performance, which suggests that it has high generalization ability.

We further evaluated the performance using other artificial datasets, which are shown in \figref{ex1-3}. In this experiment, we used 400 tasks and 3 S/T for training. As shown in the figure, MT-KSMM succeeded in modeling the continuous change of manifold shapes. In the case of the triangular manifold (b), MT-KSMM succeeded in capturing the rotation of triangular shapes. In the case of sine manifolds, all manifolds intersected on the same line, and the data points on the intersection could not be distinguished between tasks. Despite this, MT-KSMM succeeded in modeling the continuous change of the manifold shape. \tblref{results} summarizes the experiment on artificial datasets. For all cases, MT-KSMM consistently demonstrated the best performance among the three methods.

\begin{figure*}[t]
  \centering
  \includegraphics[width=\textwidth]{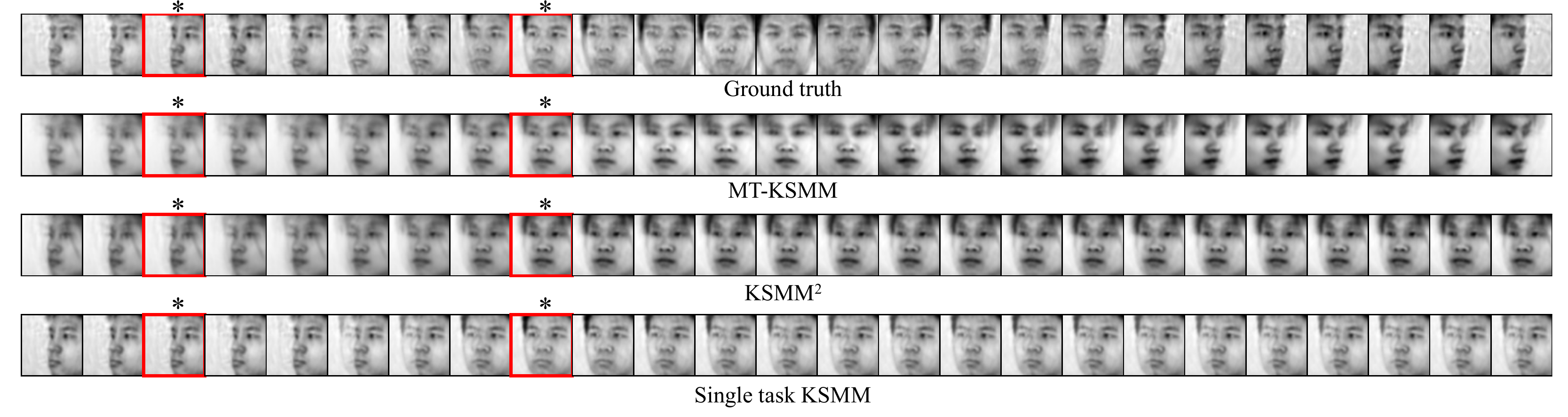}\\(a)\\
  \includegraphics[width=\textwidth]{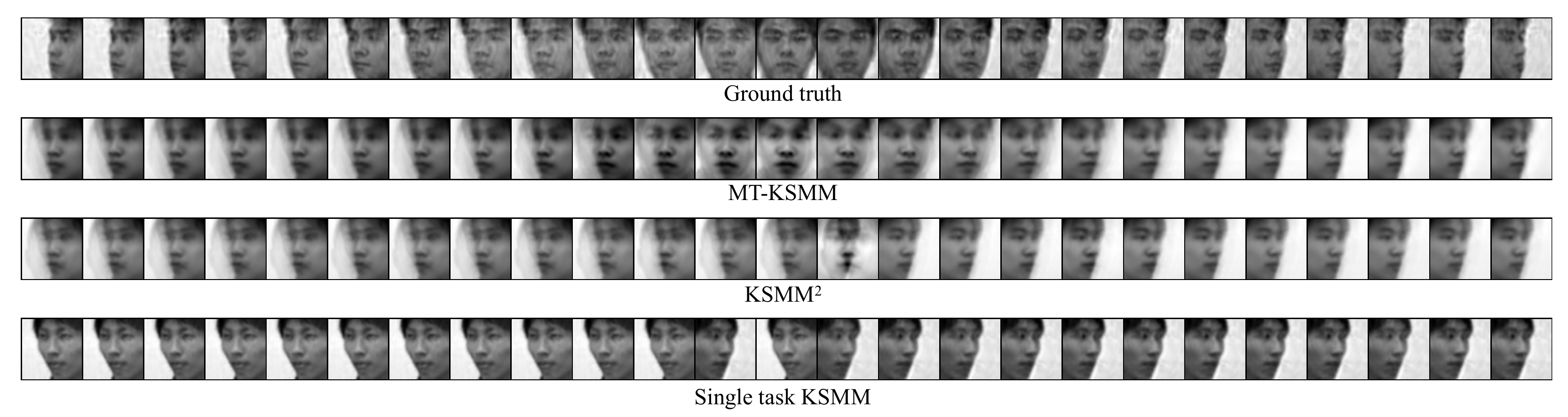}\\(b)
  \caption{Results for the face image dataset A, when 2 S/T was used for training.  Original and reconstructed face images with various poses for (a) an existing task and (b) a new task. The images indicated by asterisks (*) were used for training and the remainder were used for testing.}
  \figlabel{ex2-1}
\end{figure*}

\begin{figure}[t]
  \centering
  \includegraphics[width=0.9\columnwidth]{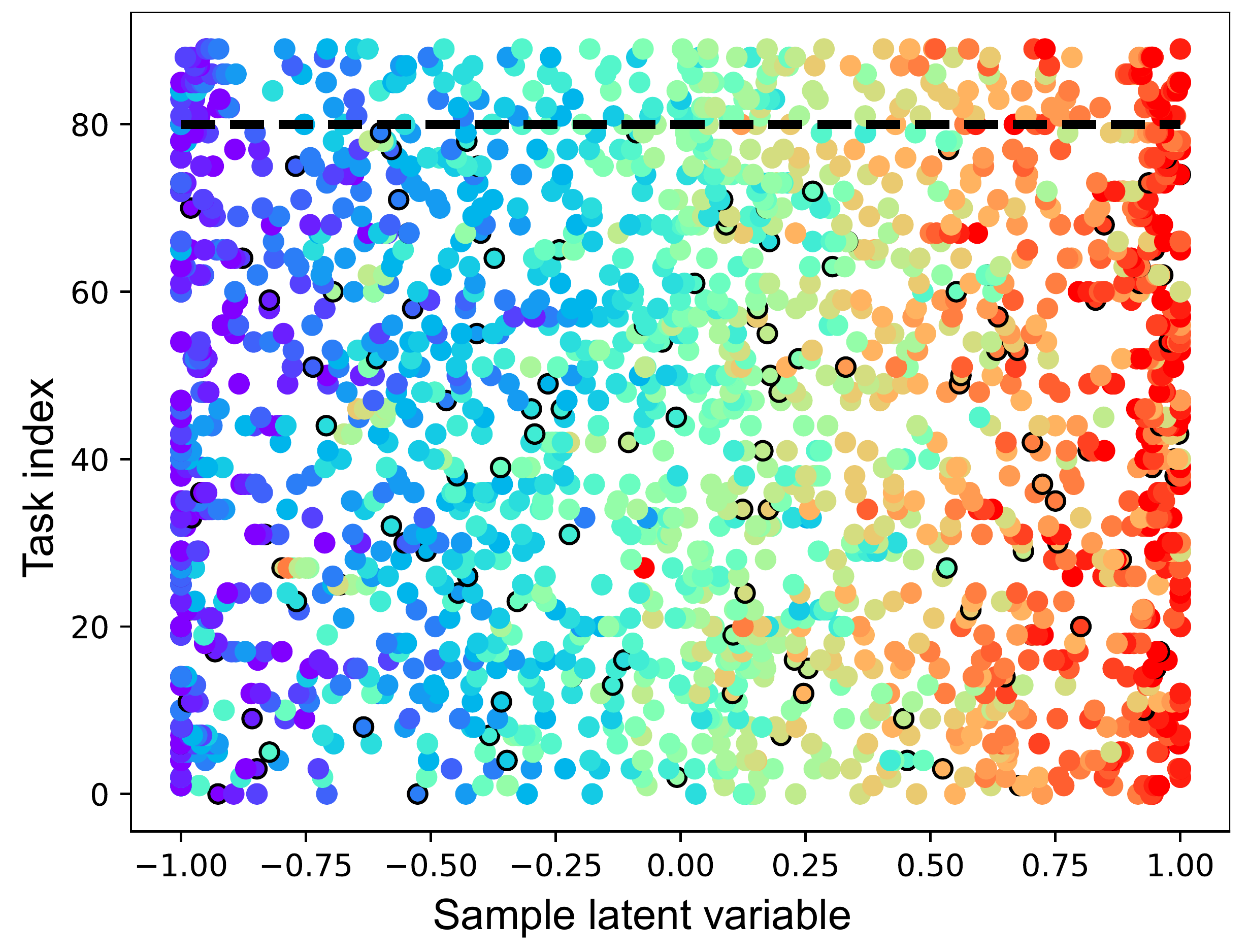}
  \caption{Sample latent variables of the face image dataset A estimated using MT-KSMM{}. MT-KSMM was trained using 2 S/T{}. Colors indicate the true angles of poses (blue: $-60^\circ$, red: $+60^\circ$). Tasks 0--79 (below the dashed line) were used for training and tasks 80--89 (above the dashed line) were used for testing as the new task. The markers outlined in black represent the training data.}
  \figlabel{ex2-2}
\end{figure}
\begin{figure}[t]
  \centering
  \includegraphics[width=0.9\columnwidth]{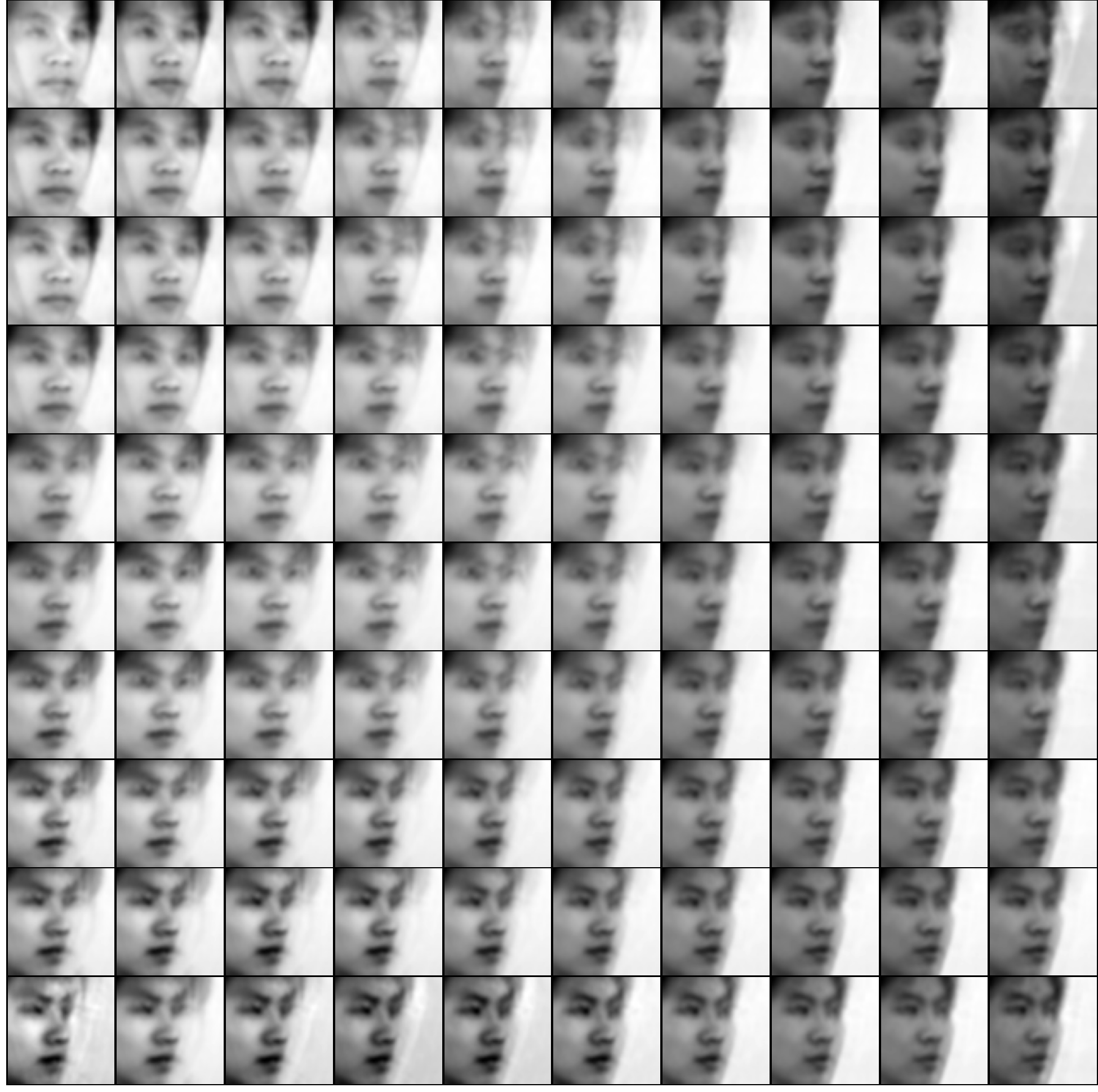}
  \caption{Face images generated by MT-KSMM by changing the task latent variable $\vu$, while the sample latent variable $\vz$ was fixed at $0.9$. Two S/T were used for training.}
  \figlabel{ex2-3}
\end{figure}
\begin{figure}[t]
  \centering\small
  \begin{tabular}{cc}
    \includegraphics[scale=0.2]{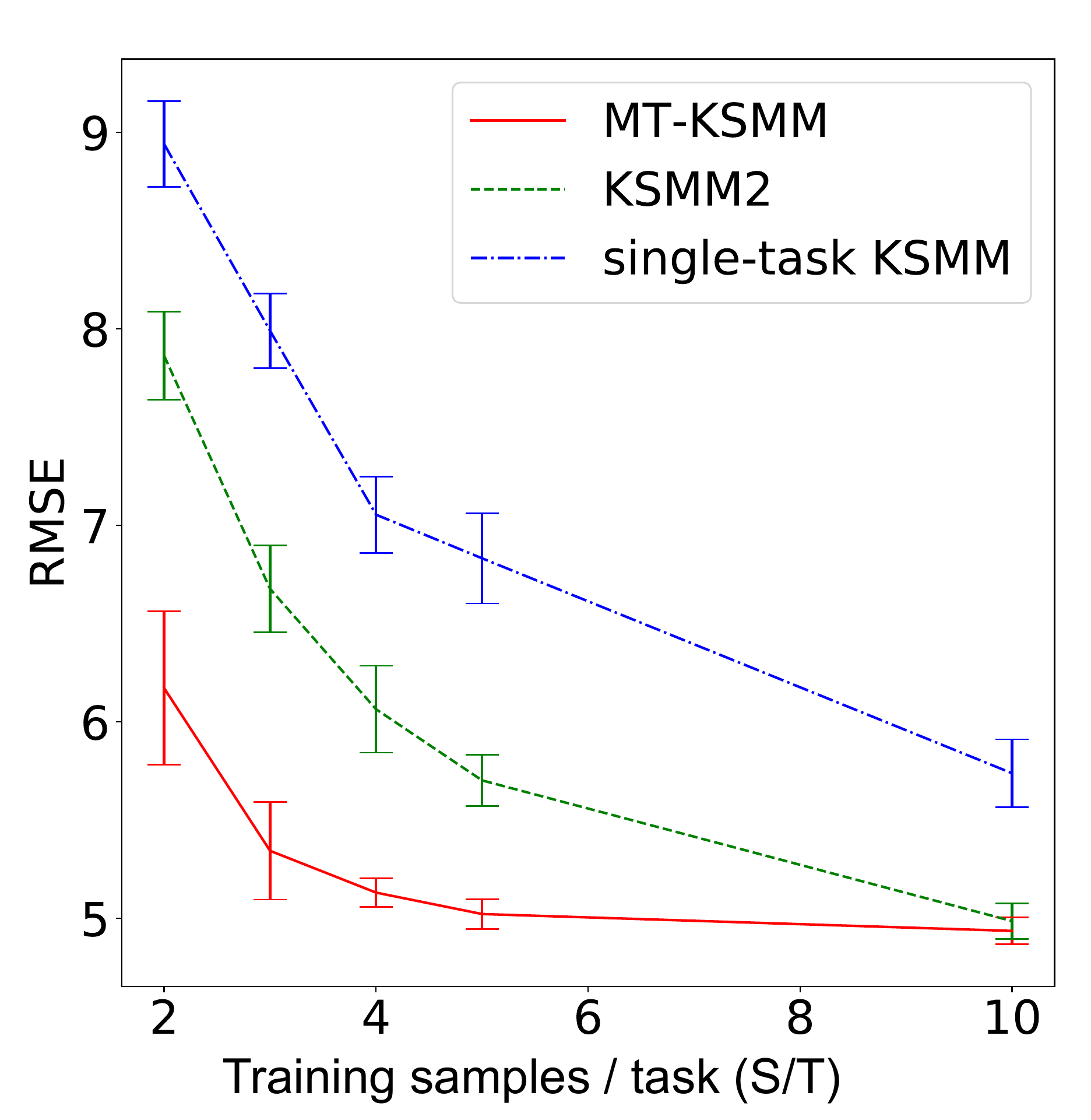} &
    \includegraphics[scale=0.2]{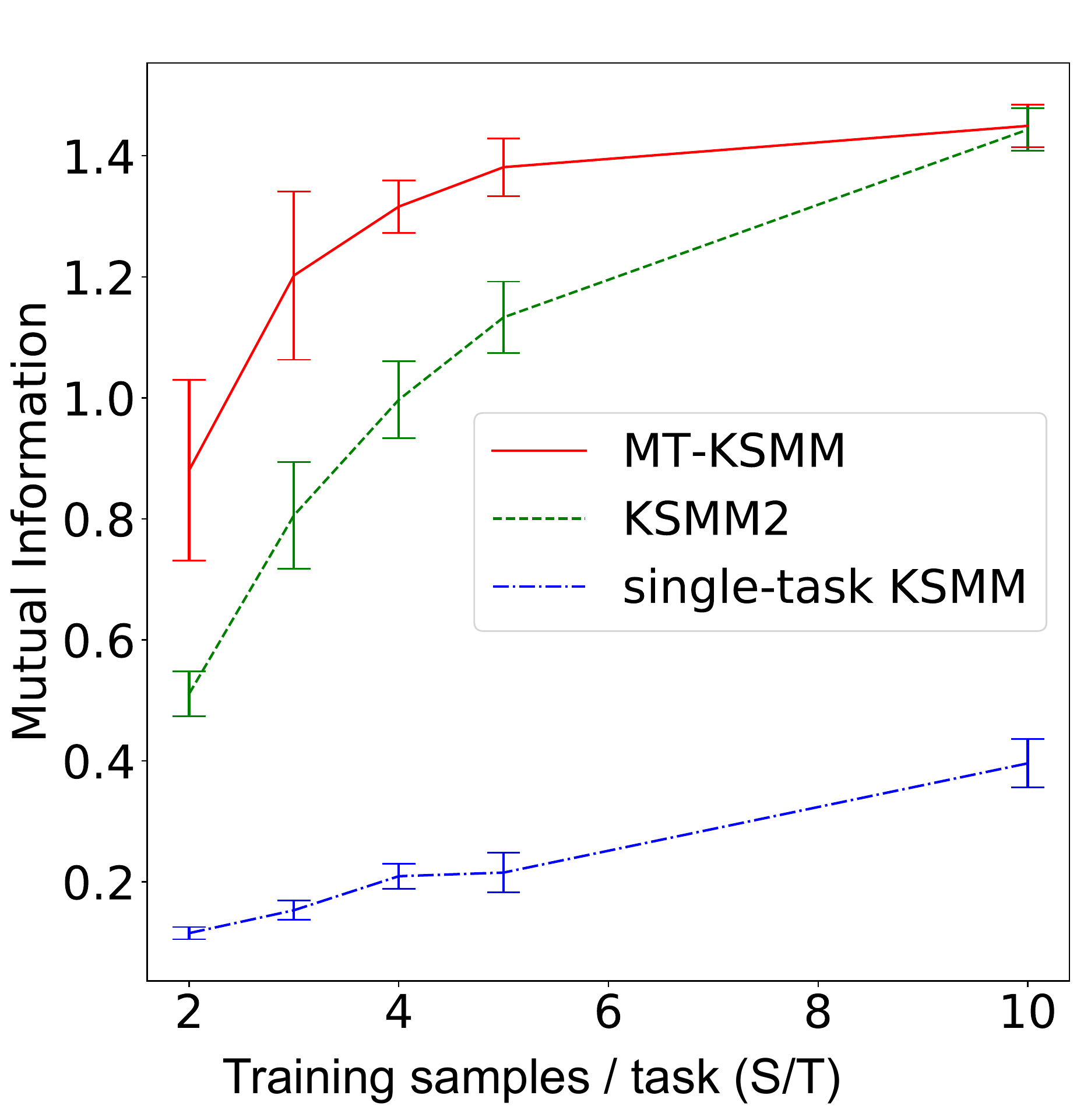} \\
    (a) & (b) \\
    \includegraphics[scale=0.2]{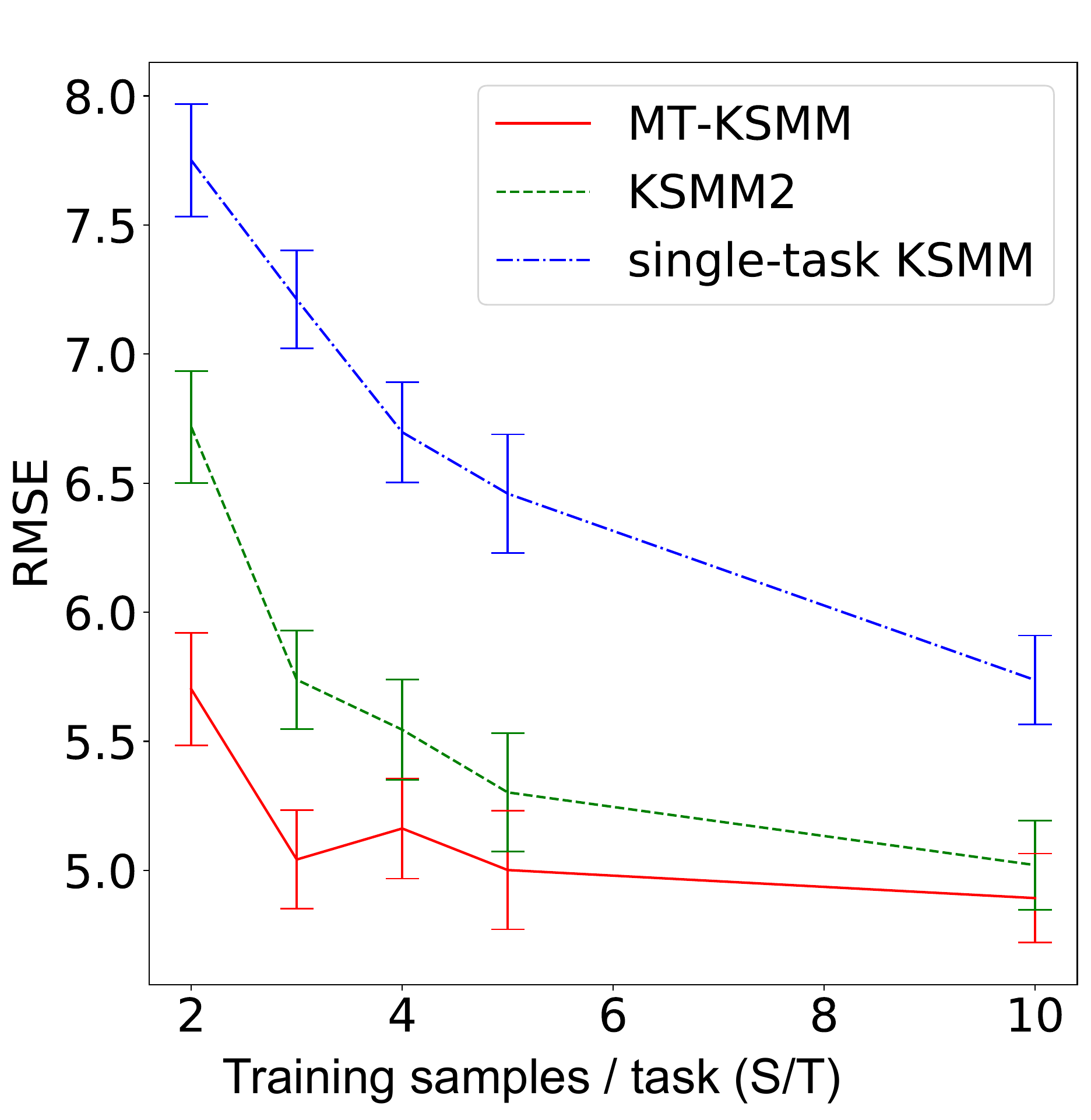} &
    \includegraphics[scale=0.2]{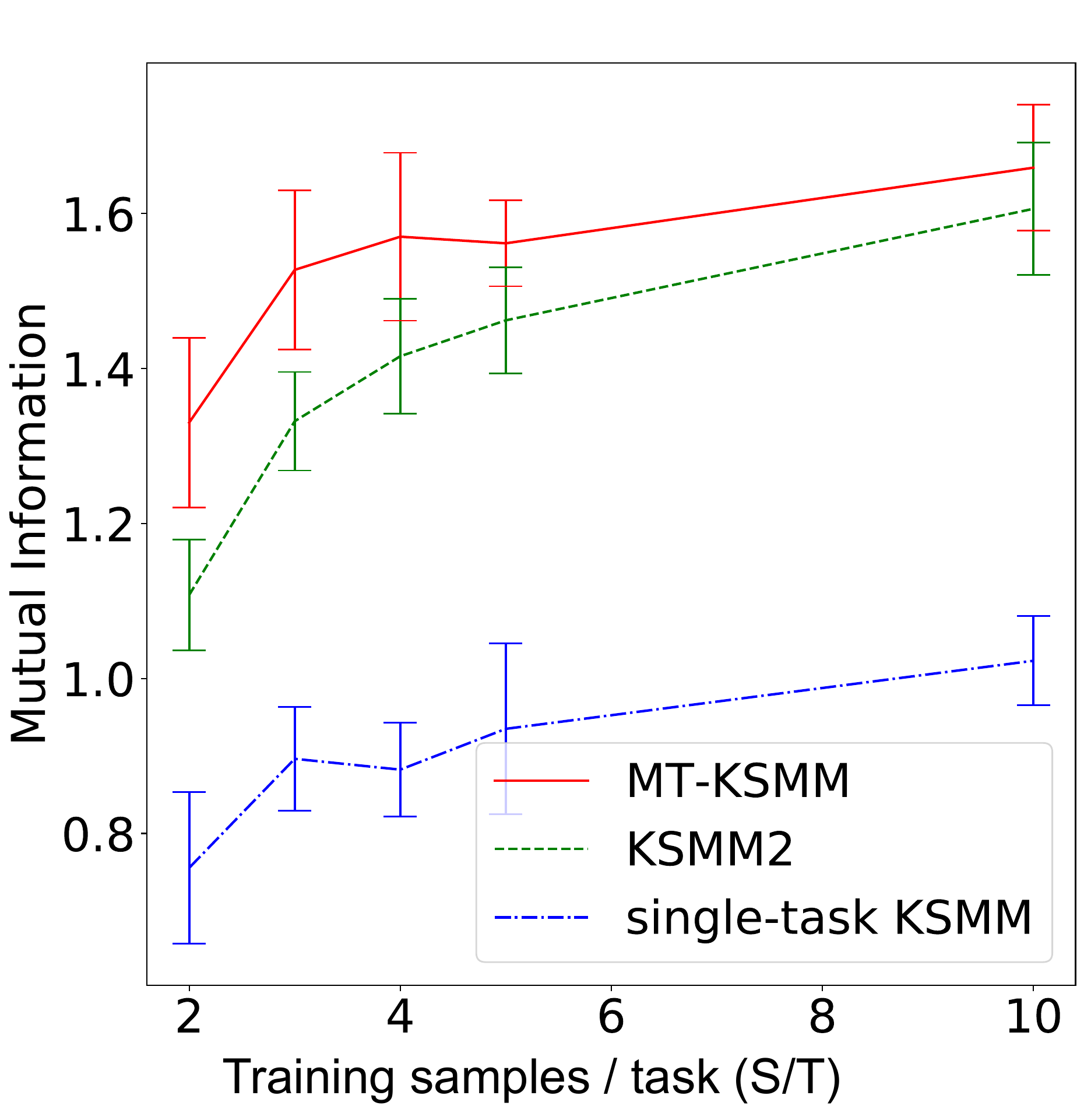} \\
    (c) & (d)
  \end{tabular}
  \caption{Quantitative evaluation for the face image set A, when the training sample number per task was changed. RMSE of the reconstructed images, MI between the pose angles, and the sample latent variables were evaluated for new images of existing subjects that appeared in training, and for new images of new subjects. (a) RMSE for existing subjects. (b) MI for existing subjects. (c) RMSE for new subjects. (d) MI for new subjects.}
\figlabel{ex2-4}
\end{figure}

\begin{figure*}[tp]
  \centering \small
  \includegraphics[width=\textwidth]{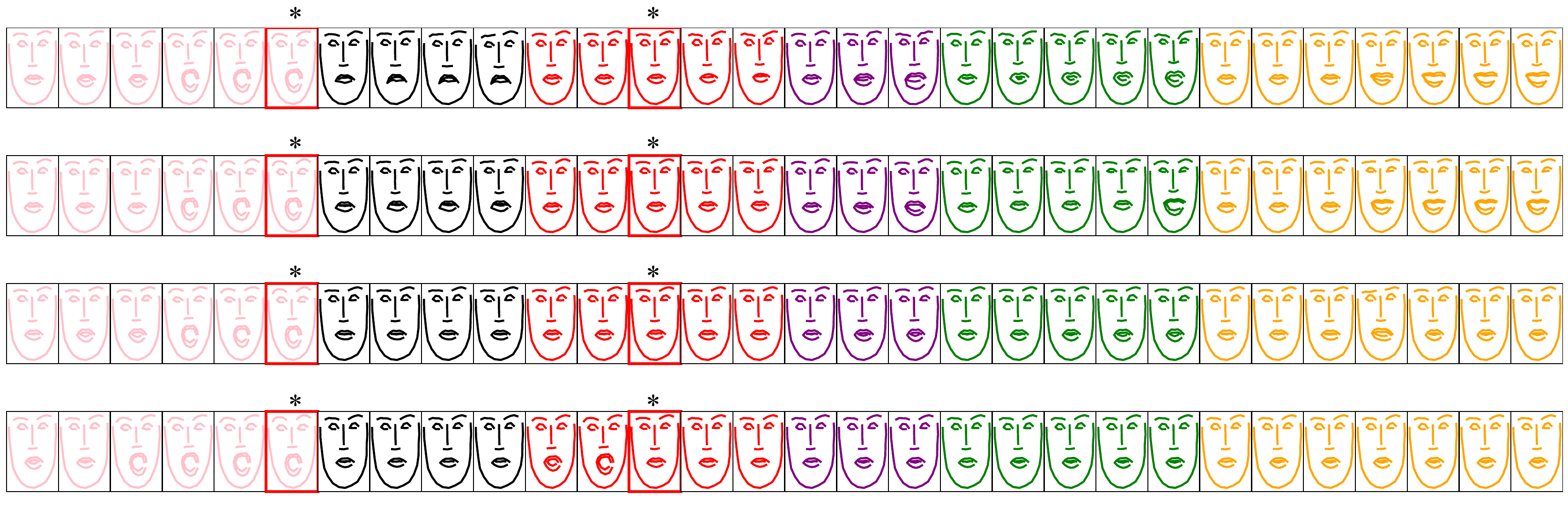}
  \caption{Reconstructed face images for dataset B{}, in which each face image is represented by a set of 64 landmark points. In this experiment, 2 S/T were used for training (indicated by an asterisk (*)). (From top to bottom) ground truth, and images reconstructed by MT-KSMM, \KSMM2, and single-task KSMM{}. The color represents the type of emotion: pink: surprise, red: anger, purple: fear, green: disgust, orange: happiness, black: label missing.}
  \figlabel{ex3-1}
  \vspace{3mm}
  \begin{tabular}{cc}
    \includegraphics[scale=0.35]{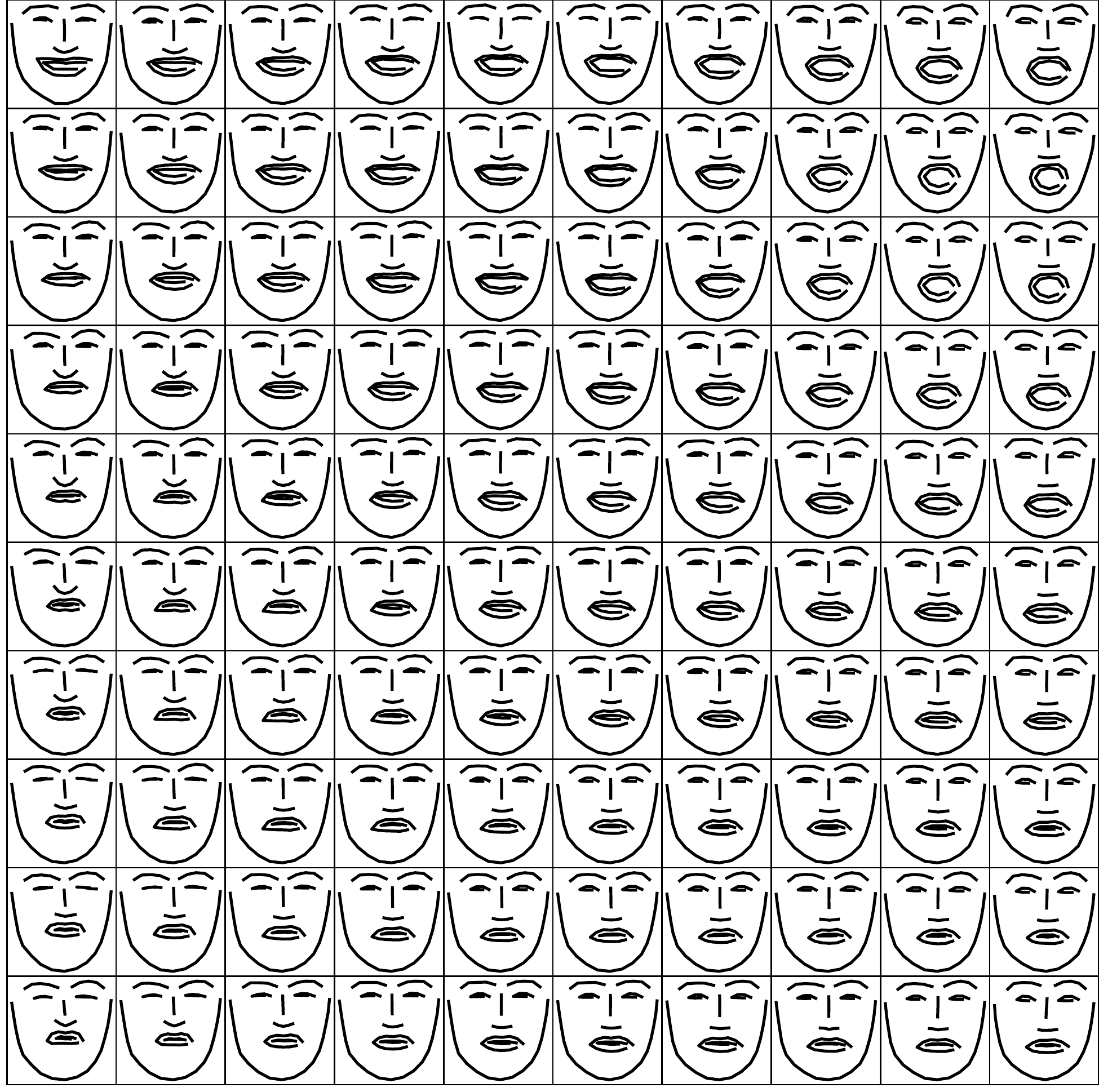}&
    \includegraphics[scale=0.35]{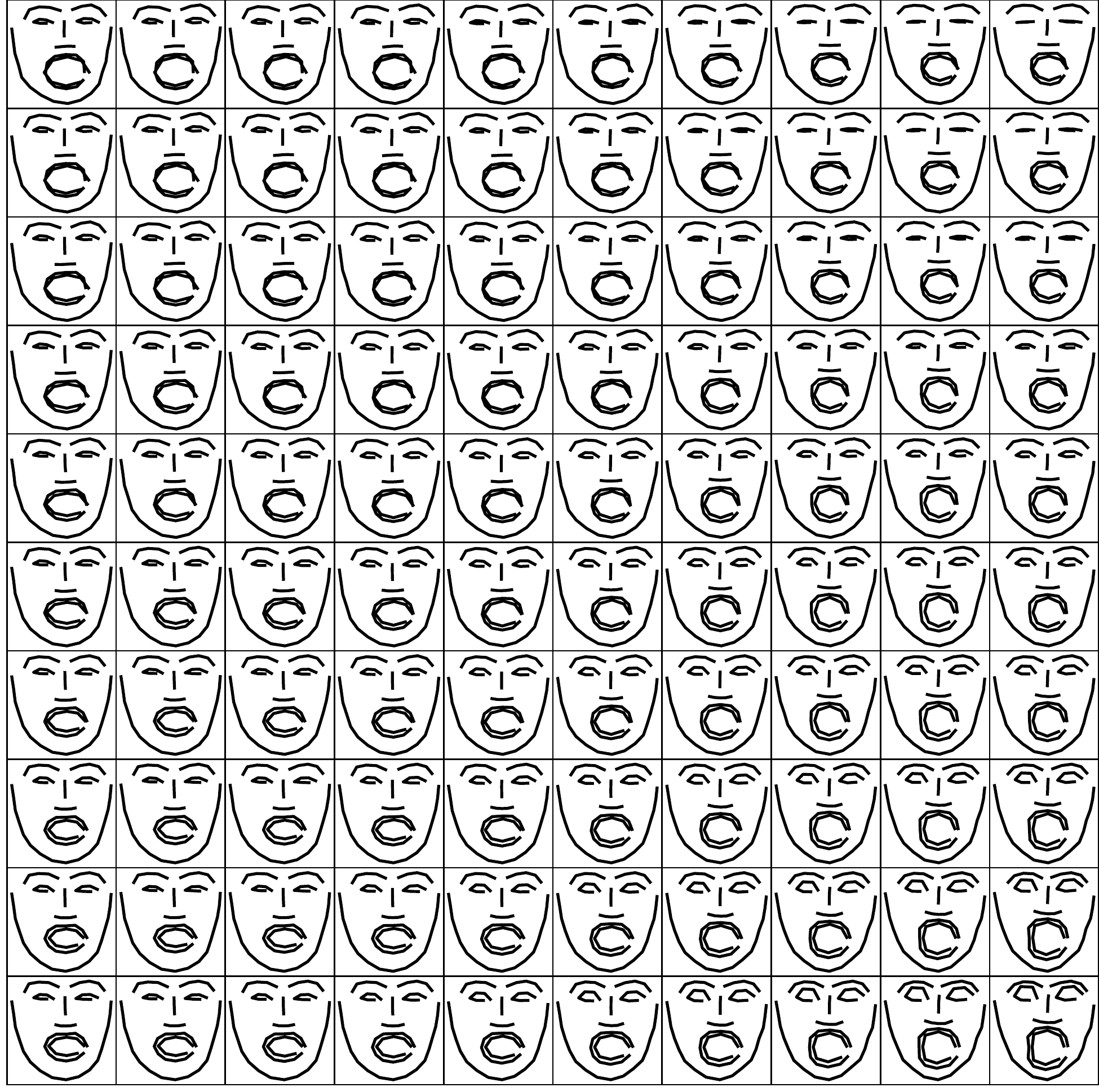}\\
    (a) & (b)
  \end{tabular}
  \caption{Face images generated by MT-KSMM, which was trained using 2 S/T. (a) Various expressions of a subject generated by changing the sample latent variable $\vz$. (b) Various styles of `surprise' generated by changing the task latent variable $\vu$.}
  \figlabel{ex3-2}
  \vspace{3mm}
  \begin{tabular}{cccc}\small
    \includegraphics[scale=0.33]{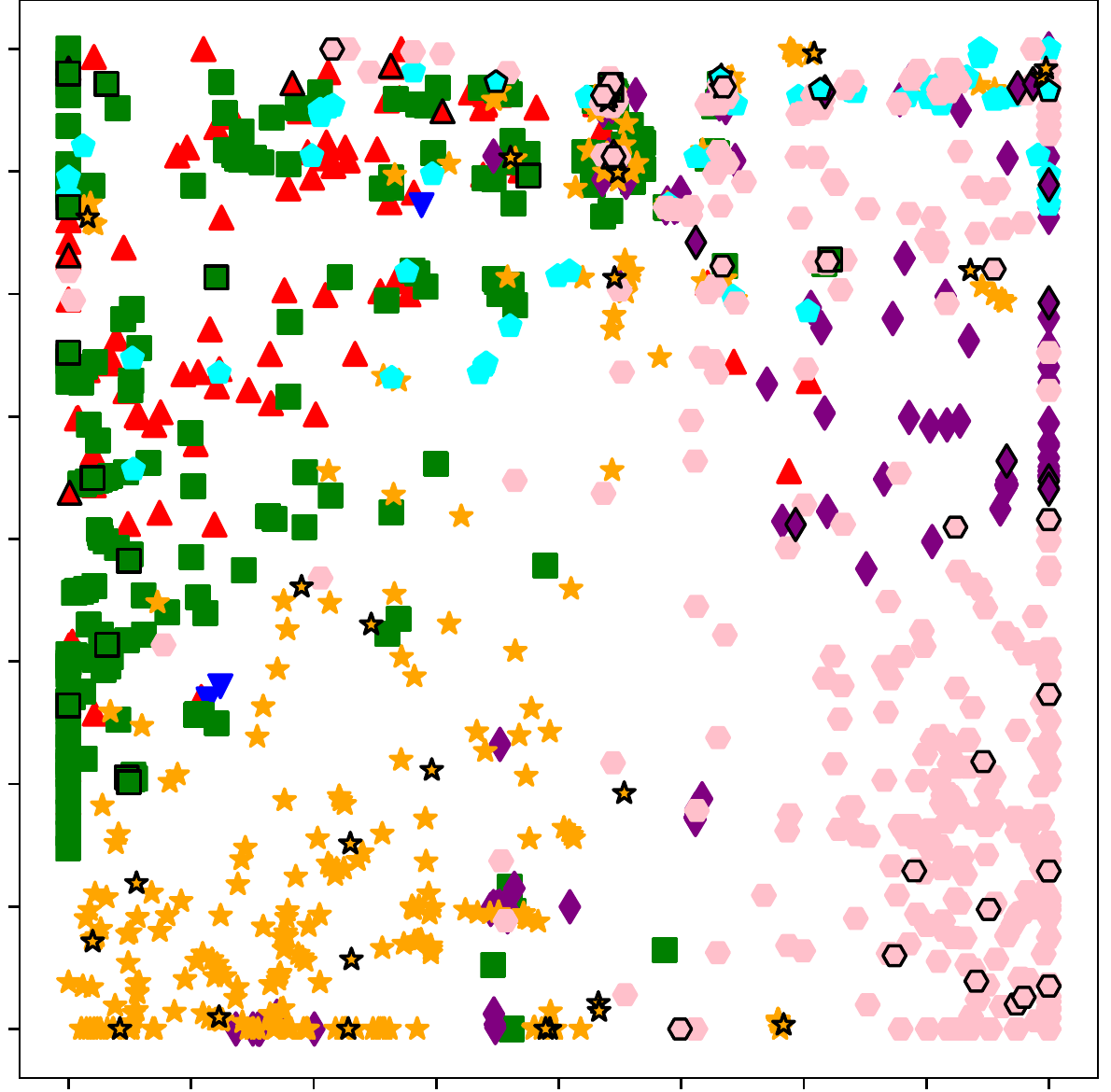} &
    \includegraphics[scale=0.33]{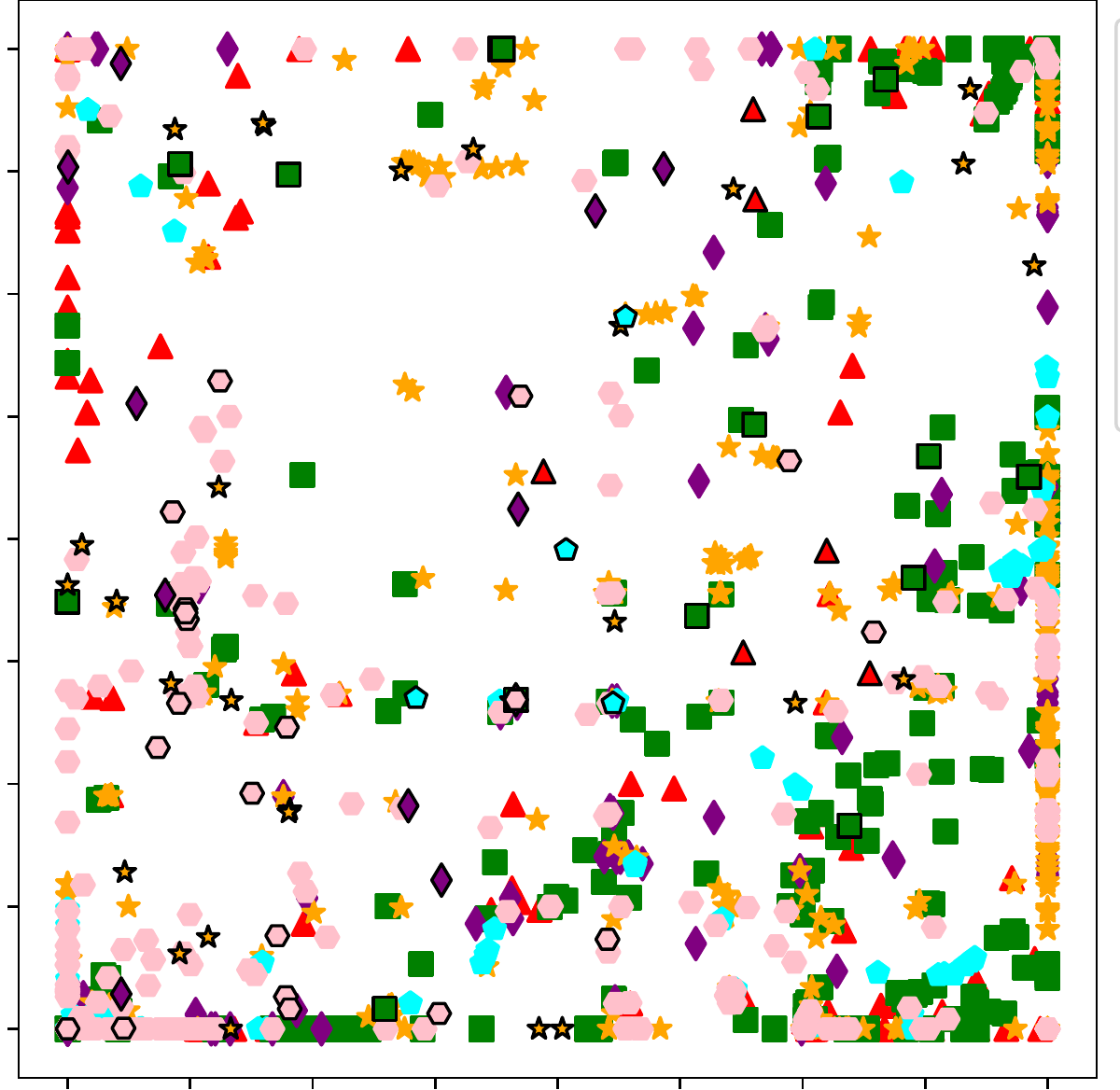} &
    \includegraphics[scale=0.33]{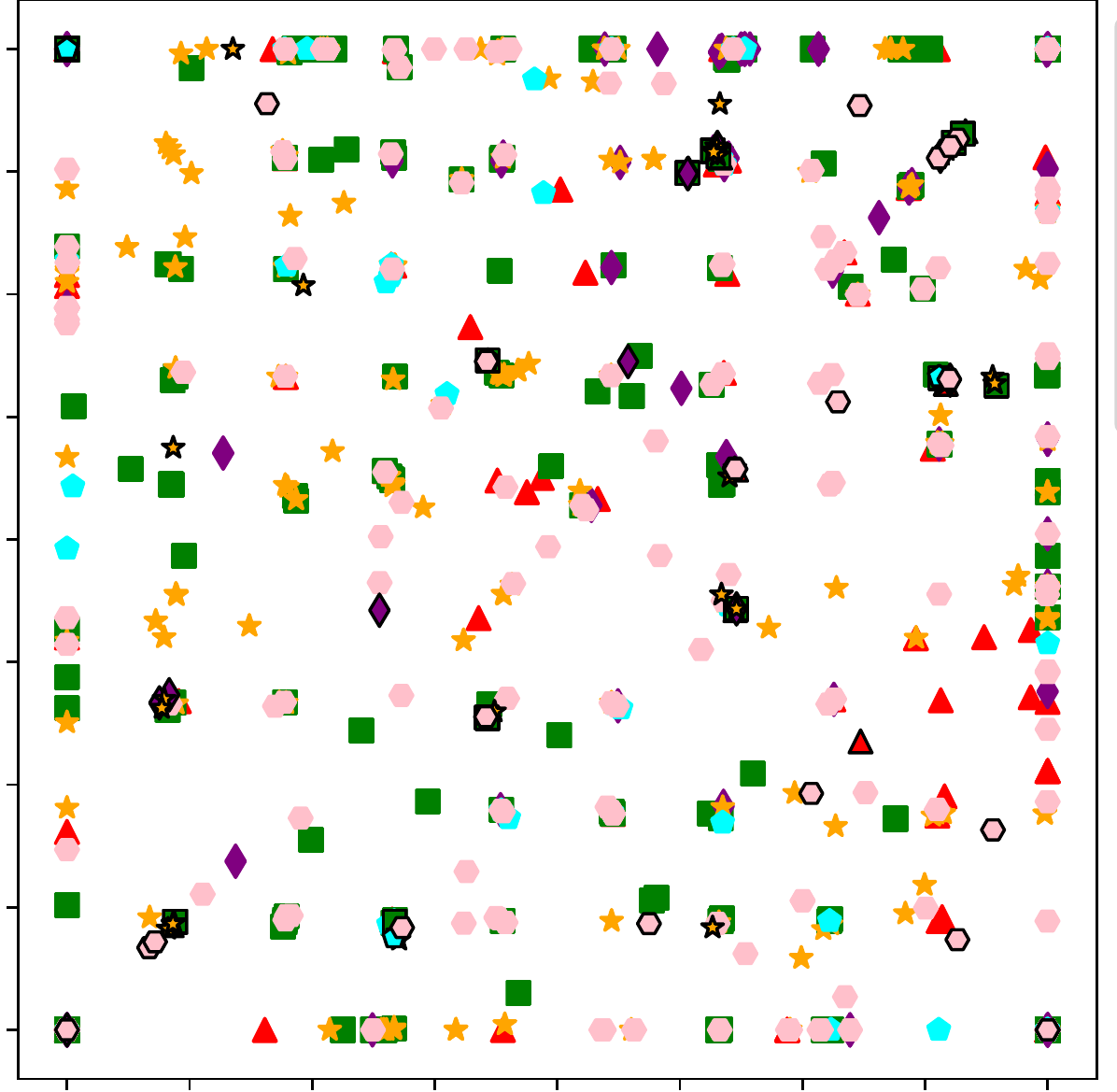} &
    \includegraphics[scale=0.5]{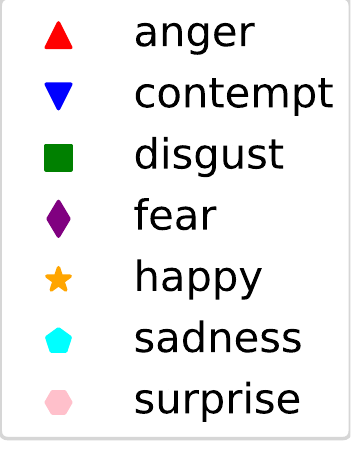} \rule{20mm}{0mm}\\
    (a) & (b) & (c) \\[2mm]
  \end{tabular}
  \caption{Estimated sample latent variables. The markers outlined in black are the training data. (a) MT-KSMM. (b) \KSMM2. (c) Single-task KSMM.}
   \figlabel{ex3-3}
\end{figure*}

\subsection{Face image dataset A: various poses}

We applied the proposed method to a face image dataset with various poses\footnote{\url{http://robotics.csie.ncku.edu.tw/Databases/FaceDetect_PoseEstimate.htm}}. The dataset consisted of the face images of 90 subjects taken from various angles. We used 25 images per subject, which were taken every $5^\circ$ from $-60^\circ$ to $+60^\circ$. The face parts were cut out from the original images with $64\times64$ pixels and a Gaussian filter was applied to remove noise. In the experiment, we used the face images of 80 subjects for the training tasks and the remaining images of 10 subjects for the test tasks. In this experiment, the dimensions of the sample latent space $\cL$ and the task latent space $\cT$ were 1 and 2, respectively.

\figref{ex2-1} shows the reconstructed images for an existing task (a) and new task (b). In this case, only two images per subject were used for training (i.e., 2 S/T). In the case of the existing task (\figref{ex2-1} (a)), we used only left face images of the subject for training, but MT-KSMM supplemented the right face images by mixing the images of other subjects. As a result, MT-KSMM successfully modeled the continuous pose change\footnote{We make a brief comment about image quality. Because the manifolds were represented as a linear combination of training data, it was unavoidable that the reconstructed images became cross-dissolved images of the training images, particularly when the number of training data was small. To reconstruct more realistic images, it is necessary to use more training images, and/or to implement prior knowledge into the method, but this is out of scope for this study.}. By contrast, \KSMM2 and single-task KSMM failed to model the pose change. This ability of MT-KSMM was also observed for new tasks, for which no image was used for training. As \figref{ex2-1} (b) shows, MT-KSMM reconstructed face images of various angles, whereas \KSMM2 and single-task KSMM failed.

\figref{ex2-2} shows the sample latent variables $\Hat Z=\{\Hat\vz_n\}$ estimated by MT-KSMM{}. The estimated latent variables were consistent with the actual pose angles for both existing tasks (below the dashed line) and new test tasks (above the dashed line). Thus, the face manifolds were aligned for all subjects, so that the latent variable represented the same property regardless of the task difference.

By changing the task latent variable $\vu$ and fixing the sample latent variable $\vz$, various faces in the same pose were expected to appear. \figref{ex2-3} shows the generated face images using MT-KSMM when $\vu$ was changed. The results indicate that MT-KSMM worked as expected. In terms of the fiber bundle, the manifolds of various poses (\figref{ex2-1}) corresponds to the fibers, whereas the manifold that consisted of various subjects corresponds to the base space (\figref{ex2-3}).

\figref{ex2-4} shows the RMSE and MI for the test data for existing tasks ((a) and (b)) and new tasks ((c) and (d)). In this evaluation, RMSE was measured using the error between the true image and reconstructed image, and MI was measured between the true pose angle and sample latent variable $\vz$. Like the case of the artificial dataset, MT-KSMM performed better than the other methods, particularly when the number of training data was small.

\subsection{Face image dataset B: various expressions}

\begin{figure}[t]
    \centering
      \begin{tabular}{cc}\small
      \includegraphics[scale=0.2]{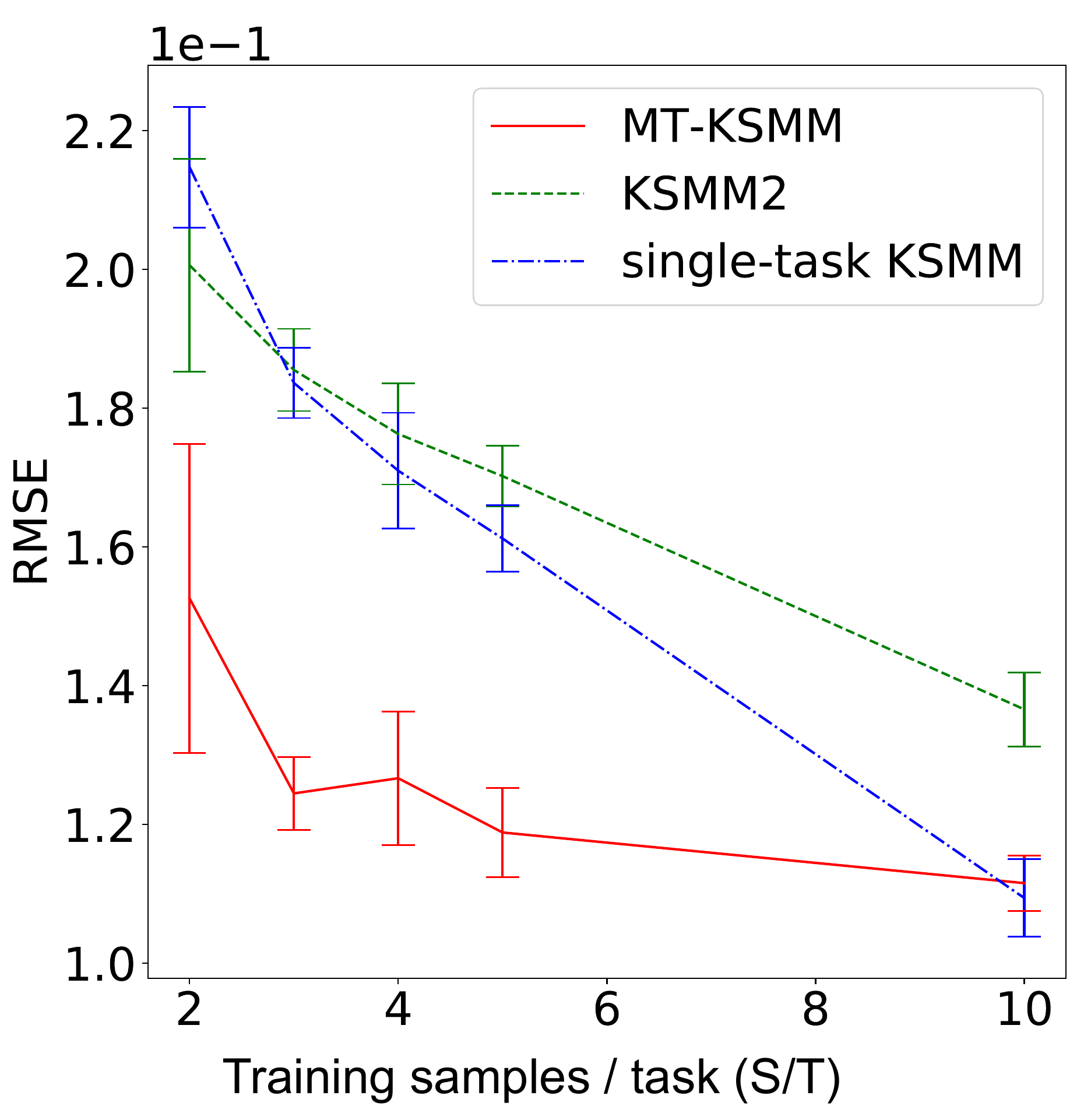} &
      \includegraphics[scale=0.2]{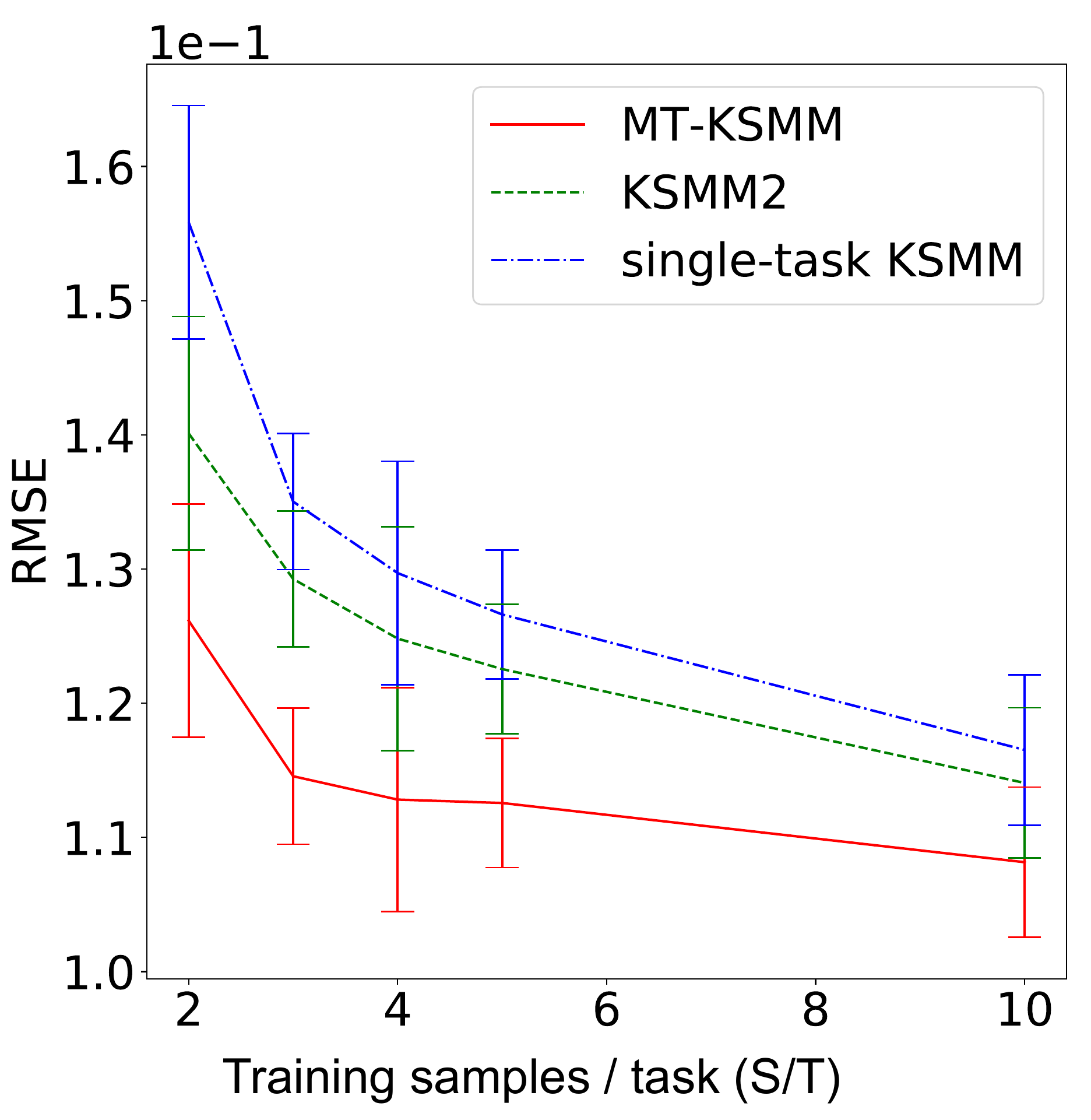}\\
      (a) & (b)
      \end{tabular}
    \caption{RMSE of the reconstructed images when the training samples per task (S/T) were changed. (a) For test data of existing tasks. (b) For test data of new tasks.}
    \figlabel{ex3-4}
\end{figure}

\begin{figure}
  \centering
  \begin{tabular}{cc}\small
    \includegraphics[width=\linewidth]{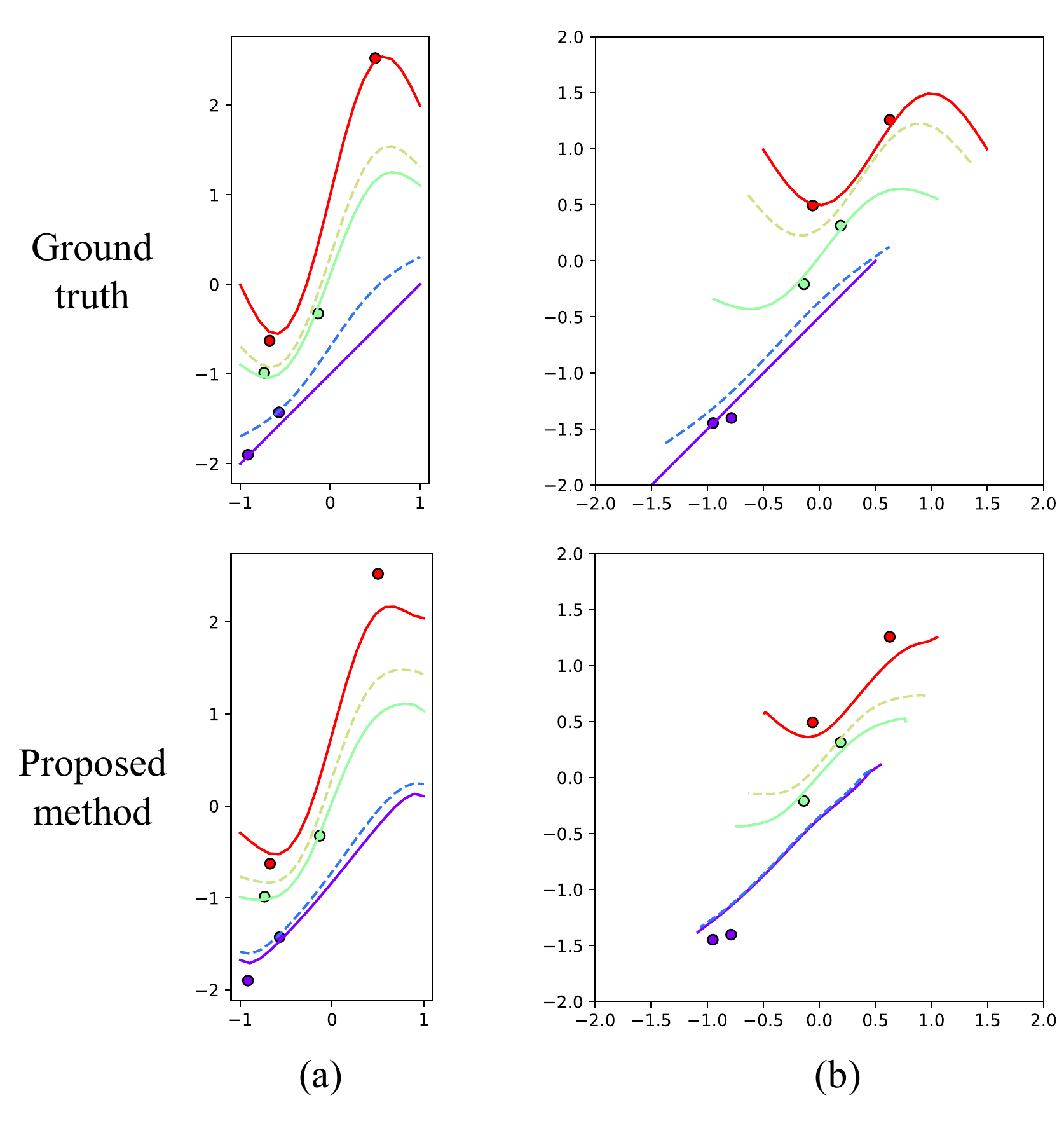} &
  \end{tabular}
  \caption{Application to supervised learning cases. (a) Multi-task regression. (b) Multi-task regression with domain shift. Solid curves are training tasks (3 out of 400 tasks are indicated in the figure), and dashed curves are new tasks. Markers denote the training samples (2 S/T in this experiment).}
  \figlabel{ex4}
\end{figure}

We applied the proposed method to another face image dataset, in which subjects displayed various expressions \cite{Kanade2000,Lucey2010}. The dataset consisted of image sequences, which ranged from a neutral expression to a distinct expression. We chose 96 subjects from the dataset, each of which had 30 images. Eighty subjects were used for training as existing tasks and the remaining 16 subjects are used for testing as new tasks. In the dataset, the sequences were labeled according to the emotion type, but the label information was not used for training.

The face image data that we used were pre-processed as follows: From each face image, 64 landmark points, such as eyes and lip edges, were extracted, and 2D coordinates were obtained from the image. Thus, each face image was transformed into 128-dimensional vector data. To separate the emotional expression from the face shape, the landmark vectors were subtracted from the vector of the neutral expressions. Thus, for all subjects, the origin was the neutral expression. To learn the dataset, the 2D latent space was used for both $\cL$ and $\cT$.

\figref{ex3-1} shows the reconstructed images for an existing task, trained using 2 S/T{}.  Although only two images were used for training (`surprise' and `anger' for this subject), MT-KSMM reconstructed other expressions successfully, whereas \KSMM2 and single-task KSMM failed.

MT-KSMM was expected to generate images expressing various emotions according to the sample latent variable $\vz$, whereas it was expected to generate images of various expression styles of the same emotion type according to the task latent variable $\vu$. \figref{ex3-2} shows the images generated by MT-KSMM, suggesting that the proposed method successfully represented the faces as expected. In terms of the fiber bundle, \figref{ex3-2} (a) represents a fiber, whereas \figref{ex3-2} (b) shows the base space.

\figref{ex3-3} shows the estimated sample latent variables $\Hat Z=\{\Hat \vz_n\}$. The results indicate that MT-KSMM (\figref{ex3-2} (a)) roughly classified the face images according to the expression, whereas the \KSMM2 and single-task KSMM did not demonstrate such a classified representation of emotion types in the latent space. Finally, we evaluated the RMSE of the reconstructed images (\figref{ex3-4}). Again, MT-KSMM performed better than the other methods for both existing and new tasks.

\begin{figure}[p]
    \centering
    \includegraphics[width=0.9\linewidth]{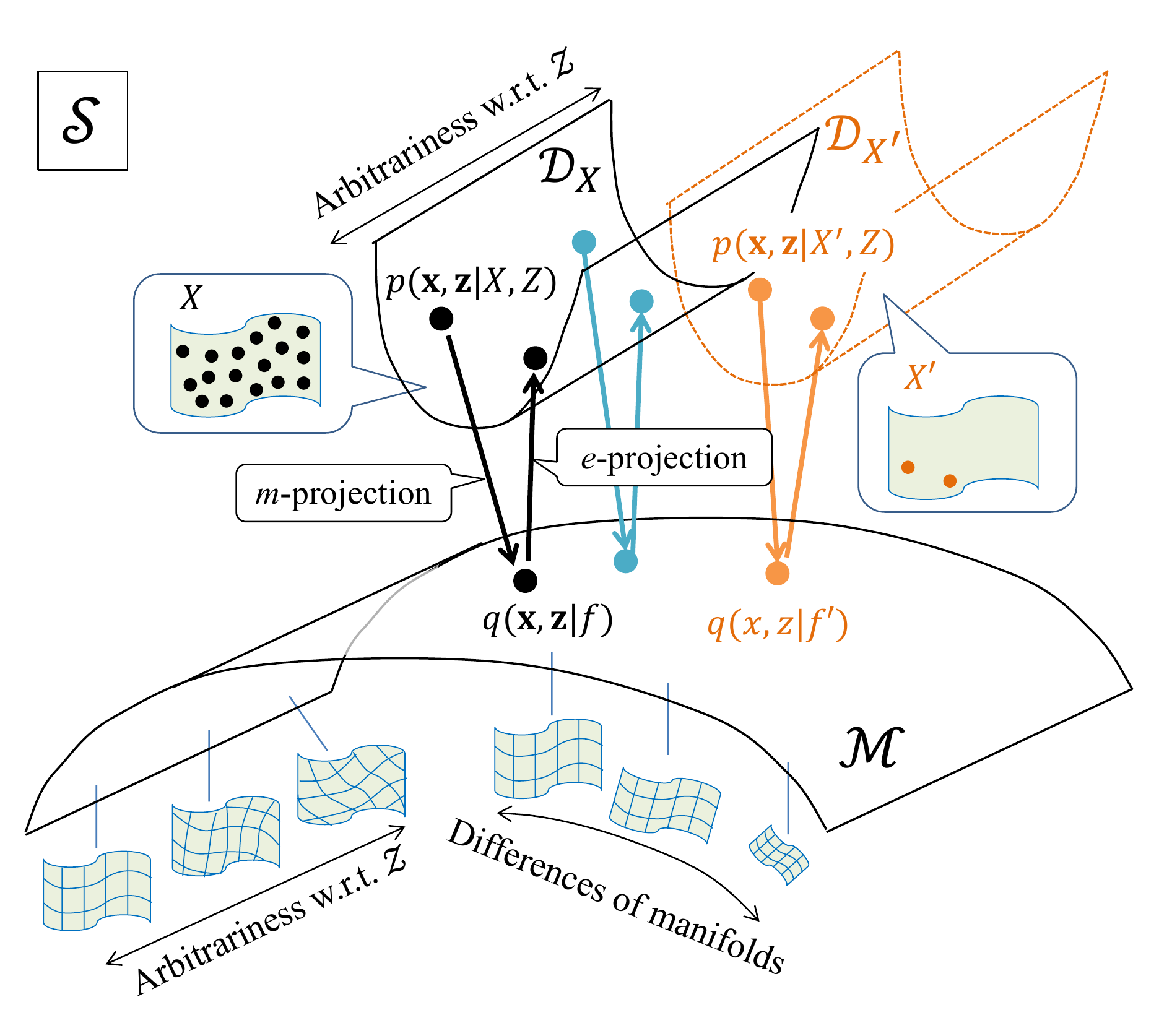}\\(a)\\
    \includegraphics[width=\linewidth]{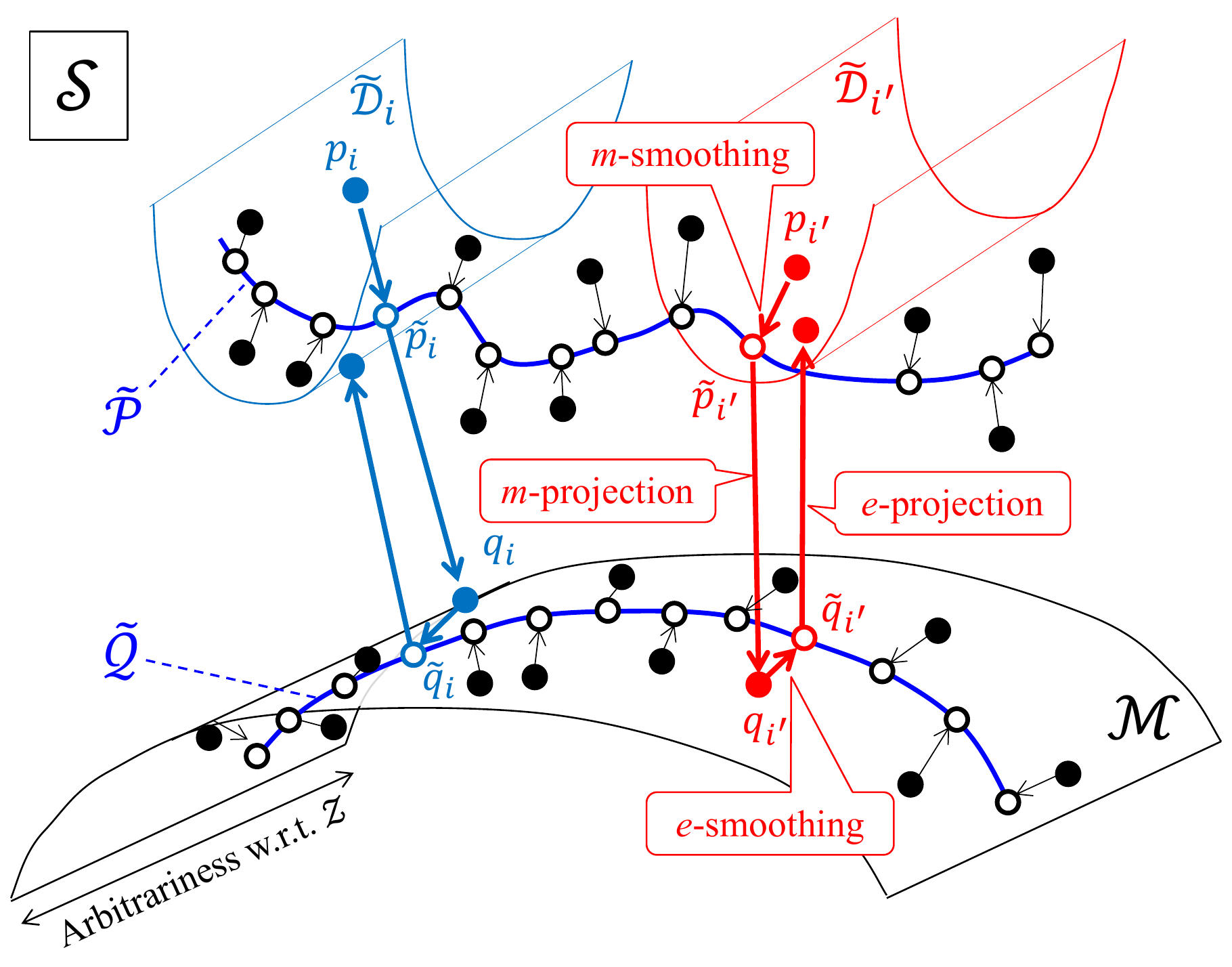}\\(b)
    \caption{Conceptual figures from the viewpoint of information geometry. (a) Single-task case. The learning process of manifold modeling is the iteration of \m- and \e-projections. The arbitrariness with respect to the coordinate system of the latent space $\cL$ is indicated for the depth direction (e.g., black and blue represent equivalent solutions). When the number of training samples is small, the data distribution becomes further from the true distribution (indicated in orange). (b) Multi-task case. Instance transfer is translated as kernel smoothing in the \m-mixture manner (\m-smoothing), which generates manifold $\Tilde\cP$. By contrast, model transfer is kernel smoothing in the \e-mixture manner (\e-smoothing), which generates manifold $\Tilde\cQ$. Additionally, \e-smoothing solves the manifold alignment problem.}
    \figlabel{scheme}
\end{figure}

\subsection{Application to multi-task supervised learning}

Although the proposed method was designed for multi-task manifold learning, the concepts of instance transfer and model transfer can be applied to other learning paradigms. \figref{ex4} shows two toy examples of multi-task regression. In this experiment, 400 tasks were generated by slanted sinusoidal functions $s=(au) \sin (t+bu) +(cu)t +du$, where $t$ and $s$ are the input and output, respectively; $u$ is the task latent variable; and $a, b, c, d$ are fixed parameters. There were 2 S/T, which were generated by the uniform random input.

In the first case (\figref{ex4} (a)), the input was distributed in $[-1,+1]$ uniformly. For this dataset, the latent variable $z$ was simply replaced by the input $t$, whereas the corresponding output $s$ was regarded as the observed data $x$. Thus, Step 5 in Algorithm \ref{algorithm} was omitted. As shown in \figref{ex4} (a), the proposed method successfully estimated the continuous change of sinusoidal functions from small sample sets.

The proposed method can be applied to multi-task regression when domain shifts occur. In the second case (\figref{ex4} (b)), the input distribution also changed according to the task latent variable $u$. For this case, MT-KSMM was directly applied by regarding each input--output pair $(t,s)$ as the training sample of MT-KSMM; thus, $\vx=(t,s)$. Although the border area was shrunk because of the absence of training samples, MT-KSMM successfully captured the continuous change of the function shapes, as well as the change of the input domain. Note that the estimation of the input distribution was also difficult for the small sample case.

\section{Discussion} \label{sec:discussion}

The key idea of the proposed method is to use two types of information transfer: instance transfer and model transfer. We discuss how these transfers work from the viewpoint of information geometry. We also discuss the problem formulation of multi-task manifold learning from the viewpoint of optimal transport.

\subsection{From the information geometry viewpoint}

Generative manifold model can be regarded as an infinite Gaussian mixture model (GMM) in which the Gaussian components are arranged continuously along the nonlinear manifold. Like the ordinary GMM, the EM algorithm can be used to solve it \cite{Heskes2001,Verbeek2005}. We first depict the single-task KSMM case from the information geometry viewpoint (\figref{scheme} (a)) and then describe the multi-task case (\figref{scheme} (b)).

\subsubsection{Single-task case}

Information geometry is geometry in statistical space $\cS$, which consists of all possible probability distributions. In our case, $\cS$ is the space that consists of all joint probabilities of $\vx$ and $\vz$, that is, $\cS=\{p(\vx,\vz)\}$.

When both $X=\{\vx_n\}_{n=1}^N$ and $Z=\{\vz_n\}_{n=1}^N$ are given, we assume that the data distribution is represented as
\begin{align*}
    & p(\vx,\vz\mid X,Z) \nncr
    &\qquad  =\frac1N \sumN \cN(\vx\mid\vx_n,\betai\vI_\DV)\, \cN(\vz\mid\vz_n, \betai\vI_\DL).
\end{align*}
Thus, the data distribution $p(\vx,\vz \mid X,Z)$ corresponds to a point in $\cS$. Because dataset $X$ is known, whereas the latent variables $Z$ are unknown, the set of data distributions for all possible $Z$ becomes a data manifold $\cD_X=\{p(\vx,\vz\mid X,Z)\mid Z=(\vz_n)\in \cL^N\}$. Therefore, estimating $Z$ from $X$ means determining the optimal point in the data manifold $\cD_X$.

It should be noted that there is arbitrariness with respect to the coordinate system of $\cL$. For example, if the coordinate system of $\cL$ is rotated or nonlinearly distorted, we obtain another equivalent solution of $Z$. Therefore, there is an infinite number of equivalent data distributions with respect to $Z$ (indicated for the depth direction in \figref{scheme} (a)). Additionally, when the number of samples is small, $\cD_X$ is far from the true distribution, and there is a large variation depending on the sample set $X$.

Similarly, we can consider a manifold consisting of all possible model distributions. In this study, we represent a model distribution as
\begin{align}
    q(\vx,\vz\mid f) = \cN(\vx\mid f(\vz),\betai\vI)\,p(\vz), \notag
\end{align}
where $f\in\cH$. Thus, the set of all possible model distributions forms a model manifold $\cM$, which is homeomorphic to $\cH$. Therefore, estimating $f$ from $X$ means determining the optimal point in $\cM$.  Like the case of data distribution, there is also arbitrariness with respect to the coordinate system of $\cL$.

The aim of KSMM is to minimize the cross-entropy between $p(\vx,\vz\mid X,Z)$ and $q(\vx,\vz\mid f)$, that is, minimize the Kullback--Leibler (KL) divergence $\DKL{p}{q}$\footnote{Rigorously speaking, the self-entropy of $p(\vx,\vz\mid X,Z)$ needs to be considered because it also depends on $Z$.}. The KSMM algorithm is the EM algorithm by which KL-divergence is minimized with respect $f$ and $Z$ alternately. From the information geometry viewpoint, the optimal solutions $\Hat Z$ and $\Hat f$ are given as the nearest points of two manifolds $\cD_X$ and $\cM$. To determine the optimal pair $(\Hat Z, \Hat f)$, projections between two manifolds are executed iteratively. Thus, in the M step, $p$ is projected to $\cM$, whereas in the E step, $q$ is projected to $\cD_X$. These two projections are referred to as {\em m-projection} and {\em e-projections} (\figref{scheme} (a)). \m- and \e-projections are executed repeatedly until $p$ and $q$ converge to the nearest points of $\cD_X$ and $\cM$ \cite{Amari1995}. We note again that there are infinite solution pairs $(\Hat Z,\Hat f)$ because of the arbitrariness of the coordinate system of $\cL$.

\subsubsection{Information transfers}

As shown in \eqref{m-ks}, the data distribution set $P=\{p_i\}$ is mixed by the instance transfer as
\begin{align}
    \Tilde p_i(\vx,\vz) \coloneqq \sum_{i'} \alpha^m_{ii'}\, p_{i'}(\vx,\vz),
  \eqlabel{m-smoothing}
\end{align}
where
\begin{align}
    \alpha^m_{ii'} = \frac{\rho(\vu_i\mid\vu_{i'})}{\sum_{i''} \rho(\vu_i\mid\vu_{i''})}. \notag
\end{align}
Then we have $\Tilde P=\{\Tilde p_i\}$. In information geometry, Eq.~\eqref{m-smoothing} is called the \m-mixture of the data distribution set $P$. Therefore, the instance transfer is translated as the {\em transfer by the \m-mixture} in terms of information geometry.

By contrast, as shown in \eqref{model_transfer3}, the log of the model distributions $Q=\{q_i\}$ are mixed by the model transfer as
\begin{align}
  \log \Tilde q_i(\vx,\vz) \coloneqq \sum_{i'} \alpha^e_{ii'}\log q_{i'}(\vx,\vz) + \const,
  \eqlabel{e-smoothing}
\end{align}
where
\begin{align}
    \alpha^e_{ii'} = \frac{h_\cT(\vu_i\mid\vu_{i'})}{\sum_{i''} h_\cT(\vu_i\mid\vu_{i''})}. \notag
\end{align}
Then we have $\Tilde Q=\{\Tilde q_i\}$. Eq.~\eqref{e-smoothing} is  called the \e-mixture of model distributions $Q$ in information geometry. Therefore, the model transfer is translated as the {\em transfer by the \e-mixture}. It has been shown that the \e-mixture transfer improves multi-task learning in estimating the probability density \cite{Takano2016}.

\subsubsection{Multi-task case}

Based on the above discussion, the learning process of MT-KSMM can be interpreted as follows (\figref{scheme} (b)): Suppose that we have $I$ data distributions $P=\{p_i\}$. In Step~1, $P=\{p_i\}$ are mapped to $\Tilde P=\{\Tilde p_i\}$ using instance transfer (i.e., \m-mixture transfer). However, MT-KSMM not only models the given tasks, but can also model unknown tasks. Therefore, we have a data distribution $\Tilde p(\vx,\vz\mid\vu)$ for any $\vu$:
\begin{align}
    \Tilde p(\vx,\vz\mid\vu)
    &= \frac{\sum_i \rho(\vu\mid\vu_i) p_i(\vx,\vz\mid X_i, Z_i)}
       {\sum_{i'} \rho(\vu\mid\vu_{i'})}. \notag
\end{align}
As a result, we have a smooth manifold $\Tilde\cP=\{\Tilde p(\vx,\vz\mid\vu)\mid \vy\in\cT\}$ in $\cS$, which is homeomorphic to $\cT$. Manifold $\Tilde\cP$ is obtained by the kernel smoothing of the data distributions $P=\{p_i\}$ in the \m-mixture manner. We refer to this process as {\em \m-smoothing}. Thus, the instance transfer can be translated as \m-smoothing, which generates the manifold $\Tilde \cP$.

In Step~2, manifold $\Tilde\cP$ is projected to $\cM$ by the \m-projection \eqref{step2a}, and we obtain the model distribution set $Q=\{q_i\}$ in $\cM$. Then, in Step~3, we obtain a manifold $\Tilde\cQ$ from $Q$:
\begin{align}
    \log \Tilde q(\vx,\vz\mid\vu) &\eqc
    \frac{\sum_i h_\cT(\vu\mid\vu_i)\,\log q_i(\vx,\vz\mid f_i)}
    {\sum_{i'} h_\cT(\vu\mid\vu_{i'})}. \notag
\end{align}
Because manifold $\Tilde\cQ$ is obtained by the kernel smoothing of $Q$ in the \e-mixture manner, we refer to this process as {\em \e-smoothing}. Thus, the model transfer is now regarded as \e-smoothing, which generates the manifold $\Tilde\cQ$ in $\cM$. In the manifold $\Tilde\cQ$, the model distributions of unknown tasks are also represented.

After \e-smoothing, in Steps~4 and 5, the latent variables $Z=\{\vz_n\}$ and $U=\{\vu_i\}$ are estimated by the \e-projection, by which $\Tilde\cP$ is updated. Thus, the MT-KSMM algorithm is depicted as the bidirectional \m- and \e-projections of $\Tilde\cP$ and $\Tilde\cQ$, which are generated by \m- and \e-smoothing, respectively.

It is worth noting that \m-smoothing tends to make the distributions broader, whereas \e-smoothing tends to make them narrower. Thus, if either \m- or \e-smoothing is used alone, it will not work as expected, particularly when S/T is small. Therefore, the combination of \m- and \e-smoothing is essentially important for multi-task manifold modeling. Additionally, \e-smoothing plays another important role in MT-KSMM{}. By performing \e-smoothing, the neighboring models come closer to each other in $\cM$, and their coordinate systems also come closer to each other. As a result, the manifolds are aligned between the tasks, and the latent variables acquire a consistent representation.

To summarize, the MT-KSMM algorithm consists of the following steps: (Step 1) By \m-smoothing, we obtain manifold $\Tilde\cP$ from the data distribution set $P$. This process is the instance transfer. (Step 2) $\Tilde\cP$ is projected to $\cM$ by \m-projection, and we obtain the model distribution set $Q$. (Step 3) By \e-smoothing, we obtain manifold $\Tilde\cQ$ from $Q$. This process is the model transfer. (Steps 4 and 5) $\Tilde\cQ$ is projected to $\cD$ by \e-projection, and latent variables $U$ and $Z$ are updated.

In this discussion, we described the learning process of MT-KSMM from the viewpoint of information geometry. We discussed the depiction of two manifolds $\Tilde\cP$ and $\Tilde\cQ$ in $\cS$, which are generated by \m- and \e-smoothing, respectively, and are projected alternately by \m- and \e-projections, respectively. We believe that such geometrical operations are the essence of the two information transfer styles in multi-task manifold modeling.

\subsection{From the optimal transport viewpoint}

In this study, we assumed that the latent space $\cL$ was common to all tasks. For example, for the face image set, we expected the latent variable $\vz_n$ to represent the pose or expression of the face image $\vx_n$ regardless of which task it belonged to. Thus, if we had two manifolds $\cX_1$ and $\cX_2$, and the corresponding embeddings $f_1$ and $f_2$, we considered that $\vx_1=f_1(\vz)\in\cX_1$ and $\vx_2=f_2(\vz)\in\cX_2$ would represent the same intrinsic property, even if $\vx_1\ne \vx_2$. Based on this assumption, information was transferred between tasks. However, some questions arise. The first question is how we can know whether latent variables that belong to different tasks represent the same intrinsic property. The second question is how we can regularize multi-task learning so that the latent variables represent the common intrinsic property. Note that the intrinsic property is unknown. Therefore, we need to redefine the objective of multi-task manifold learning without using such ambiguous terms.

In this study, we defined the distance between two manifolds $\cX_1$ and $\cX_2$ as \eqref{distance}. Thus, to measure the distance, the corresponding embeddings $f_1$ and $f_2$ need to be provided. However, because there is arbitrariness with respect to the coordinate system of $\cL$, there are many equivalent embeddings that generate the same manifold. Therefore, if $f_1, f_2$ are not specified, the distance between two manifolds cannot be determined uniquely. In such a case, it is natural to choose the pair $(f_1, f_2)$ that minimizes \eqref{distance}. Thus, the distance $D(\cX_1,\cX_2)$ is defined as
\begin{align}
    D^2[\cX_1,\cX_2] = \min_{\substack{f_1\in \cF_{\cX_1}\\ f_2\in\cF_{\cX_2}}}
      \int_\cL \norm{f_1(\vz)-f_2(\vz)}^2\,dP(\vz),
  \eqlabel{distance2}
\end{align}
where $\cF_{\cX_i}$ is a set of embeddings, which yield the equivalent data distribution on $\cX_i$; that is, for any $f_i,f_i'\in\cF_{\cX_i}$, they satisfy $q(\vx\mid f_i)=q(\vx\mid f'_i)$, where $q(\vx\mid f_i)=\int\cN(\vx\mid f_i(\vz),\betai\vI)\,dP(\vz)$. Such a distance between manifolds \eqref{distance2} becomes the optimal transport distance between $\cX_1$ and $\cX_2$. Based on this, we can say that manifolds $(\cX_1, \cX_2)$ are aligned with respect to the coordinate system induced by $(f_1, f_2)$, if and only if $(f_1, f_2)$ minimizes \eqref{distance2}. Additionally, if two manifolds are aligned, we assume that the latent variable represents the intrinsic property that is independent of the task.

Now we recall the problem formulation in terms of the fiber bundle (\figref{Model} (d)), in which each task manifold is regarded as a fiber. Then we can consider the integral of the transport distance between fibers along the base space $\cT$, which equals the length of manifold $\cY\subseteq\cH$ (\figref{Model} (c)). When all task manifolds are aligned with neighboring ones, the total length of manifold $\cY$ is expected to be minimized. As already described, SOM (in addition to KSMM) works like an elastic net. Therefore, roughly speaking, the higher-KSMM tends to shorten the length of manifold $\cY$, whereby the task manifolds are gradually aligned. Therefore, in MT-KSMM, the transport distance along the base space is implicitly regularized to be optimized. It would be worth trying to implement such optimal transport-based regularization in the objective function explicitly, and this is our future work. The underlying learning theory from the optimal transport viewpoint would be also an important issue in the future.

\section{Conclusion}

In this paper, we proposed a method for multi-task manifold learning by introducing two information transfers: instance transfer and model transfer. Throughout the experiments, the proposed method, MT-KSMM performed better than \KSMM2 and single-task KSMM{}. Because in the comparison we consistently used KSMM as a platform for manifold learning, we can conclude that these results were only caused by differences in the manner of information transfer between tasks. Thus, the combination of instance transfer and model transfer are effective for multi-task manifold learning, particularly when the sample per task is small. It should be noted that \KSMM2 (i.e., the preceding method \SOM2) learns appropriately when sufficient samples are provided. Because \KSMM2 uses model transfer only, we can also conclude that instance transfer is necessary for the small sample size case. By contrast, model transfer is necessary for manifold alignment, whereby the latent variables are estimated consistently across tasks, representing the same intrinsic feature of the data.

The proposed method projects the given data into the product space of the task latent space and sample latent space, in which the intrinsic feature of the data content is represented by the sample latent variables, whereas the intrinsic feature of the data style is represented by the task latent variable. For example, in the case of the face image set, `content' refers to the expressed emotions and `style' refers to the individual difference in the manner of expressing emotions. Such a learning paradigm is referred to as {\em content--style disentanglement} in the recent literature \cite{Kazemi2019}. Therefore, the proposed method is not only multi-task manifold learning under the small sample size condition, but also has the ability to perform content--style disentanglement under the condition where data are observed partially for each style. Because recent studies on content--style disentanglement implement the autoencoder (AE) platform \cite{Na2020}, which typically requires a larger number of samples, applying the proposed information transfers to an AE would be a challenging theme in future work.

\subsubsection*{Acknowledgements}

We thank S. Akaho, PhD, from National Institute of Advanced Industrial Science and Technology of Japan, who gave us important advice for this study.
This work was supported by JSPS KAKENHI [grant number 18K11472, 21K12061, 20K19865]; and ZOZO Technologies Inc. We thank Maxine Garcia, PhD, from Edanz (https://jp.edanz.com/ac) for editing a draft of this manuscript.


\begin{table*}[p]
\centering
\caption{Symbol list used in this paper}\tbllabel{symbols}
\small
\begin{tabular}{ll}
  \toprule
  \multicolumn{2}{c}{For problem formulation (single task case)} \\
  \midrule
  $\cV \equiv \bbR^{D_\cV}$ & High-dimensional visible data space with dimension $D_\cV$\\
  $\cL \subseteq \bbR^{D_\cL}$ & Low-dimensional latent space with dimension $D_\cL$\\
  $X=\{\vx_n\}_{n=1}^N$ & Observed dataset containing $N$ samples, where $\vx_n\in\cV$ \\
  $Z=\{\vz_n\}_{n=1}^N$ & Latent variable set of samples, where $\vz_n\in\cL$ \\
  $\cX\subseteq\cV$  & Manifold representing the data distribution \\
  $f(\vz)\equiv\pi\inv(\vz)$ & Embedding from $\cL$ to $\cX$, referred to as the {\em model} \\
  $p(\vz)$ & Prior of $\vz\in\cL$ \\
  $q(\vx,\vz\mid f)$ & Probabilistic generative model of $\vx$ and $\vz$\\
  $\beta$ & Inverse variance (precision) of observation noise \\
  \toprule \multicolumn{2}{c}{For problem formulation (multi-task case)} \\
  \midrule
  $X_i=\{\vx_{ij}\}_{j=1}^{J_i}$ & Dataset of the $i$th task containing $J_i$ samples\\
  $X=\{\vx_n\}_{n=1}^N$ & Entire dataset of $I$ tasks, where $N=\sum_i J_i$\\
  $\vX=\big(\vx_n^T\big)$ & $N\times D_\cV$ design matrix consisting of the entire dataset of $I$ tasks \\
  $I$ & Number of tasks \\
  $i_n$ & Task to which sample $\vx_n$ belongs \\
  $\cN_i=\{\vx_n\mid i_n=i\}$ & Sample set belonging to task $i$\\
  $\cX_i$ & Manifold of task $i$\\
  $f_i\in\cH$ & Embedding from $\cL$ to $\cX_i$, referred to as the {\em task model} \\
  $\cH$ & Function space (RKHS) consisting of embeddings from $\cL$ to $\cV$\\
  $\cT\subseteq\bbR^{D_\cT}$ & Low-dimensional latent space for tasks with dimension $D_\cT$ \\
  $\cY\subseteq\cH$ & Task manifold embedded into $\cH$ \\
  $U=\{\vu_i\}_{i=1}^I$ & Latent variable set for tasks, where $\vu_i\in\cT$ \\
  $g(\vu)$ & Embedding from $\cT$ to $\cY$ \\
  $G(\vz,\vu)\equiv\left[g(\vu)\right](\vz)$ & Embedding from $\cL\times\cT$ to $\cV$, referred to as the {\em general model} \\
  $p(\vu)$ & Prior of $\vu\in\cT$ \\
  $q_i(\vx,\vz\mid f_i)$ & Probabilistic generative model of task $i$\\
  $q(\vx,\vz,\vu \mid G)$ & Probabilistic generative model of $\vx$, $\vz$, and $\vu$ \\
  \toprule \multicolumn{2}{c}{For KSMM (single task case)} \\
\midrule
$h_\cL(\vz\mid\vz')\equiv\N{\vz}{\vz',\lambda^2_\cL\vI}$
    & Non-negative smoothing kernel with length constant $\lambda_\cL$\\
$\Hat Z = \{\Hat\vz_n\}_{n=1}^N$ & Tentative estimators of latent variables $Z=\{\vz_n\}$\\
$\Hat f$ & Tentative estimator of mapping $f$ \\
$p(\vx,\vz\mid X,Z)$ & Joint empirical distribution of $\vx$ and $\vz$ under $X$ and $Z$ are given\\
$\vvphi(\vz)=\left(\varphi_l(\vz)\right)$ & Basis functions for a parametric representation \\
$\vV$ & Coefficient matrix to represent $f$ parametrically\\
\toprule
\multicolumn{2}{c}{Multi-task case (MT-KSMM)} \\
\midrule
$\rho(\vu_i,\vu_{i'})$ & Function that determines the weight of the instance transfer from task $i'$ to $i$ \\
$\lambda_\rho$ & Length constant of $\rho$ \\
$\Tilde X_i=\{(\vx_n,\rho_{in})\}$ & Merged dataset of task $i$ obtained by instance transfer\\
& $\Tilde X$ is a weighted set where $\rho_{in}$ denotes the weight of $\vx_n$\\
$\Tilde Z_i=\{\vz_n, \rho_{in})\}$ & Merged latent variable set obtained by instance transfer\\
$\Tilde J_i\equiv\sum_n \rho_{in}$ & Number of merged sample sets of task $i$\\
$\Tilde f_i(\vz)$ & Task model $f_i$ after model transfer is executed \\
$\Tilde p_i(\vx,\vz\mid\Tilde X_i, \Tilde Z_i)$ & Joint empirical distribution after instance transfer is executed\\
$\Tilde q_i(\vx,\vz\mid \Tilde f_i)$ & Generative model for the $i$th task after model transfer is executed \\
$h_\cT(\vu,\vu')\equiv\N{\vu}{\vu',\lambda^2_\cT\vI}$ & Smoothing kernel for the higher-KSMM with length constant $\lambda_\cT$\\
$\vpsi(\vu)=\left(\psi_k(\vu)\right)$ & Basis function for the higher-KSMM \\
$\vV_i$ & Coefficient matrix to represent task model $f_i$ parametrically\\
$\uV=(\vV_i)$ & Coefficient tensor consisting of $\{\vV_i\}$\\
$\uW$ & Coefficient tensor to represent general model $G$ parametrically\\
\bottomrule
\end{tabular}
\end{table*}

\appendix
\section{Symbols and notation}
In this paper, scalars and functions are usually written in italics (e.g., $\lambda$, $f$, and $G$). Indices and their upper limits are written in lowercase and uppercase italics, respectively (e.g., $n$ and $N$). Sets are also written in uppercase italics (e.g., $X$ and $Z$). Vectors and matrices are written in lowercase and uppercase boldface, respectively (e.g., $\vx$ and $\vX$), whereas tensors are written in uppercase boldface and underlined (e.g., $\uW)$. Continuous spaces and manifolds are written in cursive script (e.g., $\cV$ and $\cX$).

The list of symbols used in this paper is shown in \tblref{symbols}.

\section{Details of the equation derivation}
\subsection{Derivation of the KSMM algorithm}

The cost function of KSMM is given by \eqref{KSMM1}. First, we show that this cost function equals the cross-entropy between \eqref{p(x,z)} and \eqref{q(x,z)}, except the constant. The cross-entropy is given by
\begin{align*}
    &H[p(\vx,\vz),q(\vx,\vz)] \\
    &= -\iint p(\vx,\vz) \log q(\vx,\vz)\,d\vx\,d\vz \\
    &= -\frac1N \sum_n \iint \N{\vx}{\vx_n,\beta\inv\vI} h_\eL(\vz\mid\vz_n) \\
    & \qquad \times \log \left[\N{\vx}{f(\vz),\beta\inv\vI}\,p(\vz)\right]\,d\vx\,d\vz.
\end{align*}
Note that, in this study, we assume that the prior $p(\vz)$ is a uniform distribution. Thus,
\begin{align*}
    &H[p(\vx,\vz),q(\vx,\vz)] \\
    &= -\frac1N\sum_n\iint \N{\vx}{\vx_n,\beta\inv\vI}\,h_\cL(\vz\mid\vz_n) \\
    &\qquad \times \left[-\frac\beta2\norm{\vx-f(\vz)}^2+\frac{D_\cV}{2}\log\frac{\beta}{2\pi}\right]
      \,d\vx\,d\vz \\
    &= \frac\beta{2N}\sum_n \int h_\cL(\vz\mid\vz_n)\norm{\vx_n-f(\vz)}^2\,d\vz +\const,
\end{align*}
and we obtain the cost function \eqref{KSMM1}. We apply the expectation of the square error for the normal distribution:
\begin{align*}
    \int_\cV \N{\vx}{\vmu,\beta\inv\vI}\,{\norm{\vx-f(\vz)}^2}\,d\vx
    &= \norm{\vmu-f(\vz)}^2 + \frac{D_\cL}{\beta}.
\end{align*}

To estimate the latent variables, we need to minimize the cost function \eqref{KSMM1} with respect to each $\vz_n$ as \eqref{KSMM-Estep1}. Because $\Hat f(\vz)$ is obtained by kernel smoothing, $\Hat f(\vz)$ can be considered sufficiently smooth to be locally approximated as linear. By considering the first-order Taylor expansion, the square error $\norm{f(\vz)-\vx_n}^2$ is approximated as a quadratic function centered on the minimum point. Because $h_\cL(\vz\mid\vz_n)\equiv\N{\vz}{\vz_n,\lambda_\cL^2\vI}$ is symmetric around $\vz_n$, the minimum point of \eqref{KSMM-Estep1} equals the minimum point of \eqref{KSMM-Estep2}.

By contrast, to estimate mapping $f$, we need to optimize \eqref{KSMM1} using the variation technique.
Because the functional derivative of $E$ is given by
\begin{align*}
    \frac{\delta E}{\delta f(\vz)}
    &= \frac\beta{N} \sum_n h_\cL(\vz\mid\vz_n) \left(f(\vz)-\vx_n\right),
\end{align*}
the stationary point of $f$ should satisfy
\begin{align*}
    \sum_n h_\cL(\vz\mid\vz_n)\,f(\vz) = \sum_n h_\cL(\vz\mid\vz_n)\,\vx_n.
\end{align*}
Thus, the optimal $f$ is determined as
\begin{align*}
    f(\vz) &= \frac{\sum_n h_\cL(\vz\mid\vz_n)\,\vx_n}{\sum_{n'} h_\cL(\vz\mid\vz_{n'})},
\end{align*}
and we obtain \eqref{KSMM-Mstep2}.

\subsection{Derivation of the MT-KSMM algorithm}

The extension from KSMM to MT-KSMM is straightforward: The cost function of the lower-KSMM \eqref{costfunction1} is an extension of \eqref{KSMM1} for a weighted dataset. Thus, we obtain \eqref{step4} by modifying \eqref{KSMM-Estep2}, where the weights of data are given by $\rho_{in}$. Similarly, we have \eqref{step2} by extending \eqref{KSMM-Mstep2} considering $\rho_{in}$.

The cost function of the higher-KSMM \eqref{costfunction2} looks the same as \eqref{KSMM1}.
The difference is that the square norm in \eqref{KSMM1} is defined in the ordinal vector space $\cV$, whereas the square norm in \eqref{costfunction2} is defined in the RKHS $\cH$. Similarly, we obtain
\eqref{step4} and \eqref{step3}.

\subsection{Derivation of the parametric representation of KSMM}

Using the orthonormal basis set $\{\varphi_1,\dots,\varphi_L\}$, we represent the mapping $f$ parametrically as
\begin{align*}
    f(\vz\mid\vV) &= \sum_l \varphi_l(\vz)\vv_l = \vV\T\vvphi(\vz),
\end{align*}
where $\vvphi(\vz)=\left(\varphi_1(\vz),\dots,\varphi_L(\vz)\right)\T$. In this case, the cost function \eqref{KSMM1} becomes
\begin{align*}
   E &=\frac\beta{2N} \sum_n \int h_\cL(\vz\mid\vz_n) \norm{\vV\T\vvphi(\vz)-\vx_n}^2\,d\vz.
\end{align*}
If we differentiate $E$ with respect to $\vV$, then
\begin{align*}
    \fracp{E}{\vV}
    &= \frac\beta{N} \sum_n \int h_\cL(\vz\mid\vz_n)\,
    \vvphi(\vz)\,\left(\vV\T\vvphi(\vz)-\vx_n\right)\T\,d\vz.
\end{align*}
Because the stationary point of $\vV$ should satisfy $\partial E/\partial \vV=0$, we have the following equation:
\begin{align}
    &\sum_n \int h_\cL(\vz\mid\vz_n) \, \vvphi(\vz) \vvphi\T(\vz) \vV\,d\vz
  \eqlabel{left_eq} \\
    &\qquad = \sum_n \int h_\cL(\vz\mid\vz_n) \vvphi(\vz)\,\vx_n\T\,d\vz.
  \eqlabel{right_eq}
\end{align}
The left-hand side \eqref{left_eq} becomes
\begin{align*}
    &\left[\int \left(\sum_n h_\cL(\vz\mid\vz_n)\right)\,\vvphi(\vz)
      \vvphi\T(\vz)\,d\vz\right]\,\vV \\
    & = \left[\int \ol{h}_\cL(\vz)\,\vvphi(\vz)\,\vvphi\T(\vz)\,d\vz\right]\,\vV \\
    &\eqqcolon \vA\vV,
\end{align*}
where $\ol{h}_\cL(\vz)=\sum_n h_\cL(\vz\mid\vz_n)$ and $\vA$ is an $L\times L$ matrix. By contrast, the right-hand side \eqref{right_eq} becomes
\begin{align*}
    &\int \vvphi(\vz) \,
      \left(h_\cL(\vz\mid\vz_1),\dots,h_\cL(\vz\mid\vz_N)\right)\,
      \begin{pmatrix}\vx_1\T\\ \vdots\\ \vx_N\T\end{pmatrix}\,d\vz \\
    &= \left(\int\vvphi(\vz)\,\vh_\cL(\vz)\T\,d\vz\right)\,\vX \\
    &\eqqcolon \vB\vX.
\end{align*}
Because $\vA\vV=\vB\vX$ holds, the estimator of $\vV$ is given by
\begin{align}
    \Hat\vV = \vA\inv \vB \vX.
  \eqlabel{appendix-eq2}
\end{align}
Thus, we have \eqref{step2a}. In the KSMM algorithm, the tentative estimator of latent variables $\Hat Z=\{\Hat\vz_n\}$ is used instead of $Z=\{\vz_N\}$.

\subsection{Derivation of the parametric representation of MT-KSMM}
Similar to the non-parametric case, we obtain the parametric representation of MT-KSMM by a straightforward extension, but a tensor--matrix product is required in the notation.

First, we introduce some tensor operations. Suppose that $\vX=(X_{mn})$ is an $M\times N$ matrix, and let $\check \vx\equiv \mathrm{vec}(\vX)$ be the vector representation of $\vX$. Thus, $\check\vx$ is an $(M\times N)$-dimensional vector obtained by {\em flattening} $\vX$. When we have $L$ such matrices $\{\vX_1,\dots,\vX_L\}$, the entire matrix set is represented as a matrix $\check\vX = (\check\vx_1,\dots,\check\vx_L)$. Note that $\check\vX$ is an $L\times (M\times N)$ matrix, which is also denoted by $\uX=(X_{lmn})$ in tensor notation. We further suppose that $\check\vY=\vA \check \vX$ is a linear transformation, where $\check\vY$ is an $I\times (M\times N)$ matrix, whereas $\vA$ is an $I \times L$ matrix. Using the tensor--matrix product, this linear transformation is denoted by $\uY=\uX\ttimes1\vA$, which means $Y_{imn} = \sum_l A_{il}X_{lmn}$ in element-wise notation, where $\uY=(Y_{imn})$ is a tensor of order three and size $I\times M \times N$. Similarly, $\uY=\uX\ttimes1\vA\ttimes2\vB$ means $Y_{ijn}=\sum_l\sum_m A_{il}B_{jm}X_{lmn}$.

Using the above notation, the parametric representation of MT-KSMM becomes as follows: For the lower-KSMMs, we estimate the coefficient matrix $\vV_i$ as $\vV_i = \vA_i\inv\vB\vX_i$ for each task $i$; using this, the task model is represented as $f_i(\vz)=\vV_i \vvphi(\vz)$. Consequently, the entire coefficient matrices become a tensor $\uV=(\vV_i)$ of order three and size $I\times L \times D_\cV$. By flattening $\vV_i$ using $\check\vv_i=\mathrm{vec}(\vV_i)$, $\uV$ can also be represented by a matrix $\check\vV=(\check\vv_i)$. Because we use an orthonormal basis set $\{\vvphi_l\}$, the metric in function space $\cH$ equals the Euclidean metric in the ordinary vector space of the coefficient matrices. Thus, we consider that $\{\check\vv_i\}$ are equivalent to $\{f_i\}$, including the metric.

The higher-KSMM estimates $g(\vu)$ by regarding the task model set $\{f_i\}$ as a dataset. Thus, in the parametric representation, the aim of the higher-KSMM is to estimate a function $\vV(\vu)=g(\vu)$, where $\vu$ is the latent variable of tasks. By regarding $\check\vV$ as a data matrix and applying \eqref{appendix-eq2}, we have
\begin{align}
    \check\vW &= \vC\inv\vD\check\vV,
    \eqlabel{appendix-eq3}
\end{align}
where $\vC$ and $\vD$ are defined in \eqref{step3a-C} and \eqref{step3a-D}, respectively. Using tensor--matrix product notation, \eqref{appendix-eq3} is denoted by $\vW = \vV\times_1 (\vC\inv\vD)$.

Finally, the general model $G(\vz,\vu\mid \uW)$ is represented as follows: Note that \begin{align}
    G(\vz,\vu) &= \vV(\vu) \vvphi(\vz)
  \eqlabel{appendix-eq4}\\
    \check\vv(\vu) &= \check\vW \vpsi(\vu).
  \eqlabel{appendix-eq5}
\end{align}
Using tensor--vector product notation, \eqref{appendix-eq5} is denoted by $\vV(\vu) = \uW\ttimes1\vpsi(\vu)$, which means $V_{ld}(\vu)=\sum_k W_{kld}\psi_k(\vu)$. Then \eqref{appendix-eq4} becomes \eqref{general_model} as follows:
\begin{align*}
    G(\vz,\vu) &= \uW \ttimes1 \vpsi(\vu) \ttimes2 \vvphi(\vz),
\end{align*}
which means $G_d(\vz,\vu)=\sum_k \sum_l W_{kld}\psi_k(\vu)\varphi_l(\vz)$, where $G_d$ denotes the $d$th entry of the $D_\cV$-dimensional vector $G$.

\end{document}